%% file: faults_that_matter.tex
\documentclass[runningheads]{llncs}
\usepackage[T1]{fontenc}
\usepackage{geometry}
\usepackage[USenglish]{babel} %francais, polish, spanish, ...
\usepackage[ansinew]{inputenc}
\usepackage[noadjust]{cite}
\usepackage{lmodern} %Type1-font for non-english texts and characters
\usepackage{graphicx} %%For loading graphic files
\usepackage{float}
\usepackage{textcomp}
\usepackage{amsmath}
\usepackage{amsfonts}
\usepackage{amssymb}
\usepackage{bm}
\usepackage{url}
\usepackage{hyperref}
\usepackage{booktabs}
\usepackage{subcaption}
\usepackage{tabularx}
\usepackage{makecell}
\usepackage{romannum}
\usepackage[super]{nth}
\usepackage[dvipsnames]{xcolor}
\usepackage{academicons}
\graphicspath{{./Pics/}}
\renewcommand{\baselinestretch}{0.97}

\input{abbr}

\usepackage{pgfplots, gincltex}
\usepackage{lipsum}
\usepackage{tikz}
\usepackage{wrapfig}
\usetikzlibrary{patterns}
\pgfplotsset{compat=newest}
\pgfplotsset{plot coordinates/math parser=false}
\begin{document}

\title{\bf Hardware faults that matter: Understanding and Estimating the safety impact of hardware faults on object detection DNNs
% \thanks{This project has received funding from the European Union's Horizon $2020$ research and innovation programme under grant agreement No $956123$.}
}
\titlerunning{Hardware faults that matter}

\author{Syed Qutub\inst{1} \and Florian Geissler\inst{1} \and Yang Peng\inst{1} \and Ralf Gr\"afe\inst{1} \and Michael Paulitsch\inst{1} \and Gereon Hinz\inst{2} \and Alois Knoll\inst{2}}
\institute{Dependability Research Lab, Intel Labs, Germany \\
\email{syed.qutub@intel.com}
\and Technical University of Munich, Germany
}
\authorrunning{Syed Qutub .}
% \author{
% \authorblockN{Syed Qutub{$^*$}\thanks{{$^*$} Email: syed.qutub@intel.com}, Florian Geissler, Yang Peng, Ralf Gr\"afe, and Michael Paulitsch}\\
% \authorblockA{Dependability Research Lab, 
% Intel Labs\\
% 85579 Neubiberg, Germany
% %Email: florian.geissler@intel.com 
% } }
\date{\today}
\maketitle 

\vspace{-10pt}
\begin{abstract}
Object detection neural network models need to perform reliably in highly dynamic and safety-critical environments like automated driving or robotics. Therefore, it is paramount to verify the robustness of the detection under unexpected hardware faults like soft errors that can impact a system's perception module. % The system's knowledge building and decision-making process happens over a single or collection of predicted frames rather than . 
% The system builds its knowledge by understanding the entire scene and its decision making process considers a single or collection of predicted frames to
Standard metrics based on average precision produce model vulnerability estimates at the object level rather than at an image level. As we show in this paper, this does not provide an intuitive or representative indicator of the safety-related impact of silent data corruption caused by bit flips in the underlying memory but can lead to an over- or underestimation of typical fault-induced hazards. 
With an eye towards safety-related real-time applications, we propose a new metric \textbf{\novelMetric} (Image-wise Vulnerability Metric for Object Detection) to quantify vulnerability based on an incorrect image-wise object detection due to false positive (FPs) or false negative (FNs) objects, combined with a severity analysis.
The evaluation of several representative object detection models shows that even a single bit flip can lead to a severe silent data corruption event with potentially critical safety implications, with e.g., up to  $\gg 100$ FPs generated, or up to $\sim 90\%$ of true positives (TPs) are lost in an image.
Furthermore, with a single stuck-at-1 fault, an entire sequence of images can be affected, causing temporally persistent ghost detections that can be mistaken for actual objects (covering up to $\sim 83\%$ of the image). Furthermore, actual objects in the scene are continuously missed (up to $\sim 64\%$ of TPs are lost). 
Our work establishes a detailed understanding of the safety-related vulnerability of such critical workloads against hardware faults.

\end{abstract}

%Results:
%\begin{itemize}
%\item We quantify the intrinsic resilience of object detection models against the considered faults for different metrics (ap50, map, sdc, ratio of affected images). Demonstrate interpretability of results depending on metric and explain why.
%Only a few images are affected but large impact on metric possible.
%\item Trends in the fault severity (bit position, layer, dataset type [e.g. small targets?]). Can argue with large activations in most cases.
%\item Faults cause mostly FPs, move bounding boxes (less class confusions). Put numbers and examples.
%\item Sequences of images with constant fault (bit flip or stuck-at?), do mis-predictions persist long enough to have a relevant impact?
%\item (Ranger helps to fix the problem in all studied cases. Spare details about Ranger but merely refer to previous work.)
%\item (Outlook: towards a more safety-related metric, e.g. class clusters, objects in vicinity, scoreA/scoreB work with Sreetama.)
%\end{itemize}

% TODO:
% shows: larger values -> more severe faults

%TODO:
%\begin{itemize}
%\item Sequences of images: How to inject stuck-at fault?
%\item Sequences of image: Evaluate fault persistence
%\end{itemize}
\begin{figure}[!h]
\centering
\begin{subfigure}[b]{\textwidth}
	\centering
  \frame{\includegraphics[width=0.9\linewidth]{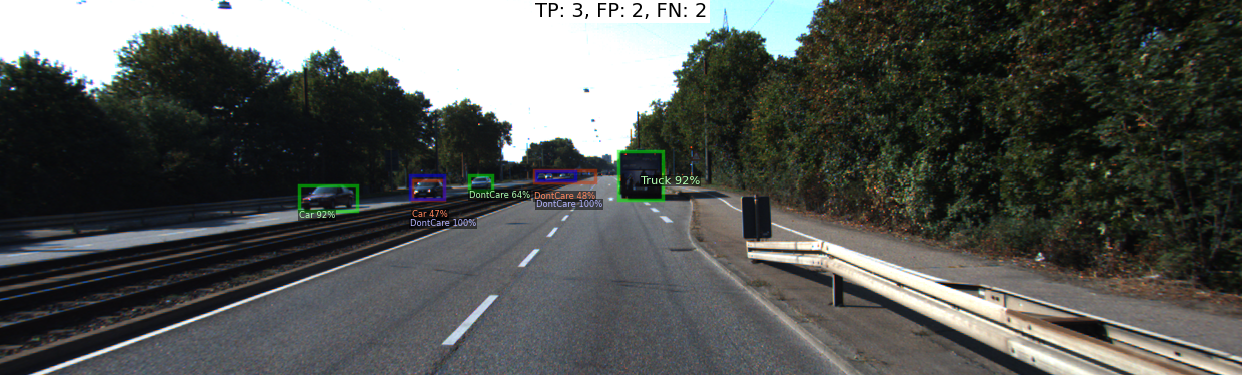}}
    %\caption{}
    \end{subfigure}
    \begin{subfigure}[b]{\textwidth}
    %\vspace{5pt}
    \centering
    \frame{\includegraphics[width=0.9\linewidth]{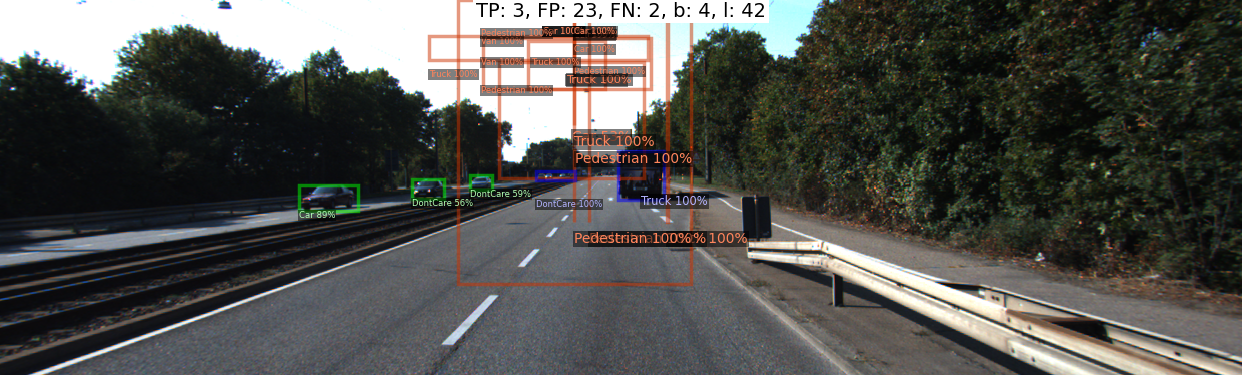}}
    \caption{Inference from Yolov3 model and Kitti dataset}
    \label{fig:example yolov3}
    \end{subfigure}% \vspace{10pt}
    \vspace{5pt}
    \begin{subfigure}[b]{\textwidth}
    \centering
      \frame{\includegraphics[width=0.9\linewidth]{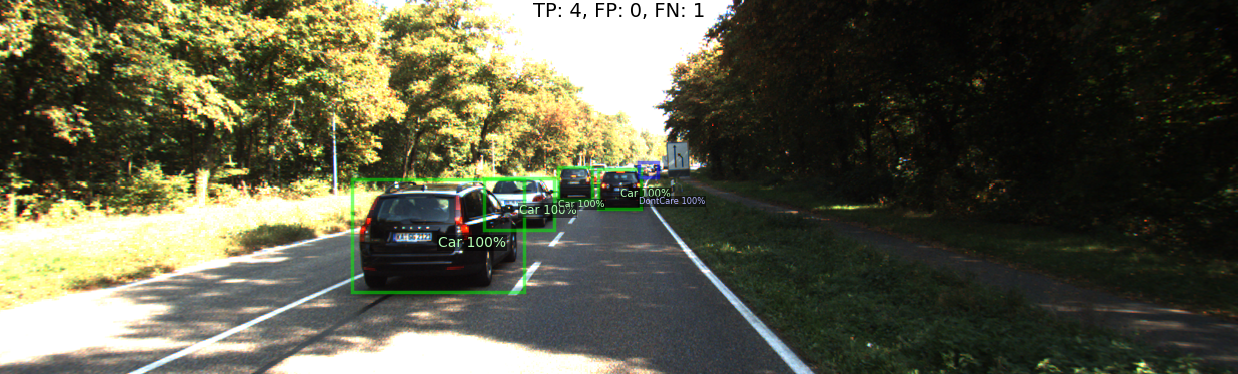}}
      %\caption{}
      \end{subfigure}
      \begin{subfigure}[b]{\textwidth}
      %\vspace{5pt}
      \centering
      \frame{\includegraphics[width=0.9\linewidth]{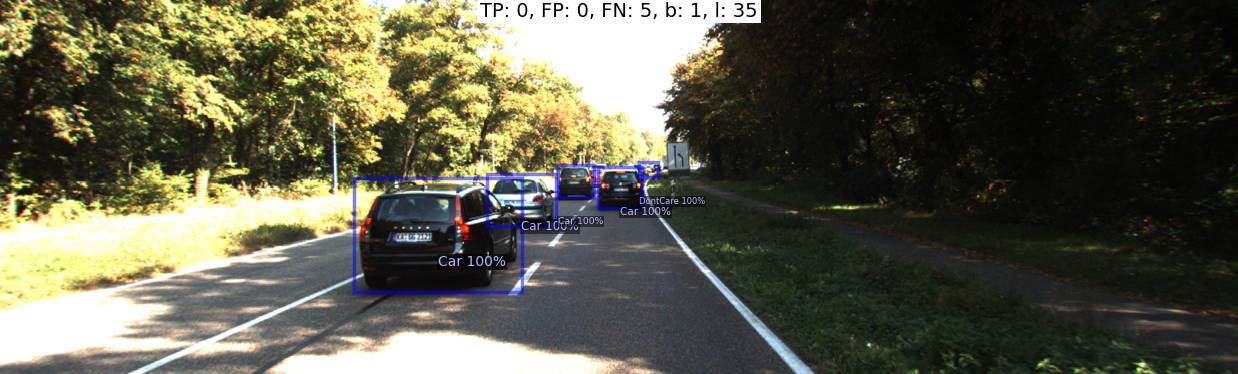}}
      \caption{Inference from Faster-RCNN model and Kitti dataset}
      \label{fig:example Faster-RCNN}
      \end{subfigure}
%\includegraphics[width=\linewidth]{examples/yolov3_kitti_767_orig.png} \\
%\vspace{.1cm}
%\includegraphics[width=\linewidth]{examples/yolov3_kitti_767_corr.png}
\vspace{-12pt}
\caption{Examples of the impact of a single neuron bit flip (at bit position $b$ and layer index $l$, see image insets). TPs are marked by green, FPs by red and FNs by blue rectangles, comparing the fault-free (top) and the faulty (bottom) predictions. In example (a) multiple FPs are generated right in front of the ego vehicle, while in (b) all previous detections  are erased due to the fault.}
\vspace{-8pt}
\label{fig:example}
\vspace{-8pt}
\end{figure}
\vspace{-15pt}
\section{Introduction}

Research communities seek to make the deployment of general artificial intelligence (AI) and deep neural networks (DNNs) used in everyday life as dependable as possible.
Significant emphasis is placed on handling corrupted input (e.g. due to visual artifacts or to attacks) provided to the model.
However, less effort has been dedicated to studying corruptions of the internal state of the model itself, most importantly caused by faults in the underlying hardware.
Such faults can occur naturally, such as memory corruption induced by external (e.g., cosmic neutron) radiation or electric leaking in the circuitry itself, typically manifested as bit flips or stuck-at-0/1s in the memory elements \cite{Athavale2020, Li2017}, which may alter the DNN model parameters (\emph{weight faults}) or the intermediate states (\emph{neuron faults}). %% TODO: Athavale2020
% \textcolor{blue}{They can also impact the input image, but in this work we focus only on the computation part of DNN}.
Platform faults can also impact the input while it is held in memory, yet this work focuses on the computational part of the DNN as our goal is to estimate the vulnerability of the model.
% DNNs are usually executed on hardware accelerators specifically designed to handle neural network-based workloads. %These accelerators already exceed the total required FIT rate of SOC without any protection \cite{Li2017}. 
%A DNN accelerator platform experiencing a fault can alter either the model weights (\emph{weight faults}) or the computational nodes (\emph{neuron faults}), leading to possible corruption of the DNN's prediction.
%
%are implemented at significant cost overhead, do not guarantee full protection and constitute significant cost overhead
The impact of these faults is often unpredictable in systems with large complexity.
Alterations can be of transient or permanent nature:
Transient faults have a short life span of the order of a few clock cycles and are therefore harder to detect by the system.
On the contrary, permanent faults may silently corrupt the system output for a longer period.
Memory protection techniques like error correcting code (ECC) can mitigate the risk of hardware faults \cite{Neale2016}; however, they are typically applied only to selected elements to avoid significant cost overheads. Given the rise in technology scaling with smaller node sizes and larger memory areas, future platforms are expected to become even more vulnerable to hardware faults \cite{Neale2016}.

%Both scenarios represent
%When remaining silent, a large  and go unnoticed for a longer time span until detected. \\

Object detection DNNs are among the most common examples of highly safety-critical DNN applications as they are in autonomous vehicles or in medical image analysis.
Typically, autonomous systems process events based on perception techniques. Hence, it is critically important that any potential hazards does not impact the system-level evaluation of events.
% These safety standards including SOTIF \cite{Sotif2019} require that the System on Chip (SoC) running the DNN algorithms provide high fault tolerance towards such soft errors (Failure-in-time rate (FIT \cite{Li2017}).
While the chances for a hardware fault to occur (for example, the chance of a neutron radiation event hitting a memory element) can be estimated statistically, it remains unclear how to quantify the safety-related impact of the failure of a DNN applied for the purpose of object detection. In contrast to simpler classification problems, the model output here typically consists of a multitude of bounding boxes and classes per image, of which a subset can be altered in the presence of a fault while others remain intact, see Fig. ~\ref{fig:example}. We find that commonly used average precision (AP) \cite{Microsoft2017} metrics inappropriately rely on the count of false objects irrespective of their interrelations (grouping in the same image or distributed across multiple frames). 
In real-time applications of DNNs, it further matters if the corrupted output is volatile or temporally stable across multiple input frames. The user is typically behind a tracking module that can regularize instantaneous alterations. We, therefore, see the need to establish a safety-related assessment of the vulnerability of object detection workloads under soft errors.
Depending on the specifications from safety assessment, we adopt a generalized notion of a safety hazard as a perturbation that causes a potentially unsafe decision by the end-user of the object detection module.

Therefore, we introduce two variants of the metric \text{\novelMetric } (Image-wise Vulnerability Metric for Object Detection), namely $\SDC$ in case of an image-wise silent data corruption (SDC), and $\DUE$ in case of detectable uncorrectable errors (DUE)s.
In this paper, we discuss the characteristics of the AP-based metrics in detail when used to quantify a model's vulnerability.
For example, AP50 is found to be hypersensitive to rare single corruption events compared to an evaluation at the image level. %We instead evaluate the chances of relevant object alteration at an image level. 
Our work supports maintaining the relationship of the system-level hazard evaluation to the impact of any hardware faults.
We find that a hardware fault - if it hits the crucial bits of either neuron or weight - can silently lead to excessive amounts of additional false positives (FPs) and increase the rate of false negatives (FNs) misses. 
%This leads to a silent data corruption event (SDC) as the system cannot recognize this easily.
%We show that these rare minor single bit flips move the bounding boxes, create class confusion or decrease the confidence scores.
We further study the impact of permanent faults in a real-time situation by considering continuous video sequences and observing a significant frequency that the error manifestation persists for a critical time interval. 
%We track occurrences of FP and FN over the entire video sequences to evaluate their impact in terms of corrupting the detection sequences. This, in turn, affects subsequent decision-making processes, ultimately leading to an unsafe scenario.

%\vspace{-8pt}
In summary, this paper makes the following contributions:
\vspace{-8pt}
\begin{itemize}
    \item We demonstrate that AP-based metrics lead to misleading vulnerability estimates for object detection DNN models (Sec. \ref{sec:ap})
    \item We propose an SCD-based/DUE-based metric \text{\novelMetric} to  quantify the vulnerability of object detection DNN models under hardware faults (Sec. \ref{sec: proposed metrics}). 
    \item We evaluate the vulnerability of various representative object detection DNN models using the proposed \text{\novelMetric}, illustrating the probability of a single bit flip resulting in a potentially safety-critical event (Sec. \ref{sec:error_probabilities}).
    \item For each such event, we propose various quantitative metrics to estimate the impact severity for typical safety-critical applications (Sec. \ref{sec:error_severities}).
    \item We extend our image-based evaluation to a video-based safety-critical system and measure the vulnerability of temporal persistency ($\Aoccfp$ and $\Aoccfn$) due to a permanent fault, by tracking the FPs and FNs across multiple video frames (Sec. \ref{sec:permanent_faults}).
		%\textbf{Severity of permanent faults:} We demonstrate the safety impact of the permanent faults on a continuous input of video stream in the automated car context by using a custom build pixel-wise tracker. 
\end{itemize}

\input{related_work}
\input{preliminaries}
\input{methodology}
\input{results_1}
\input{results_2}
\section{Conclusion}
\vspace{-5pt}
This work points out the challenges in estimating the vulnerability of object detection models under bit flip faults. Average precision-based metrics are either very sensitive or not sensitive to the corruption events, which can be misleading in a safety context.
For example, for F-RCNN+Kitti, neuron injections experiments showed almost no impact ($<0.1\%$) in the AP50 and mAP metrics. Using the image-based evaluation metric \text{\novelMetric} proposed here, however, we see that $0.7\%$ of all images lose substantial amounts ($>30\%$) of the total TP detections due to a single bit flip.
The evaluation method presented in this work allows us to come to a vulnerability estimate better addressing safety targets. Given the $\SDC$ probabilities and severities (see Fig. ~\ref{fig:fault_rates_1} and Tab.~\ref{tab:sev_features}), we conclude that the chances of safety-related corruptions due to soft errors are minor to moderate ($0.4\%-4.2\%$) in the studied setups. $\SDC$ events due to weight faults are about two times as likely as neuron faults. However, if SDC occurs, the severity can be grave. 
The \text{\novelMetric} metric should always be considered in combination with severity features for safety purposes. This is because \text{\novelMetric} does not quantify the severity, but only considers the existence of false and missed bounding boxes.
Our metric is defined relative to the original performance. This means that even if a fault also acts in a beneficial way, i.e. fixing some FP or FN occurrences, it will be categorized as a SDC here.
We estimated this severity with the help of different safety-related features. We observed that high bits of the exponent of floating point numbers, when hit by either neuron or weight faults, can lead to a significant increase in $\FPd$ and $\FNnd$. 
This effect is also translated into an average occupancy value that reflects the area portion of the image that is critically altered by a fault. We find that large average occupancies (up to $\Aoccfp \sim 81\%$ for FP and $\Aoccfn\sim86\%$ for FN) are common, reflecting significant safety hazards. Finally, we studied the use case of a sequential real-time image sequence from Lyft to show that permanent \textit{stuck-at} faults on neurons or at weights can induce FP objects covering as much as $\sim83\%$ of the image area, creating dangerous ghost objects. Similarly, up to $\sim63\%$ of the TP area in the scene can be missed. Overall, the weight faults are more likely impactful than neuron faults and have a higher severity in area occupancy (except for permanent FNs).

%the criticality of transient faults is debatable as the final FIT rate derived from several SDCs is chip-dependent. Whereas, Yolov3 trained on the Lyft dataset has a high severity condition showing that the model is most vulnerable against permanent faults. A detailed study
%of other obj-det models and datasets with continuous video sequences is required to generally conclude on the model's behavior in the presence of permanent faults.
\vspace{-10pt}
\section*{Acknowledgment}
\vspace{-10pt}
This project has received funding from the European Union's Horizon $2020$ research and innovation programme under grant agreement No $956123$.
%This work was partially funded by the German Federal Ministry of Transport, Building and Urban Development (BMVI) within the projects KoRA9 (grant No. 16AVF1032A) and Providentia++ (grant No. 01MM19008).
Our research was partially funded by the Federal Ministry of Transport and Digital Infrastructure of Germany in the project Providentia++ (01MM19008).

This version of the contribution has been accepted for publication, after peer review (when applicable) but is not the Version of Record and does not reflect post-acceptance improvements, or any corrections. The Version of Record is available online at: \href{https://doi.org/10.1007/978-3-031-14835-4_20}{{https://doi.org/10.1007/978-3-031-14835-4\_20}}. Use of this Accepted Version is subject to the publisher\textquotesingle s Accepted Manuscript terms of use \href{https://www.springernature.com/gp/open-research/policies/accepted-manuscript-terms}{{https://www.springernature.com/gp/open-research/policies/accepted-manuscript-terms}}. 
\vspace{-10pt}
%The preferred spelling of the word ``acknowledgment'' in America is without 
%an ``e'' after the ``g''. Avoid the stilted expression ``one of us (R. B. 
%G.) thanks $\ldots$''. Instead, try ``R. B. G. thanks$\ldots$''. Put sponsor 
%acknowledgments in the unnumbered footnote on the first page.

%\section*{References}
% \bibliographystyle{IEEEtran}
\bibliographystyle{splncs04}
\bibliography{faults_that_matter}

\end{document}

%% file: abbr.tex
\newcommand{\statzero}{$\text{stuck-at-0}$}
\newcommand{\statone}{$\text{stuck-at-1}$}

\newcommand{\Aoccfporig}{A_{\text{FP}_{\text{ref\_blob}}}}
\newcommand{\Aoccfnorig}{A_{\text{FN}_{\text{ref\_blob}}}}

\newcommand{\Aoccfp}{A_{\text{FP}_{\text{blob}}}}
\newcommand{\Aoccfn}{A_{\text{FN}_{\text{blob}}}}

\newcommand{\IoUeval}{\text{IoU}_{\text{eval}}}
\newcommand{\bitavg}{\text{bitavg}}
\newcommand{\avg}{\text{avg}}
\newcommand{\FPd}{{\Delta FP}_{\text{}}}
\newcommand{\FNnd}{{\Delta FN}_{\text{n}}}
\newcommand{\px}{\text{px}}
\newcommand{\novelMetric}{\text{IVMOD}}
\newcommand{\novelMetricBF}{\textbf{\novelMetric}}
\newcommand{\SDC}{\text{\novelMetric}_\text{SDC}}
\newcommand{\DUE}{\text{\novelMetric}_\text{DUE}}

\newlength{\tempdima}
\newcommand{\rowname}[1]% #1 = text
{\rotatebox{90}{\makebox[\tempdima][c]{\textbf{#1}}}}
\newcommand{\mycaption}[1]% #1 = caption

%% file: related_work.tex
\section{Related Work}
The effort to estimate the vulnerability or resilience of the DNNs against hardware faults affecting the model has been explored recently to study the safety criticality of a model when used in real-time operation. 

To this extent, faults are injected in DNNs during inference either at
the application layer on weights/neurons (\cite{Li2017, GeisslerQRAPDGP21}), or by neutron beam experiments ( (\cite{DosSantos2019, Hou2020}), black-box techniques).
Authors of Ref. \cite{Li2017, Beyer2020} considered transient faults, which are multiple event upsets occurring in data or buffers of DNN accelerators. Many prior works claimed DNNs to have inherent tolerance towards faults. Li. et al. (\cite{Beyer2020}) studied the vulnerability of DNNs by injecting faults in data paths and buffers with different data type levels and quantified it in the form of SDC probabilities and FIT (failure in time) rates.
It is seen that errors in buffers propagate to multiple locations compared to errors in data-path.
These works estimated the resiliency of the model by injecting multiple fault injections during the feed-forward inference. This analysis is limited to image classification models like AlexNet \cite{Krizhevsky2012}, VGG \cite{Simonyan2015}, and ResNets \cite{he2016deep}. Our analysis does not characterize the faults in buffer and faults in the data-path. We assume the faults will propagate to the application layer, which may impact either the weights or the neurons. Hence we analyze them independently, assuming equal probabilities.
The Ares framework \cite{Reagen2018} demonstrated that activations (neurons) in image classification networks are 50x more resilient than weights. These works focus mainly on fault models involving multiple bit flips captured by bit error rate (BER).
There is limited research done on understanding the vulnerability of object detection DNNs. The work in Ref. \cite{DosSantos2019}  quantified the architectural vulnerability factor (AVF) of Yolov3 using metrics like SDC AVF, DUE AVF, and FIT rates.
This work studies fault propagation by injecting a random value in the selected register file and not flipping a bit. The authors argue that not all SDCs are critical, given that change in objects' confidence scores after injecting faults is tolerable. The definition of SDC used in this work is not straightforward. They use the precision and recall values computed at the object level by combining all the images, obscuring the actual vulnerability.
The vulnerability of object detection DNNs is studied by injecting faults using neutron beam \cite{Lotfi2019}.
The authors analyzed both transient and permanent faults but not on continuous video sequences. Also, the dataset considered in these experiments was primarily limited to only one object per image. Also, they injected faults into the input image. We limit our fault injections to neurons and weights and only to convolution layers of the DNNs as the fully connected layers did not change much of the observed data. We believe fault-injected images do not fall into the category of the model vulnerability. They rather  find their place in adversarial input space within various adverse fault/noise models. The results obtained from many of these works are not easy to compare as the failure and SDC definitions differ and do not follow standard baseline. To our best knowledge, our paper is the first to demonstrate vulnerabilities of the object detection models in detail using the proposed (\text{\novelMetricBF}) metrics to measure the severities at the image level.
Also, we introduce a new metrics $\Aoccfp$ and $\Aoccfn$ quantifying the area occupancy of FP/FN blobs, which is essential to establish the safety criticality of the object detection models concerning specific real-time applications.
% We analyse the model's vulnerability using realistic assumptions of failure conditions (SDCs) and study the severities (TPs/FPs/FNs) in detail. To our best knowledge, our paper is the first one to demonstrate vulnerabilities of the object detection models in detail using a proposed metric , including the severities arising due to individual components like FPs and FNs considering both transient and permanent faults. Also, we introduce a new metric - area occupancy by FP/FN blob, which is essential to establish the vulnerability of the object detection models. We propose more realistic methodologies and redesigned metrics to evaluate the model sensitivity for transient and permanent faults.
\vspace{-10pt}

%% file: preliminaries.tex
\section{Preliminaries}
\subsection{Hardware faults vocabulary}
\label{sec:hardware faults vocabulary}
Our fault injection technique includes transient and permanent faults.
Transient faults refer to random bit flips (0$\rightarrow $1 or 1$\rightarrow $0) of a randomly chosen bit, which occur during a single image inference and are removed afterward.
Permanent faults are modeled as \text{\statzero} and \text{\statone} errors, meaning that a bit remains consistently in state '0' or '1' without reacting on intended updates. Those faults are assumed to persist across many image inferences.
%Fault models injected into the models for this work are transient and permanent faults.
%memory element remains consistently in state '0' or '1' without acting on intended updates.
%Permanent faults are further classified into \text{\statzero} and \text{\statone} faults.  In this case the memory element remains consistently in state '0' or '1' without acting on intended updates.
We inject faults either into intermediate computational states of the network (neurons) or into the parameters (weights) of the DNN model, focusing only on convolutional layers, which constitutes a significant part of all operations in the studied DNNs.
Both types of faults represent bit flip in the respective memory elements, holding either temporary states such as intermediate network layer outputs or learned and statically stored network parameters.
%, while weight faults affect those memory elements that store the learned network parameters.
A fault can potentially induce critical alterations of the model predictions, measured by $\SDC$ or $\DUE$ as shown in Eq.~\ref{eq:SDC_DUE}.
%\begin{itemize}
    %\item \textbf{Transient faults:}  This fault is modeled by flipping (0$\rightarrow$1 or 1$\rightarrow$0) a randomly chosen bit.
    %All transient faults in this work is restricted to a single inference of an image.
    %\item \textbf{Permanent fault - \text{\statzero}:} This fault is modeled by changing the state of a randomly chosen bit to 0 (X$\rightarrow$0) irrespective of it initial state.
    %This fault injection is constant during an inference of an entire epoch of the images.
    %\item \textbf{Permanent fault - \text{\statone}:} This fault is modeled by changing the state of a randomly chosen bit to 1 (X$\rightarrow$1) irrespective of it initial state.
    %This fault injection is constant during an inference of an entire epoch of the images.
%\end{itemize}

\vspace{-10pt}
\subsection{Experimental setup: Models, datasets and system}
We use standard object detection models - Yolov3 \cite{Redmon2018}, RetinaNet \cite{Lin2020}, Faster-RCNN (F-RCNN\cite{Ren2017}) - together with the test datasets CoCo2017 \cite{Microsoft2017}, Kitti \cite{Geiger2013a} and Lyft \cite{Lyft}.
We retrained Yolov3 on the Kitti and Lyft dataset, and the Faster-RCNN model on Kitti for comparative experiments. 
We used open-source trained weights for the rest of the models and datasets. The base performances of these models in terms of AP50 and mAP can be found in Fig. \ref{fig:fault_rates_1}. The parameter configurations used for these models (NMS threshold, confidence score, etc.) are taken from the original publications. Since fault injection is compute-intensive, we select a subset of $1000$ images for each dataset to perform the transient fault analysis and use a single Lyft sequence of $126$ images for the permanent fault analysis. All experiments adopt a single-precision floating-point format (FP32) according to the IEEE754 standard \cite{IEEE2019}. Our conclusions also apply to other floating-point formats with the same number of exponent bits, such as BF16 \cite{Intel2018}, since no relevant effect was observed from fault injections in mantissa bits. 
% All the experiments were performed using Intel\textsuperscript{\textregistered} Core\textsuperscript{TM} i9 CPUs, GeForce RTX 2080, Titan RTX, and RTX 3090 GPUs and PyTorch (version 1.8.0).
\vspace{-10pt}
% Table of parameters
% \begin{wraptable}{r}{2cm}
%     % \centering
%     \begin{tabular}{ccccc} \\\toprule
%                     Setup & NMS IoU & NMS Conf thres & Dataset & AP50\\
%                     \midrule
%         Yolov3 & 0.45 & 0.25 & Coco2017, Kitti, Lyft &  32.1, 87.6, 85.2 \\
%                     %Yolo+Kitti & 0.45 & 0.25 & 0.5 & 1K \\
%                     RetinaNet & 0.5  & 0.05 & Coco2017 & 49.7 \\
%                     Faster-RCNN & 0.5 & 0.05 & Coco2017, Kitti & 19.4, 51.6 \\
%                     \bottomrule
%     \end{tabular}
%             \caption{Experimental setup}
% \label{tab:models_datasets}
% \end{wraptable}

% \begin{table}[!h]
%         \centering
%         \begin{tabular}{p{15mm}p{7mm}p{7mm}p{20mm}p{15mm}} %{ccccc}
%                         \toprule
%                         Setup & NMS IoU & NMS Conf thres & Dataset & AP50\\
%                         \midrule
%             Yolov3 & 0.45 & 0.25 & Coco2017, Kitti, Lyft &  32.1, 87.6, 85.2 \\
%                         %Yolo+Kitti & 0.45 & 0.25 & 0.5 & 1K \\
%                         RetinaNet & 0.5  & 0.05 & Coco2017 & 49.7 \\
%                         Faster-RCNN & 0.5 & 0.05 & Coco2017, Kitti & 19.4, 51.6 \\
%                         \bottomrule
%         \end{tabular}
%                 \caption{Experimental setup}
% \label{tab:models_datasets}
% \end{table}

%% file: methodology.tex
\section{Methodology of vulnerability estimation}
\subsection{Issues with average precision}
\label{sec:ap}

\begin{figure*}[!ht]
    \centering
    
    \begin{subfigure}[b]{0.24\textwidth}
    \includegraphics[width=\textwidth]{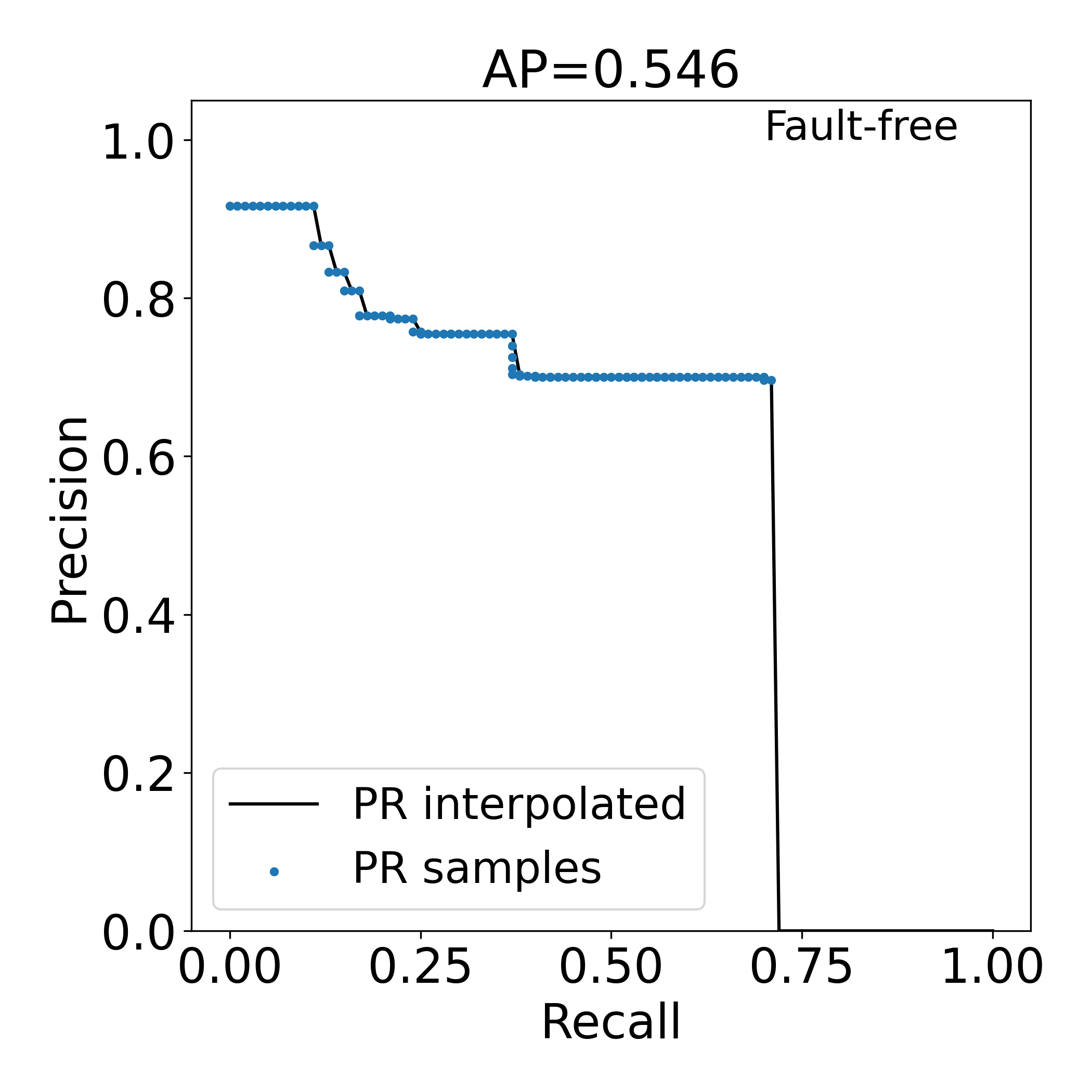}
    \vspace{-3pt}
    \caption{}
    \end{subfigure}
    \begin{subfigure}[b]{0.24\textwidth}
    \includegraphics[width=\textwidth]{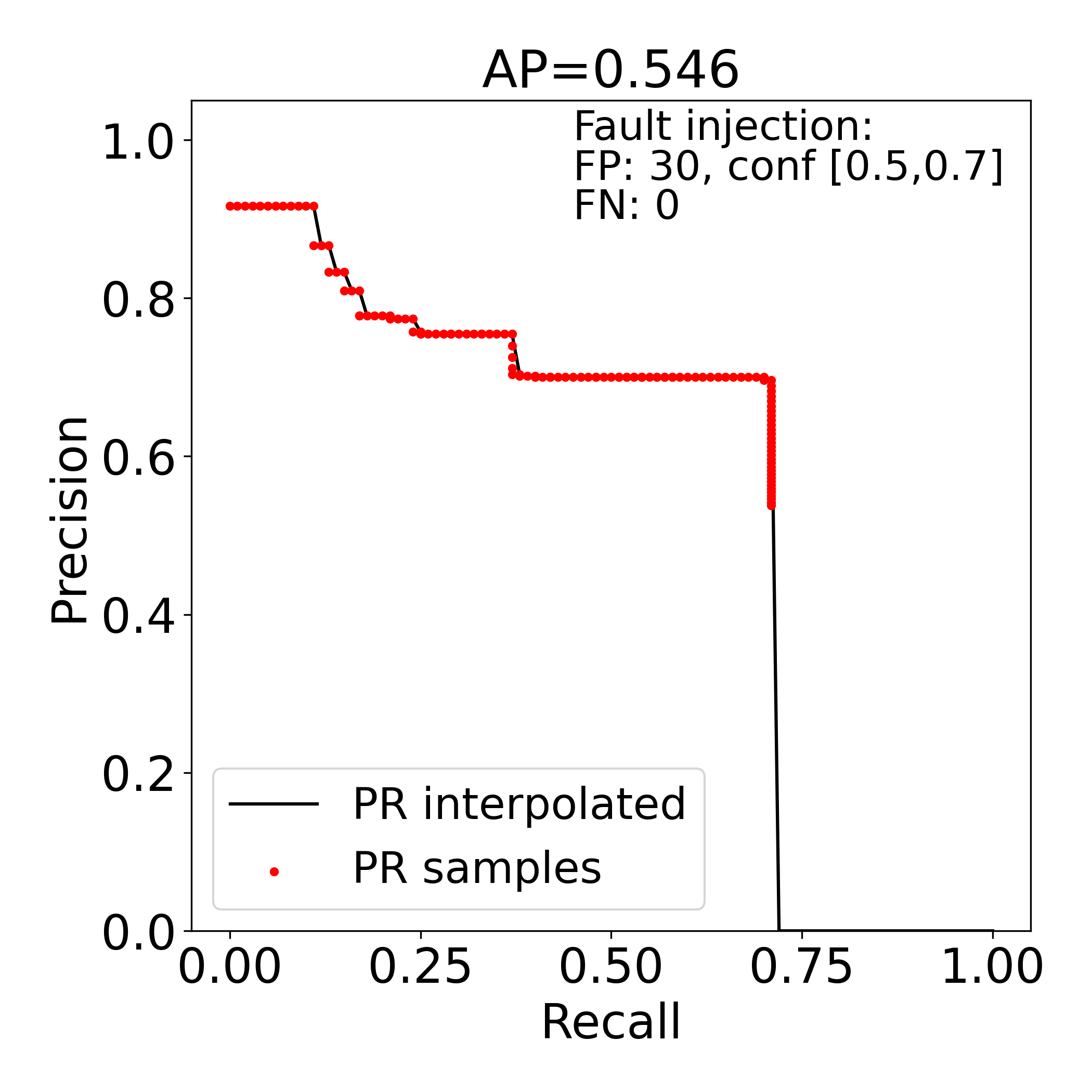}
    \vspace{-3pt}
    \caption{}
    \end{subfigure}
    \begin{subfigure}[b]{0.24\textwidth}
    \includegraphics[width=\textwidth]{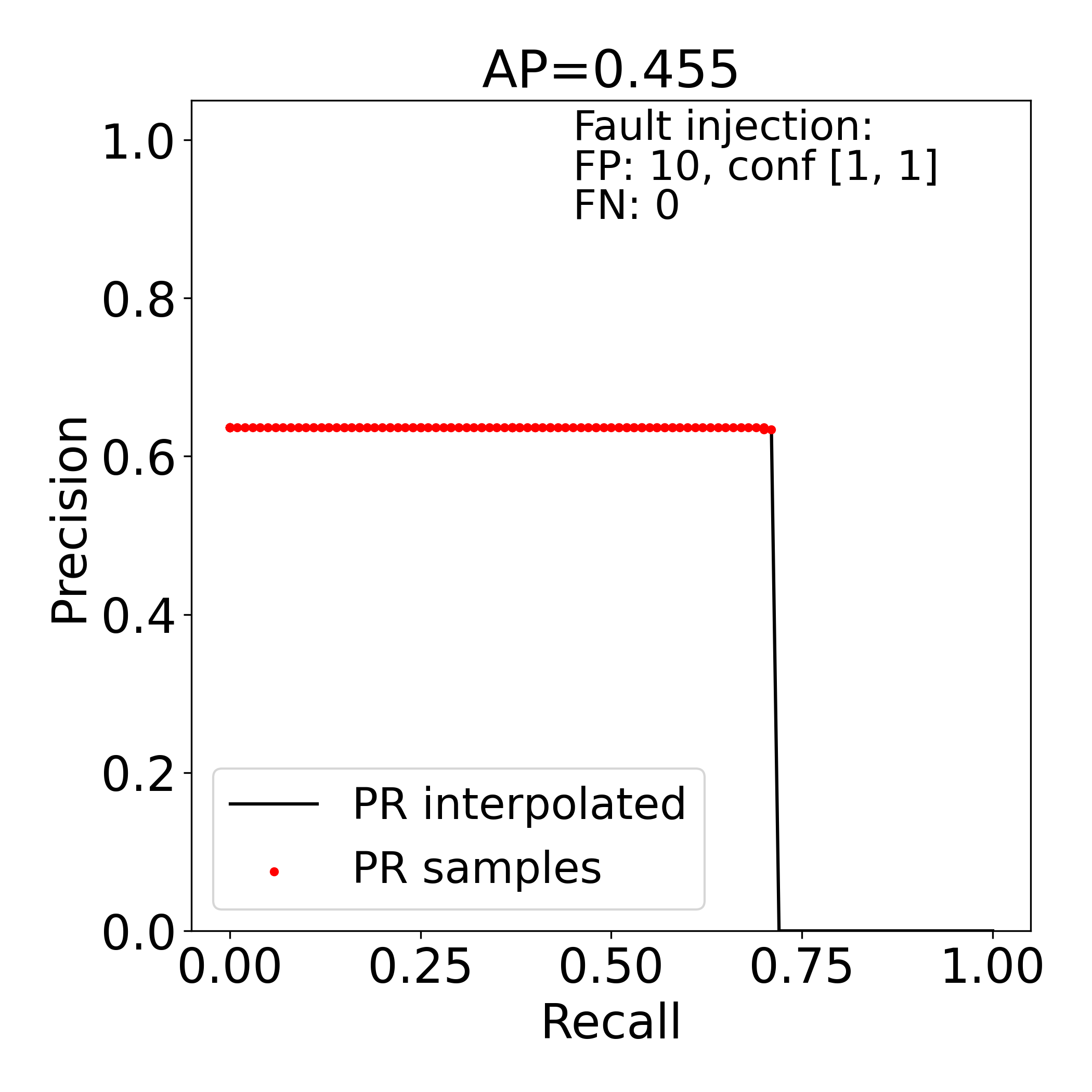}
    \vspace{-3pt}
    \caption{}
    \end{subfigure}
    \begin{subfigure}[b]{0.24\textwidth}
    \includegraphics[width=\textwidth]{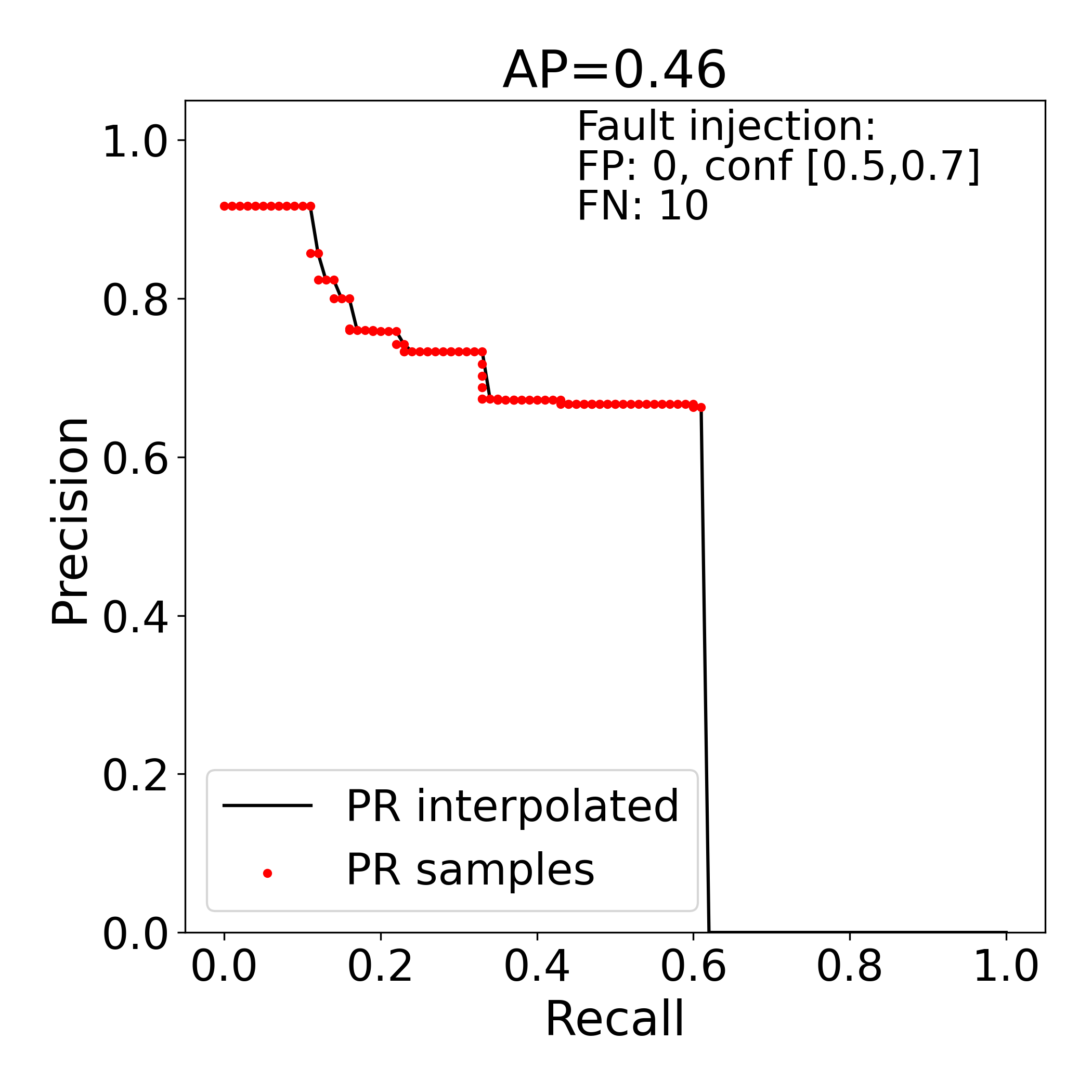}
    \vspace{-3pt}
    \caption{}
    \end{subfigure}
    \vspace{-5pt}
    \caption{Simulation of the effect of fault injection on the AP metric. Here, an artificial data set of $100$ objects was generated, where each object was classified as TP with a chance of $0.7$ or as a FN otherwise. In addition, FPs were created with a rate of $0.3$ per true detection. Both TPs and FPs are assigned random confidence values between $0.7$ and $1$. To this setup (a), additional FPs simulating the effect of fault injection were augmented or existing TPs were randomly eliminated to model fault-induced FNs ((b)-(d)). The diagrams show the PR curves and the effect of fault injection on them. Number and confidence range of the faulty objects are given in the insets.}
		\vspace{-8pt}
    \label{fig:pr_example}
    \vspace{-8pt}
\end{figure*}

%
%\begin{figure*}[!ht]
    %\centering
		%\begin{subfigure}
		%\end{subfigure}
    %\includegraphics[width=\linewidth]{metrics/pr_curves.png}\\
		%(a) \hspace{4cm}(b) \hspace{4cm}(c) \hspace{4cm}(d)
    %\caption{Illustration of the effect of fault injection on the AP metric. Here, an artificial data set of $100$ objects was generated, where each object was classified as TP with a chance of $0.7$ or became a FN otherwise. In addition, FPs were created with a rate of $0.3$ per true detection. Both TPs and FPs are assigned random confidence values between $0.7$ and $1$. To this setup (a), additional FPs simulating the effect of fault injection were augmented or existing TPs were randomly eliminated to model fault-induced FNs ((b)-(d)). The diagrams show the PR curves and the effect of fault injection on them. Number and confidence range of the faulty objects are given in the insets.}
    %\label{fig:pr_example}
%\end{figure*}
In object detection, evaluation and benchmarking methods are most commonly selected from the family of average precision (AP)-based metrics (in combination with specific IoU thresholds such as AP50 or mAP).
Libraries such as CoCo API \cite{Microsoft2017} perform the following relevant steps to obtain AP values from a set of object predictions:
%To evaluate the performance of object detection DNNs after injecting the hardware faults, we use the basic approach of CoCo-API except the assignment policy defined by us differs slightly from the typical strategy used by benchmarking tools such as CoCo API for the calculation of AP metrics. 
%In particular, the following steps are taken to obtain an AP value: 
i) ground truth and the predicted objects are collected in groups of the same class label, 
ii) within a group, the predicted objects are sorted w.r.t their confidence scores, 
iii) the sorted predictions are consecutively assigned to the ground truth objects within the same class group, using an appropriate IoU threshold, 
iv) precision and recall (PR) curves are evaluated sequentially through the confidence-ranked TP, FP, and FN objects,
v) the class-wise AP is calculated as the area under the interpolated PR curve of a class, and vi) the overall AP is determined as the average of the class-wise AP values. 

It has been pointed out that such AP metrics can lead to non-intuitive results in the detection performance of a model on a specific data set \cite{Redmon2018}. %\cite{Hall2020}
In the following, we illustrate that an AP-based evaluation can be misleading when estimating the vulnerability of a model against corruption events such as soft errors in a safety-critical real-time context concerning the probability and severity of corruption.
Corruption events lead to additional FP and FN objects merged into or eliminated from the healthy list of detected objects.  
We identified the following issues when trying to quantify model vulnerability based on AP metrics: 
\begin{itemize}
\item \textbf{Object-level evaluation:} The AP is calculated on an object level, i.e., the amount of TP, FP, FN objects accumulated across all images is used for evaluation. This does not consider how corrupted boxes are distributed across images, i.e., one image with a large number of fault-induced FP detections can have the same effect as many corrupted images with few FP detections each. From a real-time safety perspective, however, the amount of corrupted image frames is typically relevant, as this may determine, for example, the robustness of a video stream used for environment perception.
\item \textbf{Dependency of PR on confidence:} Due to the sequential and integration-based characteristic of the average precision, the fault-induced FP object's impact depends highly on those sample's confidence. This does not reflect the potential safety relevance a low-confidence FP object may have, see more below. 
\item \textbf{Dependency of box assignment on confidence:} The strict confidence ranking can, in some cases, lead to a non-optimal global assignment of bounding boxes. For example, a better matching box might have slightly lower confidence than a global optimization would demand.
\item \textbf{Class-wise average:} Common and rare classes have the same weight in the overall AP metric. However, their detection performance and vulnerability can be quite different as they typically relate to the samples the model encountered during training.
%\item \textbf{No separation of abnormal data samples:} FP or FN samples that show artifacts like \textit{Inf} and \textit{NaN} values are not handled separately which ilent and non-silent faults with different criticality.
\end{itemize}
\vspace{-6pt}
In particular, the second point above is non-intuitive; we therefore illustrate this in more detail in Fig. ~\ref{fig:pr_example} with the help of a generic example from a randomly generated data set of $100$ objects.
Additional FPs with low confidence compared to the reference set of objects have a negligible impact on the metric as they get appended to the tail of the PR curve, even when numerous and potentially safety-critical. On the contrary, few high-confidence FP objects can lead to significant drops in the AP as those samples get sorted in at the head of the PR curve to lower it. Fault-induced FNs reduce the area under the PR curve by pushing the samples towards smaller recalls.
\subsection{Proposed metrics: \text{\novelMetric}}
\label{sec: proposed metrics}
We introduce \text{\novelMetric} metrics to measure the image-wise vulnerability of the object detection DNNs. 
Our evaluation strategy described in the following seeks to counter the issues with AP-based metrics described in the last section in order to reflect vulnerability estimation better addressing safety targets.
In particular, our approach is characterized by: 
\begin{itemize}
\item \textbf{Image-level evaluation:} We evaluate vulnerability on an image level instead of an object level. 
This approach reflects that those faults jeopardize safety applications that silently alter the free and occupied space by inducing false detections in an image, particularly sequences thereof, even if such an alteration involves only few false objects per frame. 
%\item \textbf{Handling silent and non-silent corruption:} 
We register image-wise SDC and DUE events, see Sec.~\ref{sec:sdc_due}, to determine the probability of a relevant fault impact. The severity of the latter is evaluated separately in terms of the amount of induced FPs and FNs. Due to their image-based character, $\SDC$ and $\DUE$ metrics are naturally independent of the object confidences.
\item \textbf{Confidence-independent box assignment:} False-positive objects can be critical whether they have high or low confidence, which is masked in the AP metric. We apply a different assignment scheme for FPs and FNs that omits confidence ranking and hence makes the model vulnerability metric independent of the confidence of FPs, see Sec.~\ref{sec:assignment_strat}. The assignment strategy can also be varied to relax class correspondence requirements, which are often overemphasized from a safety perspective. The system can perform at degraded level if its sure of object location and not much about the class.
\item \textbf{Class-independent average:} We evaluate the overall sample mean instead of the mean of individual class categories to reflect typical imbalances in the data set concerning object classes.
\end{itemize}
%With our assignment, we address two potential issues that can be important for a practical, possibly safety-critical application: 
%i) We therefore omit the predicted confidence in the assignment process. 
%ii) The relevance of a strict class matching and subsequently a class-specific evaluation might be overemphasized for many applications. As an example, we observe that Yolov3 has difficulties in distinguishing the classes \textit{car}, \textit{truck}, and \textit{Van} of Kitti, however, for practical purposes this is often irrelevant as those objects behave in a very similar way and only the free or occupied space in the driving path matters. Our assignment strategy is therefore foremost based on the IoU, while the class correspondence can be varied for testing purposes. 
%iii) Further, we evaluate the overall sample mean instead of the mean of individual class categories.
\vspace{-15pt}
\subsubsection{Assignment policy} \label{sec:assignment_strat}
In contrast to the sequential and class-wise matching described in Sec.~\ref{sec:ap}, we calculate the cost matrix from a set of predictions and ground truth objects for a single image. The cost for matching objects is the IoU between the bounding boxes.
If the IoU is below the specified threshold $\IoUeval=0.5$, or if the classes of the two objects are not the same, we assign a maximum cost. To analyze the relevance of exact class predictions for an application, we can harden or soften the class matching from a one-to-one correspondence to compatible class clusters or neglect class matching altogether. 
A Hungarian association algorithm \cite{Kuhn1955} is then deployed to obtain the global optimal cost assignment.
As usual, the number of accepted matches per image represents the true positive (TP) cases.
%, while the amount of unmatched boxes from the ground truth and the prediction set defines the number of false negatives (FN) and false positives (FP), respectively. 
False detections are registered in the following cases: 
i) a FP and a simultaneous FN detection occurs if the IoU with the assigned ground truth object is below the threshold, independent of the predicted class, or if the IoU is sufficiently large, but the classes are not compatible, 
ii) a single FP occurs if there is a predicted object that cannot be assigned to any ground truth object with acceptable costs, 
iii) a single FN occurs if there is a non-assigned ground truth object. Fig. ~\ref{fig:example} shows an illustrative example of assigned TP, FP, FN boxes. In our setup, we clip predicted bounding boxes reaching out of the image dimensions -- e.g., due to faults -- to the actual image boundaries.
%From now on we would refer original CoCo-API as \textit{Metric-DP} used for evaluating detectors performance and the above changed metric as \textit{Metric-DV} used for evaluating detectors vulnerability.
\vspace{-15pt}
\subsubsection{\text{\novelMetric} ($\SDC$ and $\DUE$)} \label{sec:sdc_due}
%We study the impact of transient and permanent faults by injecting them either to neuron or weights of the model during inference at application layer. 
%All the study discussed in this work is limited to 1 fault per inference. 
%Our approach is guided by the idea that those faults will be most safety-critical that alter the free and occupied space in the image by inducing false positive or false negative detections. 
%After injecting faults for each image, depending on the injection policy chosen, the vulnerability of the model is computed using the metrics \textit{SDC rate} and \textit{DUE}.
We define the $\SDC$ rate as the ratio of events where a fault during inference causes a silent corruption of an image and the total number of image inferences. 
$\SDC$ is an SDC defined as a change in either of the TP, FP, or FN count of the respective image, compared to the original fault-free prediction, given that no irregular \textit{NaN} (not a number) or \textit{Inf} (infinite) values occur during the inference as shown in  Eq. \ref{eq:SDC_DUE}. Since TPs and FNs are complementary to each other, we can eliminate either TP or FN in $\SDC$ in Eq. \ref{eq:SDC_DUE}.
On the other hand, the $\DUE$ rate is the ratio of events where irregular \textit{NaN} or \textit{Inf} values are generated during inference and detected inside the layers or in the predicted output due to the injected fault in the respective image during inference and are computed using the Eq. \ref{eq:SDC_DUE}. As DUE events are naturally detectable, they typically are less critical than SDC events. Explicitly,

%\begin{equation}
\vspace{-15pt}
\begin{align}
\vspace{-25pt}
\begin{split}
        %SDC = \frac{1}{N} \sum^{N}_{i=1} \Big[(|TP_{\text{corr[i]}}- TP_{\text{orig[i]}}| \geq 1) \\
              %|\quad (|FP_{\text{corr[i]}}- FP_{\text{orig[i]}}| \geq 1) \\
              %|\quad (|FN_{\text{corr[i]}}- FN_{\text{orig[i]}}| \geq 1)\Big]
            %   SDC = \frac{1}{N} \Big[\sum^{N}_{i=1}(|TP_{\text{corr[i]}}- TP_{\text{orig[i]}}| \geq 1) \\
            %   |\quad \sum^{N}_{i=1}(|FP_{\text{corr[i]}}- FP_{\text{orig[i]}}| \geq 1) \\
            %   |\quad \sum^{N}_{i=1}(|FN_{\text{corr[i]}}- FN_{\text{orig[i]}}| \geq 1)\Big]
            \SDC &= \frac{1}{N} \sum^{N}_{i=1} \big\{\big[ (FP_{\text{orig}})_i \neq (FP_{\text{corr}})_i \lor\ (FN_{\text{orig}})_i \neq (FN_{\text{corr}})_i\big]
             \land\ \lnot \text{Inf}_i\ \land\ \lnot \text{NaN} \big\}, \\
%%\end{split}
%\label{eq:SDC}
%%\end{align}
%\end{equation}
%\begin{equation}
    %\begin{split}
        %DUE = \frac{1}{N} \sum^{N}_{i=1} \Big[(Detected\_\textit{NaN}_{\text{corr[i]}} \geq 1) \\
              %|\quad (Detected\_\textit{INF}_{\text{corr[i]}} \geq 1)\Big] \\
        \DUE &= \frac{1}{N} \sum^{N}_{i=1} \left[ \text{Inf}_i\ \lor\ \text{NaN}_i \right].
\end{split}
\vspace{-25pt}
\label{eq:SDC_DUE}
\end{align}
\vspace{-15pt}
%\end{equation}

%% file: results_1.tex
\begin{figure*}[!ht]
    \centering
    
    \begin{subfigure}[b]{.32\linewidth}
    \includegraphics[width=\linewidth]{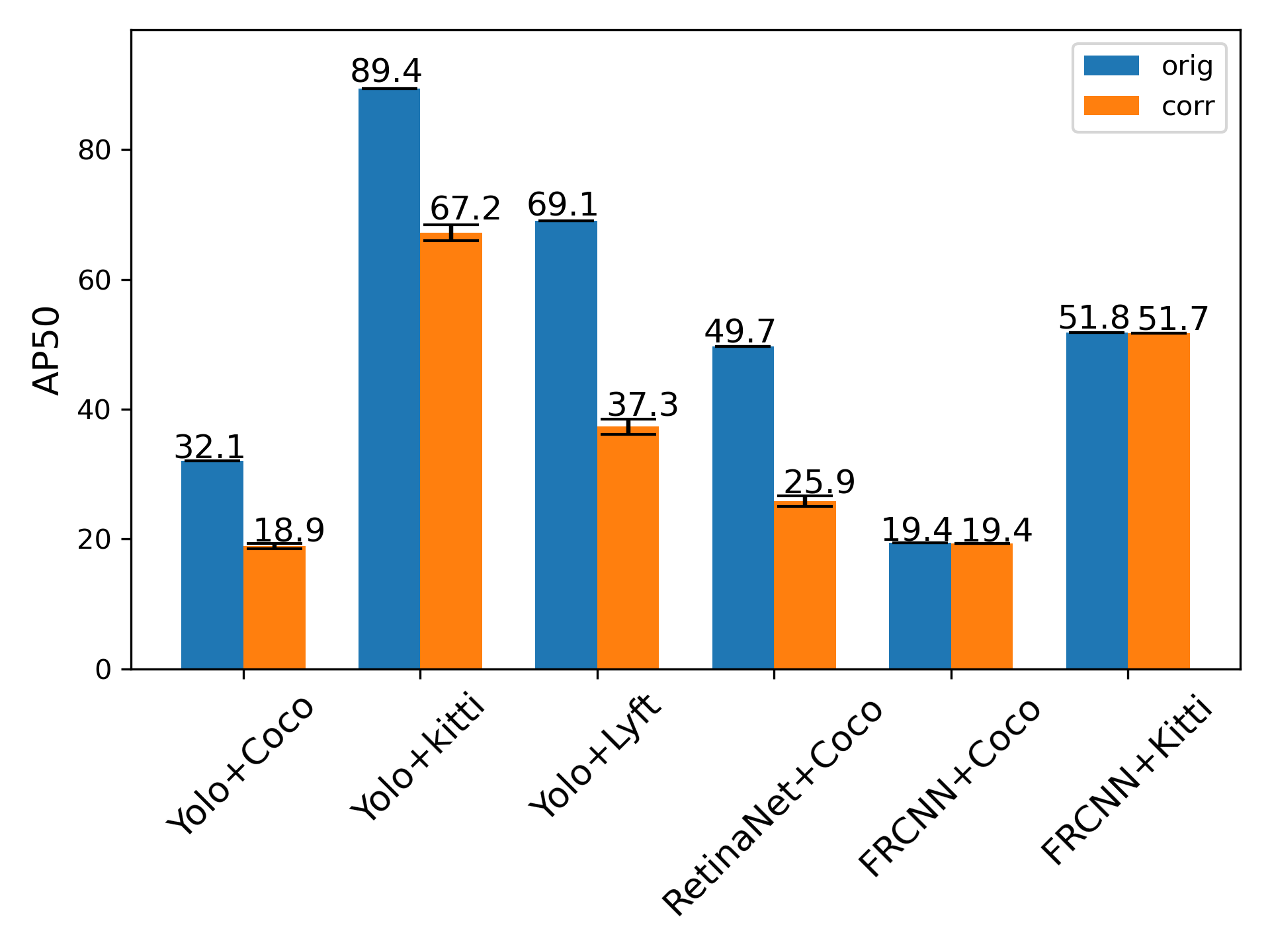}
    %\caption{A mouse}\label{fig:mouse}
    \end{subfigure}
    \begin{subfigure}[b]{.32\linewidth}
    \includegraphics[width=\linewidth]{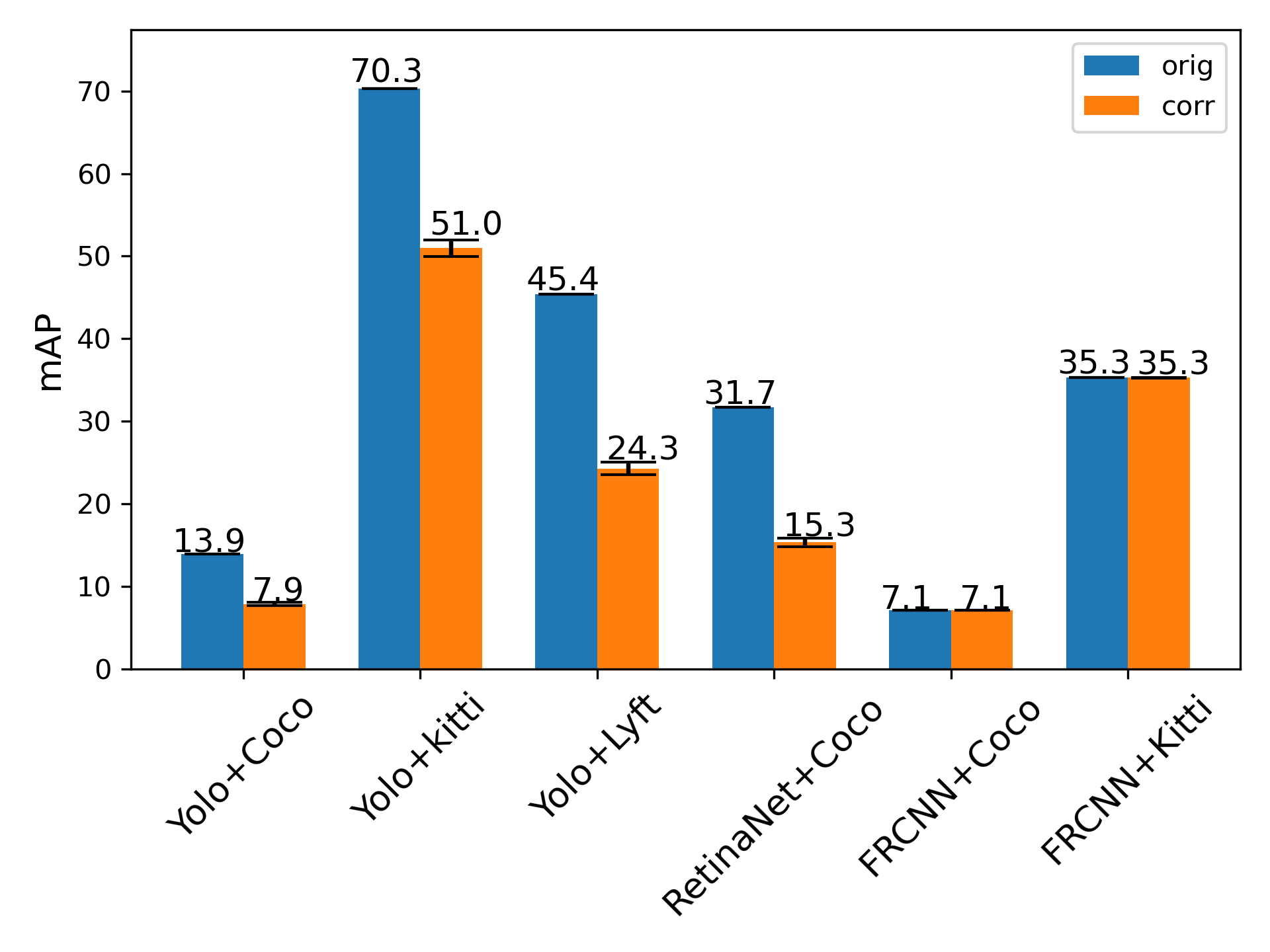}
    %\caption{A mouse}\label{fig:mouse}
    \end{subfigure}
    \begin{subfigure}[b]{.32\linewidth}
    \includegraphics[width=\linewidth]{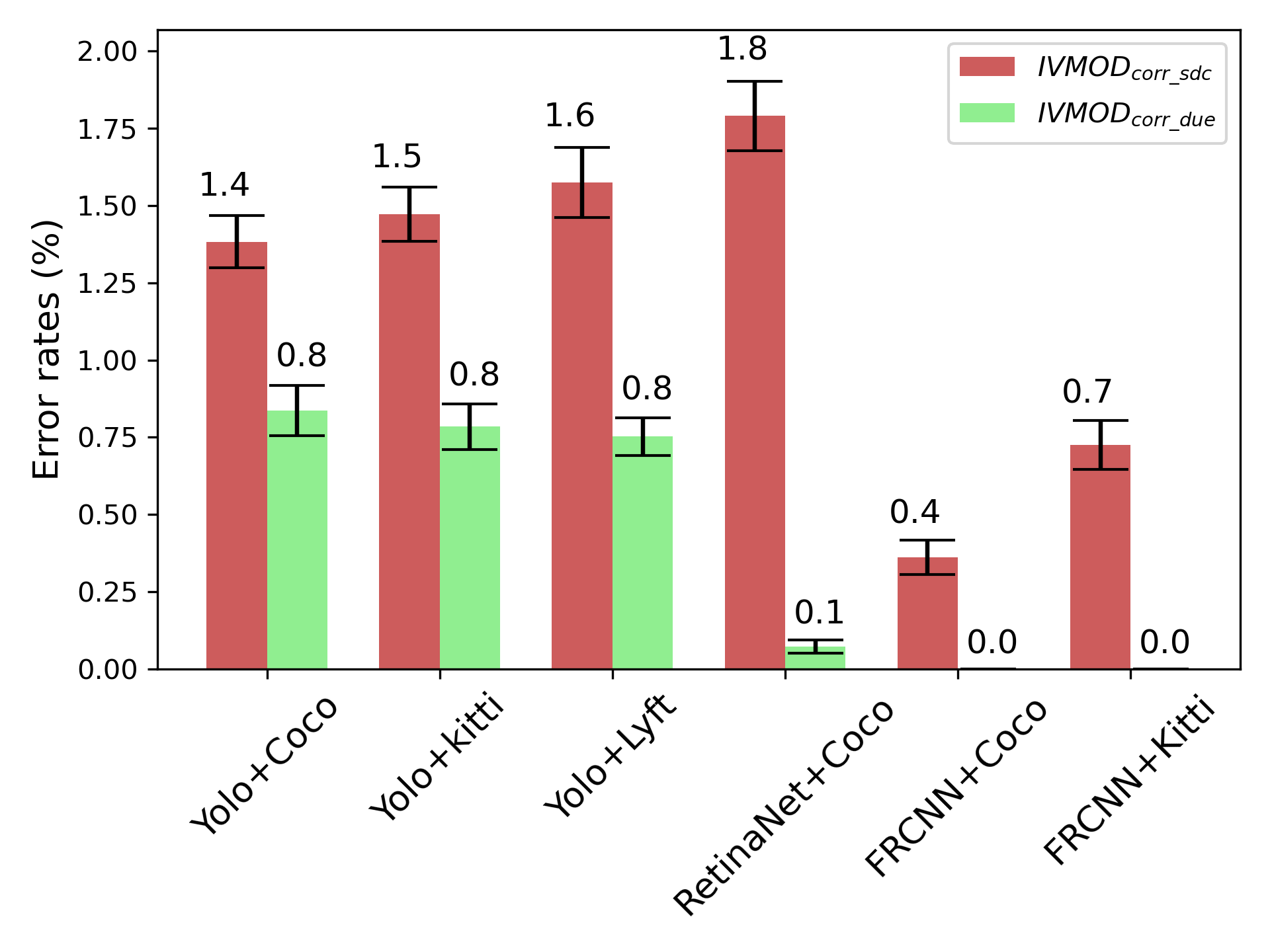}
    %\caption{A mouse}\label{fig:mouse}
    \end{subfigure}\\
    (a) Neuron faults \\
    \vspace{.1cm}

    \begin{subfigure}[b]{.32\linewidth}
    \includegraphics[width=\linewidth]{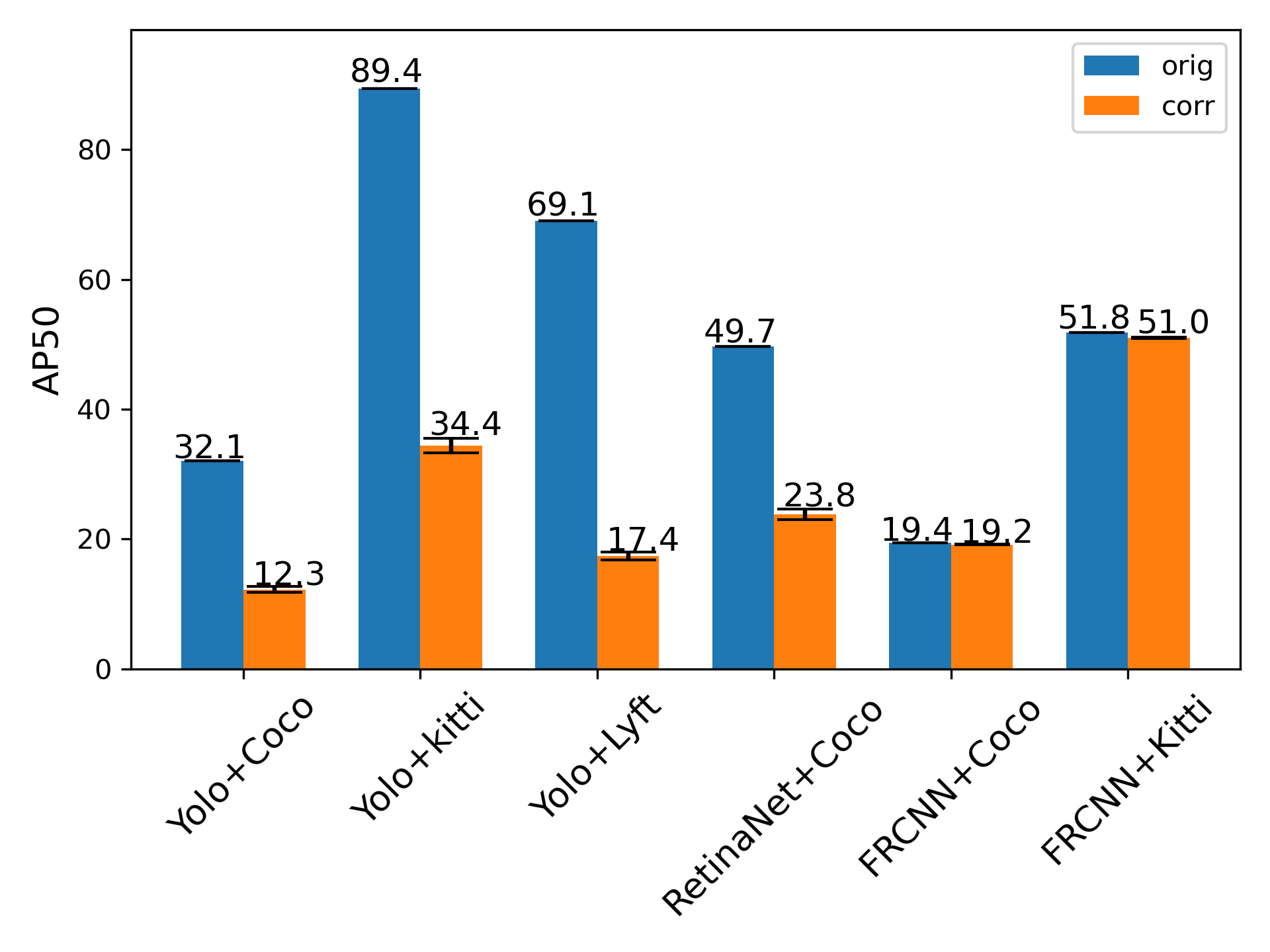}
    %\caption{A mouse}\label{fig:mouse}
    \end{subfigure}\
    \begin{subfigure}[b]{.32\linewidth}
    \includegraphics[width=\linewidth]{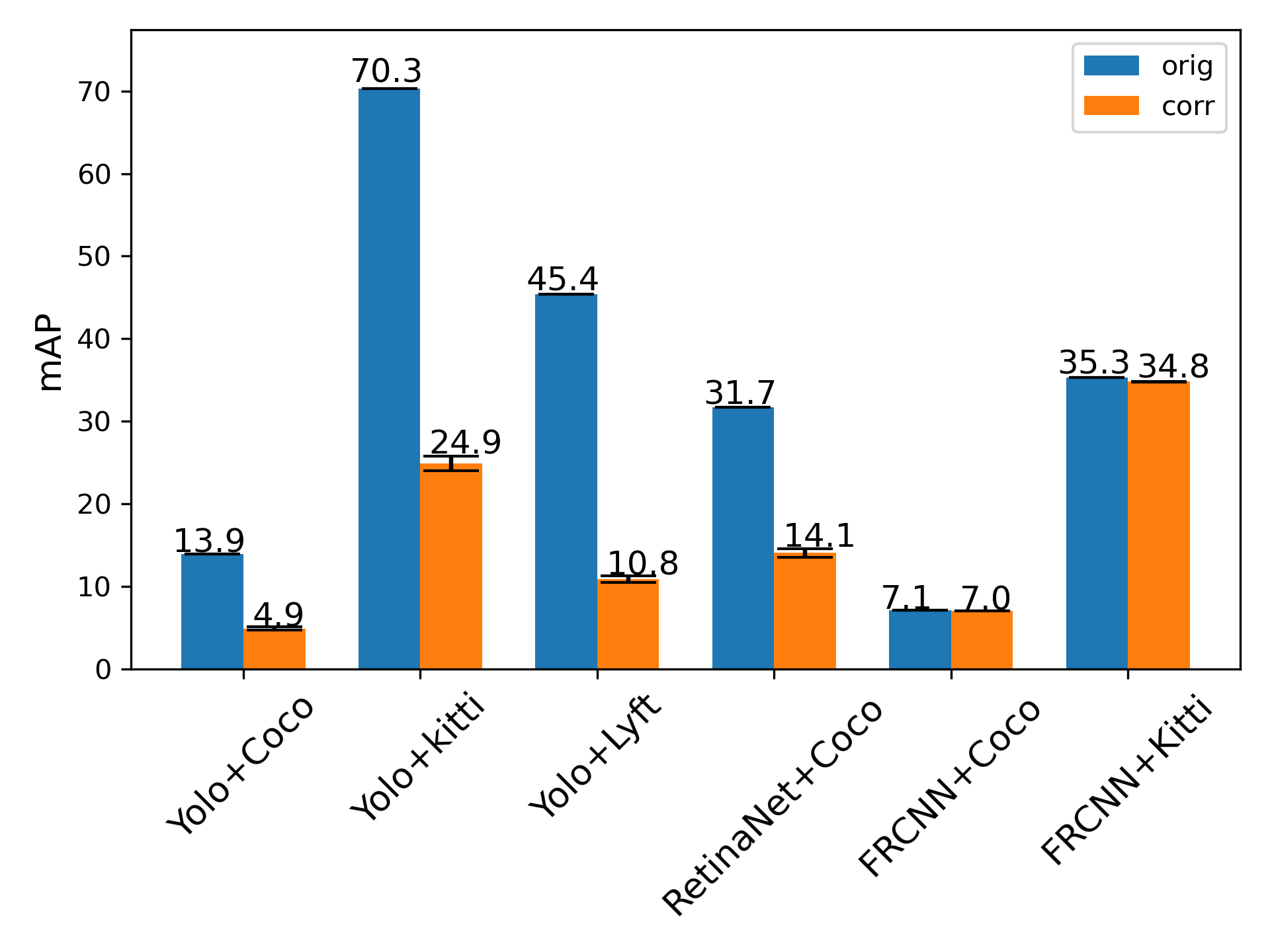}
    %\caption{A mouse}\label{fig:mouse}
    \end{subfigure}
    \begin{subfigure}[b]{.32\linewidth}
    \includegraphics[width=\linewidth]{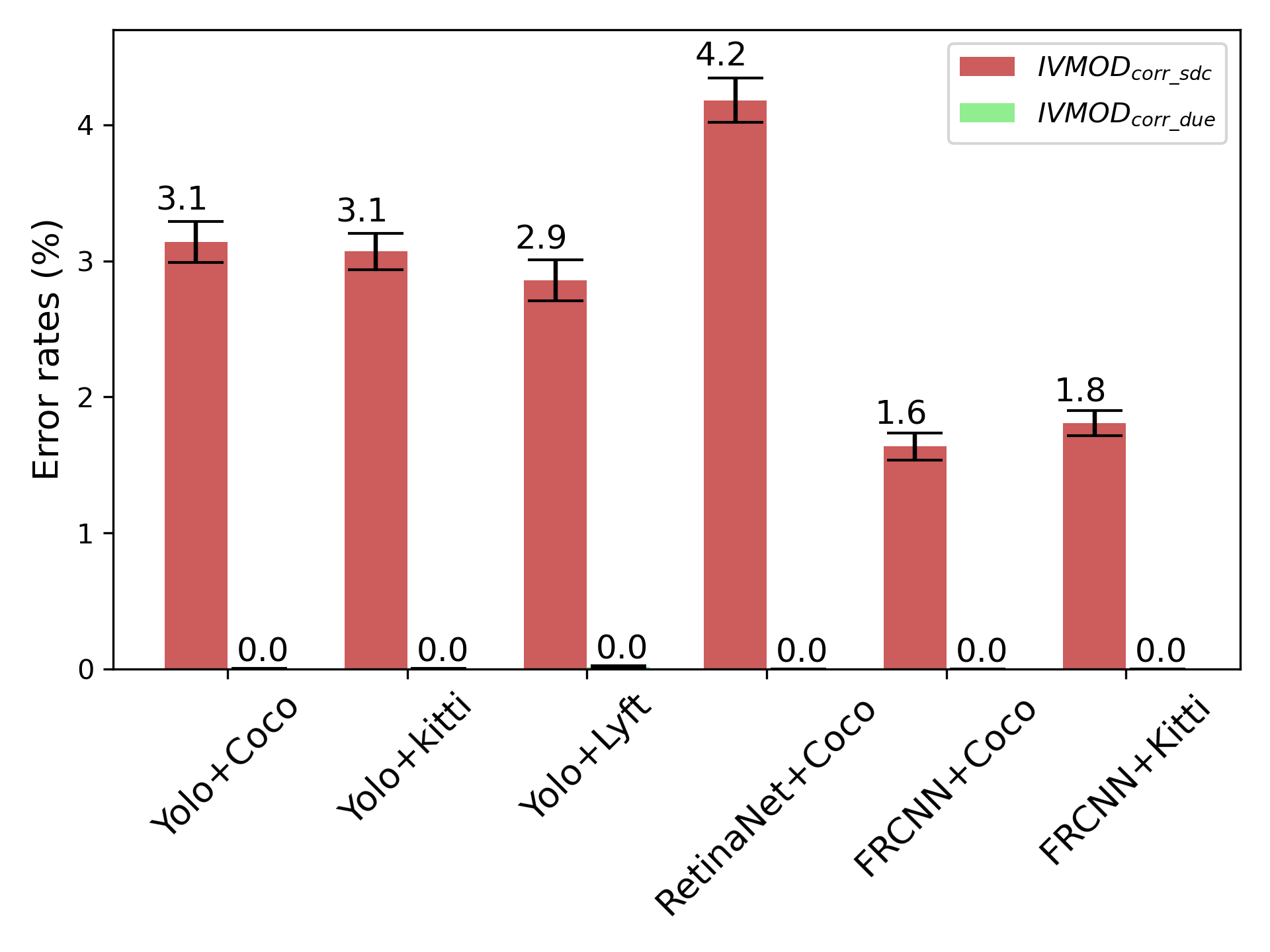}
    %\caption{A mouse}\label{fig:mouse}
    \end{subfigure}\\
    (b) Weight faults
	\vspace{-8pt}
    \caption{Key metrics to interpret the vulnerability of object detection DNNs in the presence of transient hardware faults: (left) AP50, (center) mAP, (right) error rate, distinguishing $\SDC$ and $\DUE$. We study both neuron faults (a) and weight faults (b).}
    \label{fig:fault_rates_1}
	\vspace{-8pt}
\end{figure*}

%\vspace{3mm}
\section{Transient faults}
\label{sec:transient_faults}
Our evaluation concept is guided by the assumption that in safety-critical applications, both the miss of any existing object as well as the creation of any false positive object can be potentially hazardous.
Therefore, we consider the probability that such an SDC event occurs and our primary metrics $\SDC$ and $\DUE$ (Eq.~\ref{eq:SDC_DUE}) captures the vulnerability of a model. For transient faults, this evaluation is performed in Sec.~\ref{sec:error_probabilities}. Accordingly, we independently inject 50,000 random single-bit flips in neurons and weights at each inference of the chosen test datasets.
Subsequently, Sec.~\ref{sec:error_severities} discusses the severity of each of those SDC events in terms of the average impact of additional FP and FN objects, their size, and confidence.
If a specific use case is given, the factors of probability and severity can be used to derive the risk of an error \cite{Koopman2019}.

\begin{wrapfigure}{R}{.35\textwidth}
	% \vspace{-20pt}
	\vspace{-20pt}
    \begin{minipage}{\linewidth}
    \centering
    \includegraphics[width=\linewidth]{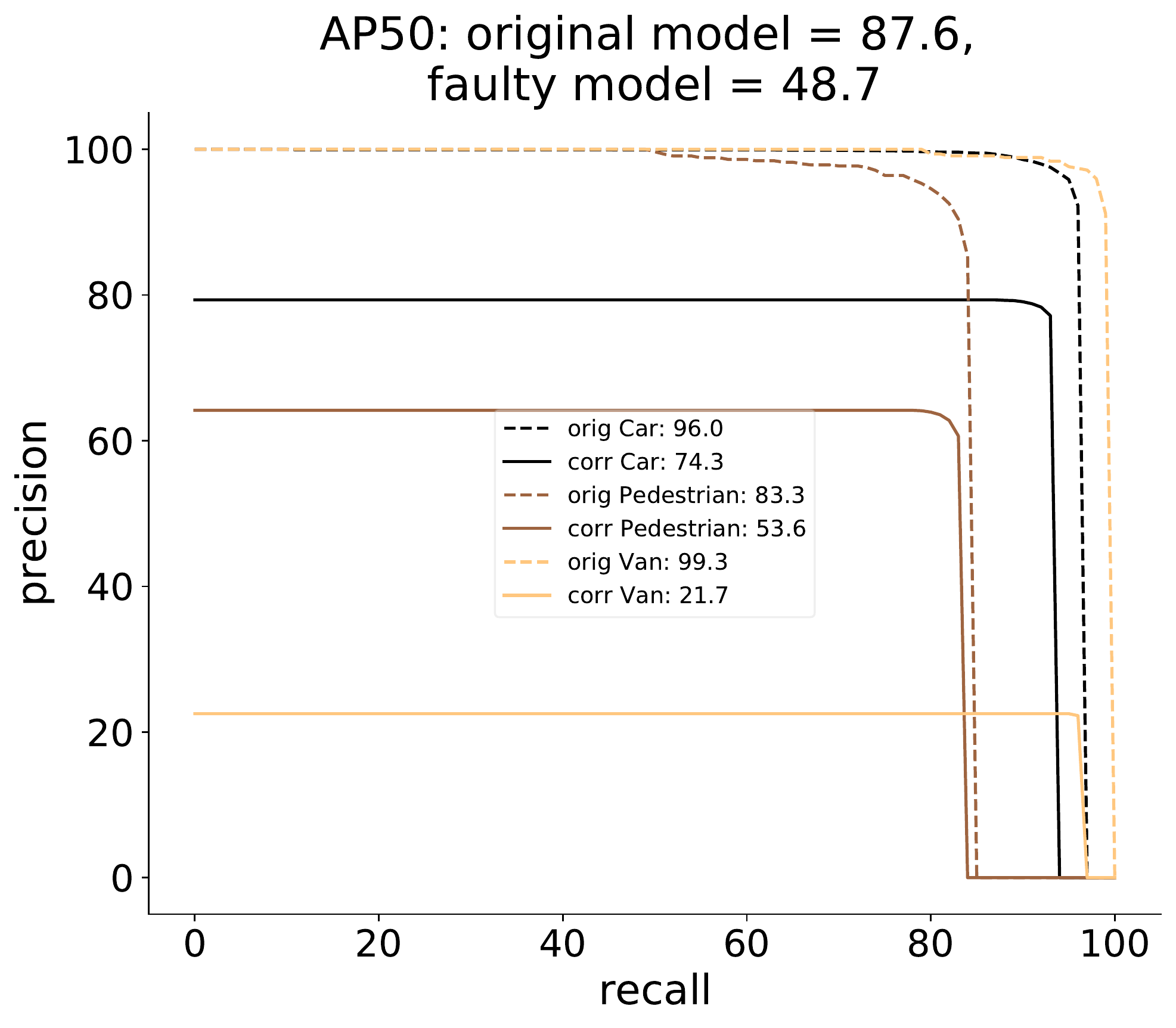}
\end{minipage}
\vspace{-8pt}
\caption{Example of the AP50 PR curves of few classes from Yolov3 and Kitti in the fault free and faulty cases.}
\vspace{-18pt}
\label{fig:pr_curve_explain}

% \begin{wrapfigure}{R}{.35\textwidth}
% 	% \vspace{-20pt}
% 	\vspace{-20pt}
%     \begin{minipage}{\linewidth}
%     \centering
%     \includegraphics[width=\linewidth]{metrics/pr_curves_explain/prCurve_orig.pdf}
% 	\par\vfill \vfill
%     \includegraphics[width=\linewidth]{metrics/pr_curves_explain/prCurve_corr.pdf}
% \end{minipage}
% \vspace{-8pt}
% \caption{Example of the AP50 PR curves of Yolov3 and Kitti in the fault free (left) and the faulty case (right).}
% \vspace{-18pt}
% \label{fig:pr_curve_explain}
\end{wrapfigure}
% \begin{wrapfigure}{R}{0.5\textwidth}
%     \centering
%     \includegraphics[width=0.3\textwidth, height=3cm]{metrics/pr_curves_explain/prCurve_orig.pdf}
% 	\smallskip\par
%     \includegraphics[width=0.3\textwidth, height=3cm]{metrics/pr_curves_explain/prCurve_corr.pdf}
%     \caption{Example of the AP50 PR curves of Yolov3 and Kitti in the fault free (left) and the faulty case (right).}
%     \label{fig:pr_curve_explain}
% \end{wrapfigure}

% \begin{wrapfigure}{R}{0.35\textwidth}
% 	\begin{center}
% 	  \includegraphics[width=0.34\textwidth]{metrics/pr_curves_explain/prCurve_orig.pdf}
% 	\end{center}
% 	\begin{center}
% 		\includegraphics[width=0.34\textwidth]{metrics/pr_curves_explain/prCurve_corr.pdf}
% 	\end{center}	
% %     \caption{Example of the AP50 PR curves of Yolov3 and Kitti in the fault free (left) and the faulty case (right).}
% %     \label{fig:pr_curve_explain}
% \end{wrapfigure}

% \begin{figure}[!ht]
%     \centering
%     \includegraphics[width=0.45\textwidth]{metrics/pr_curves_explain/prCurve_orig.pdf}
%     \includegraphics[width=0.45\textwidth]{metrics/pr_curves_explain/prCurve_corr.pdf}
%     \caption{Example of the AP50 PR curves of Yolov3 and Kitti in the fault free (left) and the faulty case (right).}
%     \label{fig:pr_curve_explain}
% \end{figure}

\subsection{Corruption probability}
\label{sec:error_probabilities}

In Fig. ~\ref{fig:fault_rates_1}, we present the fault injection campaigns of all studied networks, comparing the typical benchmark metrics AP-50 and mAP to the  $\SDC$ and $\DUE$ rate as defined in Sec.~\ref{sec:sdc_due}.
Both Yolov3 and RetinaNet show a significant change in the AP-50 and mAP metrics under the injected neuron and weight faults: The accuracy can drop as much as from $89.4\%$ to $34.4\%$ (AP-50) due to a single weight fault in the scenario of Yolov3 and Kitti.
On the other hand, F-RCNN does not showcase much sensitivity to the injected faults ($\lesssim 0.8\%$ change in AP-50). At the same time, the$\SDC$ rates vary between $0.4\%$ and $1.8\%$ (neuron faults), and from $1.5\%$ to $4.2\%$ (weight faults). This discrepancy illustrates the need for a more realistic vulnerability estimate.
As shown below, in Tab.~\ref{tab:sev_features}, fault injections in Yolov3 and RetinaNet tend to produce many FPs with statistically increased confidence. This leads to a drastic shift of the PR curves, as shown in the example in Fig. ~\ref{fig:pr_curve_explain}, where only $3.2\%$ out of 1000 samples have corrupted prediction (demonstrated the similar effect in Fig. ~\ref{fig:pr_example}(c)). Rare classes are susceptible to such faults, diminishing the class-averaged metric further. Since the induced false objects are concentrated on only a few images, the AP metric exaggerates the safety-related vulnerability of the model under software errors (see also the discussion in Sec.~\ref{sec:ap}).\\
In contrast, the F-RCNN model architecture appears to be very robust against the generation of FPs (see Tab.~\ref{tab:sev_features}). Predictions made in the presence of a soft error have nearly the same confidence as in the fault-free case. However, faults do disturb the detection of objects as a significant portion of FNs appear (on average between $10-33\%$). Nevertheless, the AP metrics for FRCNN under fault injection are hardly affected: We observe very few accuracies drops for both neurons and weights. At the same time about $0.4-0.7\%$ ($1.5-1.8\%$) of the images see silent data corruption. In this case, the AP-based metric is masking the potentially safety-critical impact of underlying faults.
We further observe that for Yolov3, $\DUE$ events are generated in $\sim0.9\%$ of the neuron injection cases, while in RetinaNet and F-RCNN and for weight injections, those are negligible ($\lesssim 0.1\%$). 
We conclude that the Yolov3 architecture stimulates neuron values that have a higher chance if being flipped to a configuration encoded as \textit{NaN} or \textit{Inf} (in FP32, all exponential bits have to be in state '1'), compared to RetinaNet and F-RCNN.
The weight values of all networks, on the other hand, are closely centered around zero, which makes it very unlikely to reach a \textit{NaN} or \textit{Inf} bit configuration \cite{GeisslerQRAPDGP21} (typically MSB and at least another exponential bit are in state '0' at the same time).
We observe that the faults injected in weights at any bit of FP32 cause higher $\SDC$ rates than the faults injected at the neuron level. They showcase $\sim 2\times$ more adverse effects on predictions than faults injected at the neuron level.

\subsection{Corruption severity}
\label{sec:error_severities}

We next aimed to understand how faults leading to $\SDC$ events corrupt images and how the severity of an $\SDC$ event on a potential safety-critical application can be estimated.
Even though the relevance of a safety feature may depend on the specific application, we identified the following fundamental features to serve as a specific indicative measure of an SDC fault severity, see Tab.~\ref{tab:sev_features}:
\begin{itemize}
\item The average number of FP objects induced by a given $\SDC$ fault and the proportion of boxes lost due to a fault, referred to as $\FPd$ and $\FNnd$, respectively as described in Eq. ~\ref{eq:delta_fp_fn} (subscript 'n' represents normalization as the upper limit of  FNs is known, in contrast to FPs).

% \item The average number of FP objects induced by a given $\SDC$ fault and the proportion of boxes lost due to a fault, referred to as $\FPd$ and $\FNnd$ respectively as described in Eq. (n subscript in the given equation represents normalization factor as FN's upper limit is known and we cannot normalize the FP as fault induced FP has no upper limit).

\item The average size of objects in the presence and absence of SDC ($\avg(\text{size})$) since a significant change of the object size can be safety-critical,
\item The average area of the image that is erroneously occupied due to $\SDC$ induced FP objects ($\Aoccfp$) and the average portion of the vacant area created by not detecting the objects due to $\SDC$ faults ($\Aoccfn$).
\item The average confidence of objects in the presence and absence of $\SDC$, $\avg(\text{conf})$.
\end{itemize}
We motivate this choice more in the following subsections.

\begin{table}[!ht]
	\vspace{-25pt}
	\caption{Severity features averaged over all $\SDC$ events.}
	\vspace{.1cm}
    \centering
		\resizebox{\textwidth}{!}
    {\begin{tabular}{c||c|c|c|c|c|c}
				~ & \textbf{Yolo+Coco} & \textbf{Yolo+Kitti} & \textbf{Yolo+Lyft} & \textbf{Retina+Coco} & \textbf{F-RCNN+Coco} & \textbf{F-RCNN+Kitti} \\ \hline \hline
				% \textbf{Neurons:} & \multicolumn{5}{l}{} & \\ \hline
				\multicolumn{6}{l}{\textbf{Neurons:}} & \\ \hline \hline
				$\avg(\FPd)$  & $\textbf{333}$ & $36$ & $174$ & $33$ & $0$ & $0$  \\ \hline
				$\avg(\FNnd)(\%)$	& $42.2$ & $41.3$ & $\textbf{46.6}$ & $16.1$ & $25.3$ & $33.3$  \\ \hline
				\begin{tabular}[x]{@{}c@{}}$\avg(\text{conf})$\\$(\text{corr}, \text{orig})$\end{tabular} %$\avg(\text{conf})(\text{corr}, \text{orig})$ 
				& $0.99, 0.52$ & $0.99, 0.51$ & $0.99, 0.65$ & $\textbf{0.79, 0.11}$ & $0.73, 0.73$ & $0.90, 0.89$  \\ \hline
				\begin{tabular}[x]{@{}c@{}}$\avg(\text{size})/1e^3 \px$\\$(\text{corr}, \text{orig})$\end{tabular}
				& $4.3, 11.2$ & $\textbf{34.5, 2.3}$ & $17.8, 7.3$ & $5.6, 20.3$ & $17.0, 18.6$ & $6.3, 6.8$ \\ \hline 
				$A_\text{fp-occ} (\%)$ & $36.8$ & $\textbf{62.5}$ & $59.8$ & $0.7$ & $1.7$ & $0.0$  \\ \hline
				$A_\text{fn-vac} (\%)$ & $4.0$ & $5.1$ & $4.8$ & $\textbf{53.1}$ & $41.1$ & $39.8$  \\ \hline \hline
				%
				% \textbf{Weights:} & \multicolumn{5}{l}{} & \\ \hline
				\multicolumn{6}{l}{\textbf{Weights:}} & \\ \hline \hline
				$\avg(\FPd)$  & $\textbf{198}$ & $59$ & $145$ & $7$ & $0$ & $0$  \\ \hline
				$\avg(\FNnd)(\%)$	& $23.3$ & $21.7$ & $21.3$ & $4.0$ & $9.6$ & $\textbf{29.6}$  \\ \hline
				\begin{tabular}[x]{@{}c@{}}$\avg(\text{conf})$\\$(\text{corr}, \text{orig})$\end{tabular} %$\avg(\text{conf})(\text{corr}, \text{orig})$ 
        		& $1.00, 0.53$  & $1.00, 0.52$ & $1.00, 0.65$ & $\textbf{0.62, 0.11}$ & $0.72, 0.73$ & $0.89, 0.88$  \\ \hline
				\begin{tabular}[x]{@{}c@{}}$\avg(\text{size})/1e^3 \px$\\$(\text{corr}, \text{orig})$\end{tabular}
				& $5.5, 12.1$ & $21.4, 2.5$ & $\textbf{30.8, 6.9}$ & $7.9, 19.8$ & $10.0, 15.0$ &  $4.9, 5.0$ \\ \hline %\multicolumn{1}{|c|}{\makecell[l]{$\avg(\text{size})/1e^3 \px$\\ $(\text{corr}, \text{orig})$} }
				$A_\text{fp-occ} (\%)$ & $40.1$ & $\textbf{81.0}$ & $79.1$ & $1.5$ & $0.3$ & $0.0$  \\ \hline
				$A_\text{fn-vac} (\%)$ & $15.1$ & $2.5$ & $6.8$ & $42.3$ & $77.8$ & $\textbf{85.8}$  \\ \hline \hline
    \end{tabular}}
	\label{tab:sev_features}
	\vspace{-25pt}
\end{table}
\vspace{-10pt}

\subsubsection{Fault-induced object generation and loss}
\label{subsec:Fault-induced object generation and loss}
Object detection is commonly used in scenarios where the number of objects, combined with their location and class, is input to safety-critical decision making. Examples include face detection or vehicle counting in traffic surveillance, automated driving, or medical object detection.
Therefore, to assess $\SDC$ severity, we quantify the impact of a fault injection by the differences (a loss in TPs equals the gain in FNs)
%FPs and FNs 
%To quantify the effect of faults on FPs and FNs, we define the bit-averaged false positive difference, $\bitavg(\FPd)$, which intuitively tells us how many FPs an SDC event with this bit position induces, on average. 
%Similarly, for FNs, the normalized bit-averaged difference, $\bitavg(\FNnd)$, represents what portion of the originally detected objects disappears due to an SDC event with a specific bit position (a loss in TPs equals a gain in FNs). That is,
%\begin{align}
%\FPbad &= \bitavg\left( FP_{\text{corr}}- FP_{\text{orig}}\right), \\
%\FNband &= \bitavg\left( (TP_{\text{orig}}- TP_{\text{corr}})/TP_{\text{orig}}\right),
%\end{align}
% \begin{wrapfigure}{r}{0.4\textwidth}
\vspace{-5pt}
\begin{align}
\vspace{-25pt}
\begin{split}
\FPd &= \left( FP_{\text{corr}}- FP_{\text{orig}} \right), \\
\FNnd &=  (TP_{\text{orig}}- TP_{\text{corr}})/TP_{\text{orig}},
% \vspace{-8pt}
\end{split}
\vspace{-25pt}
\label{eq:delta_fp_fn}
\end{align}
% \end{wrapfigure}
%where for a sample list $x$ of size $N$ and its associated flipped bit list $bit(x)$ we have $\bitavg(X) =\{\avg(\{X_i|bit(X)_i=j; i=1\ldots N\}); j\in 1\ldots 32\}$ and $\avg(X) = \sum_i X_i/|\sum_i X_i|$. \\
In Tab.~\ref{tab:sev_features}, we observe that all Yolov3 and RetinaNet scenarios exhibit large numbers of fault-induced FPs ($\gg 100$ in Yolov3 and Coco experiments). For neuron faults, the generation of FPs is, on average, more pronounced. 
Furthermore, the normalized FN rates show that already a single fault can cause a significant loss of accurate positive detections. Average FN rates are higher for neuron faults than weight faults and reach averages up to $47\%$ (Yolov3 and Lyft).
F-RCNN models are robust against the generation of FP objects but not immune against fault-induced misses (e.g. Fig.~\ref{fig:example Faster-RCNN}).
The number of generated FPs and FNs varies in a broad sample range, up to the maximum limit of allowed detections (here $1000$), due to the inhomogeneous impact of flips in different bit positions (see Sec.~\ref{sec:bit-wise transient}).
In some situations, additional objects created by faults will match actual ground truth objects, leading to a negative FP or FN difference. This effect originates from the imperfect performance of the original fault-free model and is tolerated here due to the minor impact.
By relaxing the class matching constraints from one-to-one correspondence to no class matching, we can further segment the type of FPs that the $\SDC$ events cause.
It appears that situations where an FP is due to a change in the class label only or due to a shift of the bounding box only (on average $\lesssim 3$ for Yolo models, $0$ for others). In most cases, \textit{both} the bounding box gets shifted, and the class labels is mixed up, or predicted objects cannot be matched with any ground truth object at all.
\vspace{-15pt}
\subsubsection{Object size and confidence}
Box sizes and confidence values are other severity indicators since large erroneous objects take up a more significant portion of the image space, and high-confidence objects might be handled with priority in some use cases.
Tab.~\ref{tab:sev_features} shows the change of the average box size and confidence of all model detections across the identified $\SDC$ events. 
In most models, the typical box size is reduced in the presence of faults, which is partially due to the creation of boxes with zero width or height. 
However, there are also scenarios where faults tend to induce overly large objects (Yolov3 and Kitti, Lyft, see Tab.~\ref{tab:sev_features}) that can even fill the entire picture.
An object's average confidence score after fault injection significantly increases in the scenario of Yolov3 and RetinaNet, while there is hardly any impact on F-RCNN predictions.
This explains why confidence-sensitive metrics based on AP react differently to fault injections in the respective architectures; see the discussion in Sec.~\ref{sec:ap}.

\begin{figure}[h]
    \centering
		 \begin{subfigure}[height=4cm]{0.24\textwidth}
         \centering
         \frame{\includegraphics[width=\textwidth]{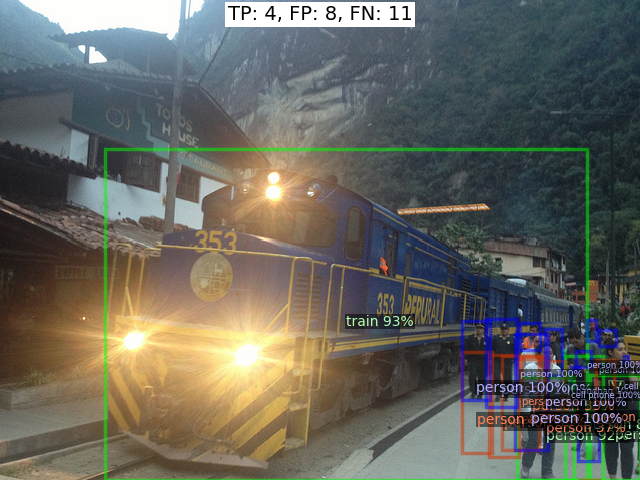}}
         \caption{orig}
         %\label{fig:y equals x}
     \end{subfigure}
		\begin{subfigure}[height=4cm]{0.24\textwidth}
         \centering
         \frame{\includegraphics[width=\textwidth]{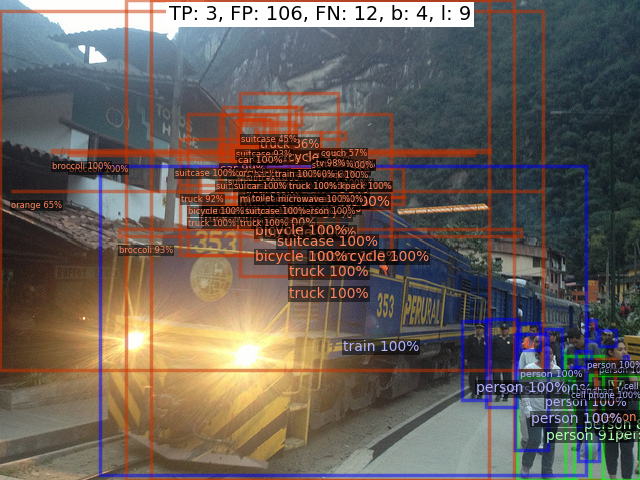}}
         \caption{corr}
         %\label{fig:y equals x}
     \end{subfigure}
		\begin{subfigure}[height=4.cm]{0.24\textwidth}
         \centering
         \frame{\includegraphics[width=\textwidth]{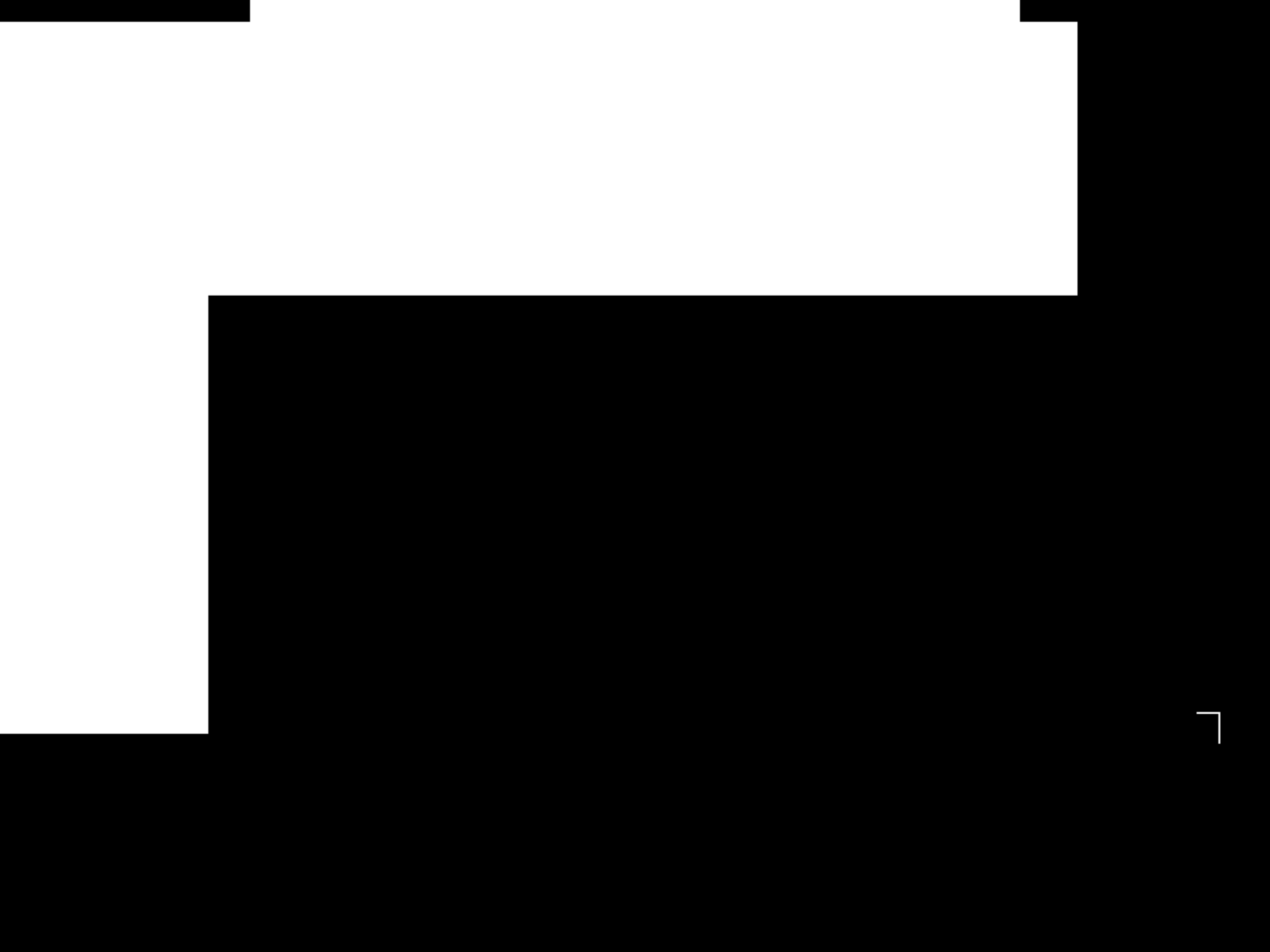}}
         \caption{$FP_{\text{blob}}$}
         %\label{fig:y equals x}
     \end{subfigure}
		\begin{subfigure}[height=4.cm]{0.24\textwidth}
         \centering
         \frame{\includegraphics[width=\textwidth]{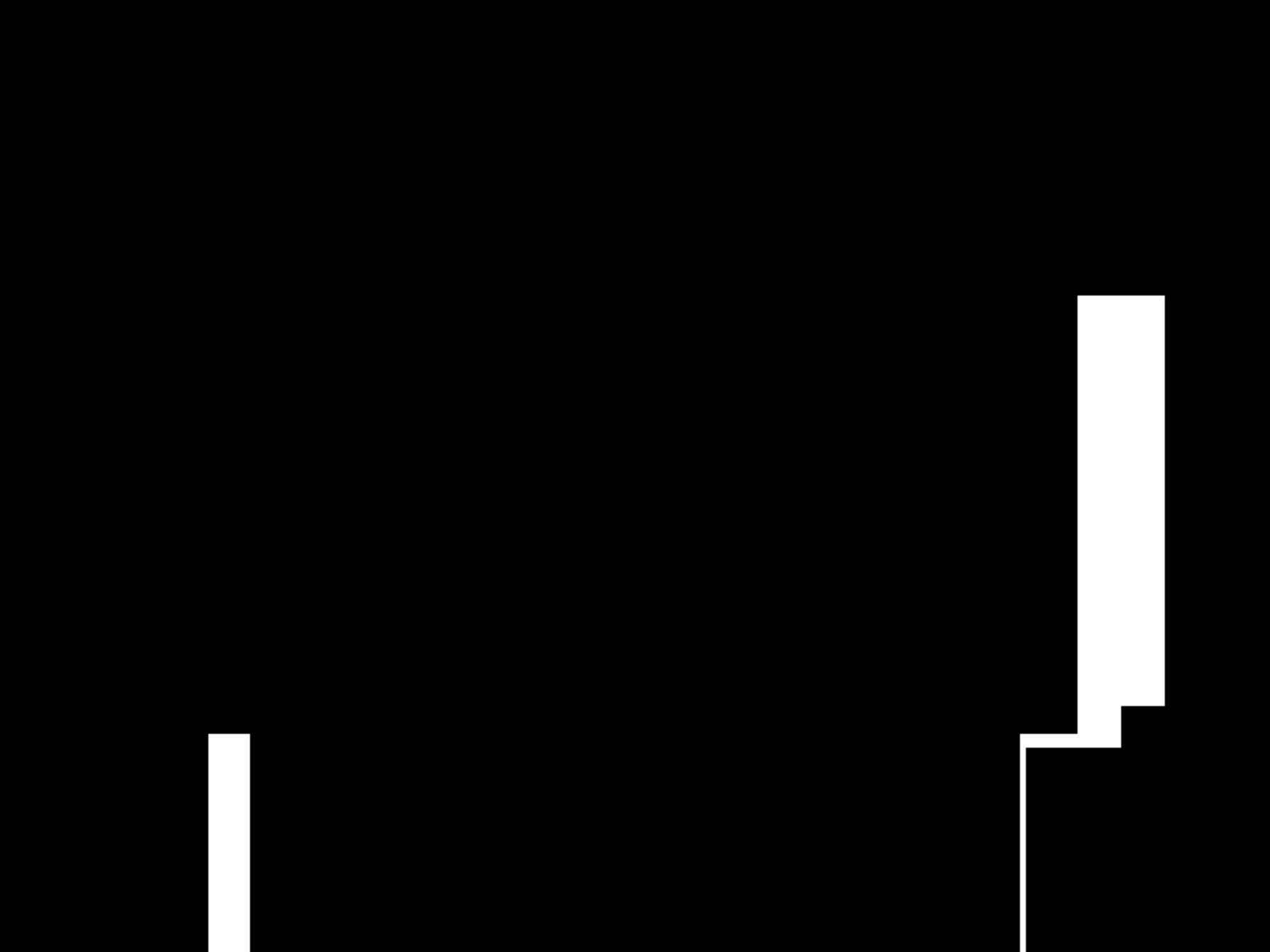}}
         \caption{$FN_{\text{blob}}$}
         %\label{fig:y equals x}
     \end{subfigure}

    %\includegraphics[height=4cm]{Pics/examples/yolov3_coco2017_76625_orig.png}&
    %\includegraphics[height=4cm]{Pics/examples/yolov3_coco2017_76625_corr.png}&
    %\includegraphics[height=4cm]{Pics/examples/test_fp_blob.png}&
    %\includegraphics[height=4cm]{Pics/examples/test_fn_blob.png}\\
		%
		%(a) orig \hspace{2cm}&(b) corr \hspace{2cm}&(c) $FP_{\text{blob}}$& \hspace{2cm}(d) $FN_{\text{blob}}$
	\vspace{-8pt}
	\caption{Illustration of the clustering of bounding boxes to binary occupancy blobs. In this example we find from (c) and (d) that $\Aoccfp=33.3\%$, $\Aoccfn=7.5\%$ (white pixels indicate space occupied by fault FPs).}%
    \label{fig:blob_example}
\end{figure}

\vspace{-15pt}
\subsubsection{Area occupancy}

safety-related decision-making in a dynamic environment is most importantly based on the detected free and occupied space. For example, an automated vehicle will determine a driving path depending on the detected drivable space.
A large number of false-positive objects, even when small in size, can, in combination, cover a significant portion of the image, which will leave only little free space. On the opposite, in some situations, they may overlay each other and occupy only a little space.
To reflect a realistic severity of free space, we first cluster all FP and FN objects to \textit{blobs} by projecting them to a binary space of occupancy and vacancy (see Fig. ~\ref{fig:blob_example}).
As we are only interested in fault-induced false objects, our blobs for a given frame at time $t$ are defined as follows:
 %\begin{equation*}

% \begin{align}
% \begin{split}
% &FP_{\text{blob}} = \mathcal{I}(\text{det}_\text{corr} - \text{det}_\text{orig}), \\
% &FN_{\text{blob}} = \mathcal{I}(\text{det}_\text{orig} - \text{det}_\text{corr}). 
%     %&FN\_blob_t = |orig\_det_t - corr\_dt_t |, 
% \label{eq:blob corr}
% \end{split}
% \end{align}
% \begin{wrapfigure}{r}{.4\textwidth}
\vspace{-10pt}
% \begin{align}
% \vspace{-25pt}
% \nonumber
% &FP_{\text{blob}} = \mathcal{I}(\text{det}_\text{corr} - \text{det}_\text{orig}), \\
% &FN_{\text{blob}} = \mathcal{I}(\text{det}_\text{orig} - \text{det}_\text{corr}). \label{eq:blob corr} \\[2pt]
% 	%&FN\_blob_t = |orig\_det_t - corr\_dt_t |, 
% \nonumber
% &\Aoccfp = |FP_{\text{blob}}|/ A_{\text{image}}, \\
% &\Aoccfn = |FN_{\text{blob}}|/ |\mathcal{I}(\text{det}_\text{orig})|. \label{eq:area_blob corr}
% \vspace{-25pt}
% \end{align}
\begin{equation}
\begin{aligned}
	% \vspace{-25pt}
	FP_{\text{blob}} = \mathcal{I}(\text{det}_\text{corr} - \text{det}_\text{orig}), \\
	FN_{\text{blob}} = \mathcal{I}(\text{det}_\text{orig} - \text{det}_\text{corr}). 
	\label{eq:blob corr} 
		%&FN\_blob_t = |orig\_det_t - corr\_dt_t |, 
	\end{aligned}
\end{equation}
% \vspace{-5pt}
\begin{equation}
\begin{aligned}
\Aoccfp = |FP_{\text{blob}}|/ A_{\text{image}}, \\
\Aoccfn = |FN_{\text{blob}}|/ |\mathcal{I}(\text{det}_\text{orig})|. 
\label{eq:area_blob corr}
% \vspace{-25pt}
\end{aligned}
\end{equation}
% \end{wrapfigure}
% symbol for positive and projected to bw 
In Eq.~\ref{eq:blob corr}, $\text{det}$ denotes the set of all detected bounding boxes (TP and FP), and $\mathcal{I}(x)$ represents the pixel-wise projection to binary occupancy space, i.e., for any pixel $u$ in a blob $x$ it is $\mathcal{I}(u<0)=0$, $\mathcal{I}(u\geq 0)=1$ (see Fig. ~\ref{fig:blob_example}). We define the occupancy coefficients in Eq. \ref{eq:area_blob corr}, where $A_{\text{image}}$ is the size of the image in pixels and $|\ldots|$ denotes the sum of all nonzero pixels in a blob.
In Tab.~\ref{tab:sev_features}, we see Yolo+Kitti creates significantly less $\FPd$ than Yolo+CoCo, but the average $\Aoccfp$, in this case, is $\sim2x$ greater than $\Aoccfp$ of Yolo+CoCo. This can even be observed using the feature $\avg(\text{size})/1e^3$ (average size of bounding boxes of all the detections combined - TPs+FPs). 
In the case of Yolo+Kitti, the $\avg(\text{size})/1e^3$ is $15x$ and $\sim 8x$ larger than its original detections when a fault is injected in neurons and weights.
This implies that $\FPd$ alone cannot determine the safety impact during an $\SDC$ event. 
Similarly, F-RCNN creates no $\FPd$, but large free space $\FNnd$ by missing the TPs. F-RCNN+Kitti, when induced with weight faults, is more safety-critical as the $\Aoccfn$ is highest compared to other studied models. Furthermore, in case of neuron faults, the RetinaNet and F-RCNN have higher $\Aoccfn$.
% TODO: This means we consider all detections (FP and TP) that do not show up in the original picture. 

%As a result, $FP_{\text{blob}}$ represents the pixels 
% With that, 

% \begin{align}
%\nonumber
%&\Aoccfporig = |FP_{\text{orig\_blob}}|/ A_{\text{image}}, \\
%&\Aoccfnorig = |FN_{\text{orig\_blob}}|/ |\mathcal{I}(\text{det}_\text{orig})|. \label{eq:area_blob orig}\\[5pt] 
%\nonumber
% \begin{split}
% %\nonumber
% &\Aoccfp = |FP_{\text{blob}}|/ A_{\text{image}}, \\
% &\Aoccfn = |FN_{\text{blob}}|/ |\mathcal{I}(\text{det}_\text{orig})|. 
% % \label{eq:area_blob
% \label{eq:area_blob corr}
% \end{split}
% \end{align}
\begin{figure}[!h]
\centering
\begin{subfigure}[b]{0.48\linewidth}
    \centering
		%\hspace{-.8cm}
    \includegraphics[width=0.9\textwidth, height=4cm]{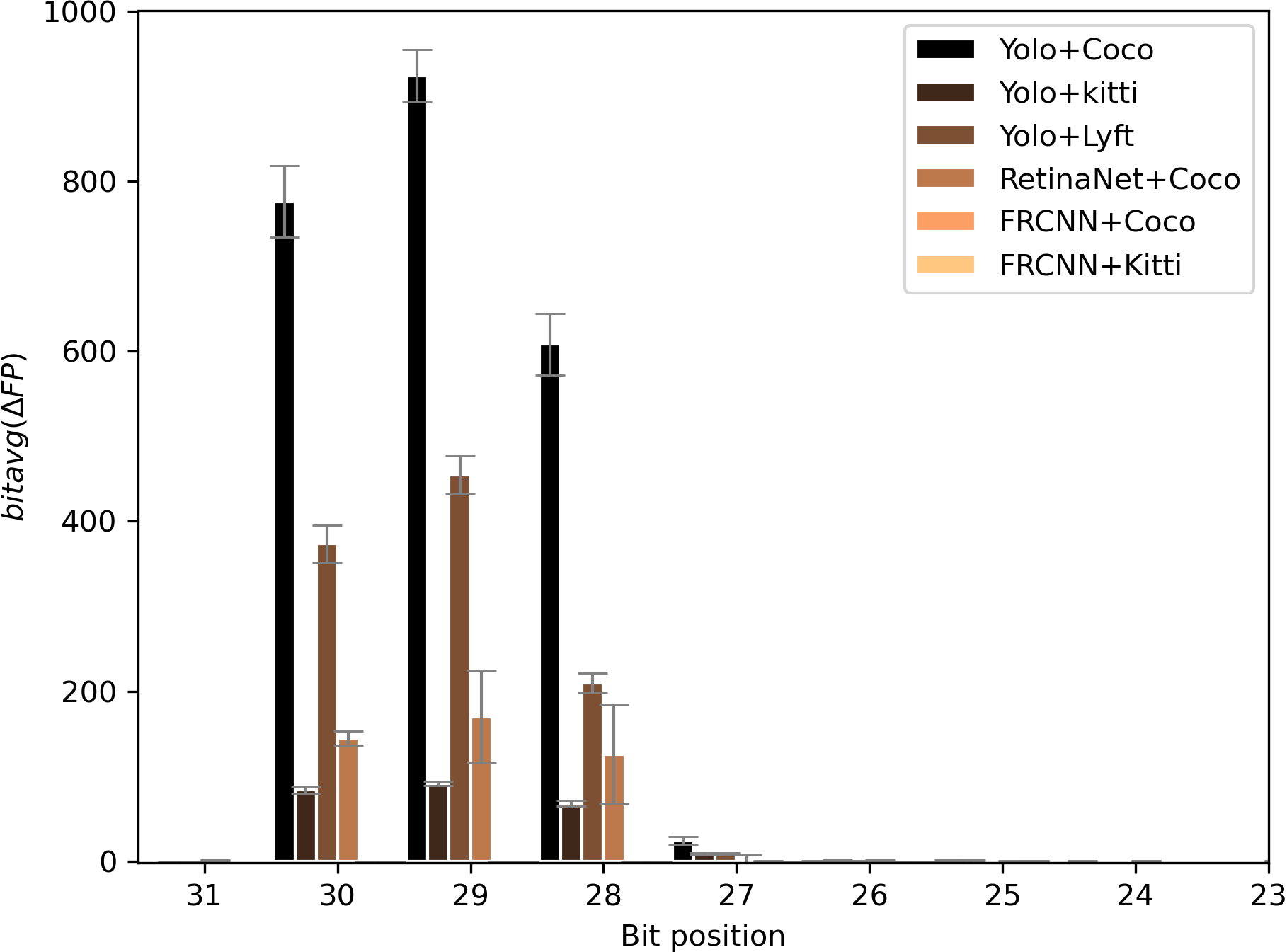}
    \caption{FP Neurons}
    \end{subfigure}
		%\hspace{-.8cm}
    \begin{subfigure}[b]{0.48\linewidth}
    \centering
		\includegraphics[width=0.9\textwidth, height=4cm]{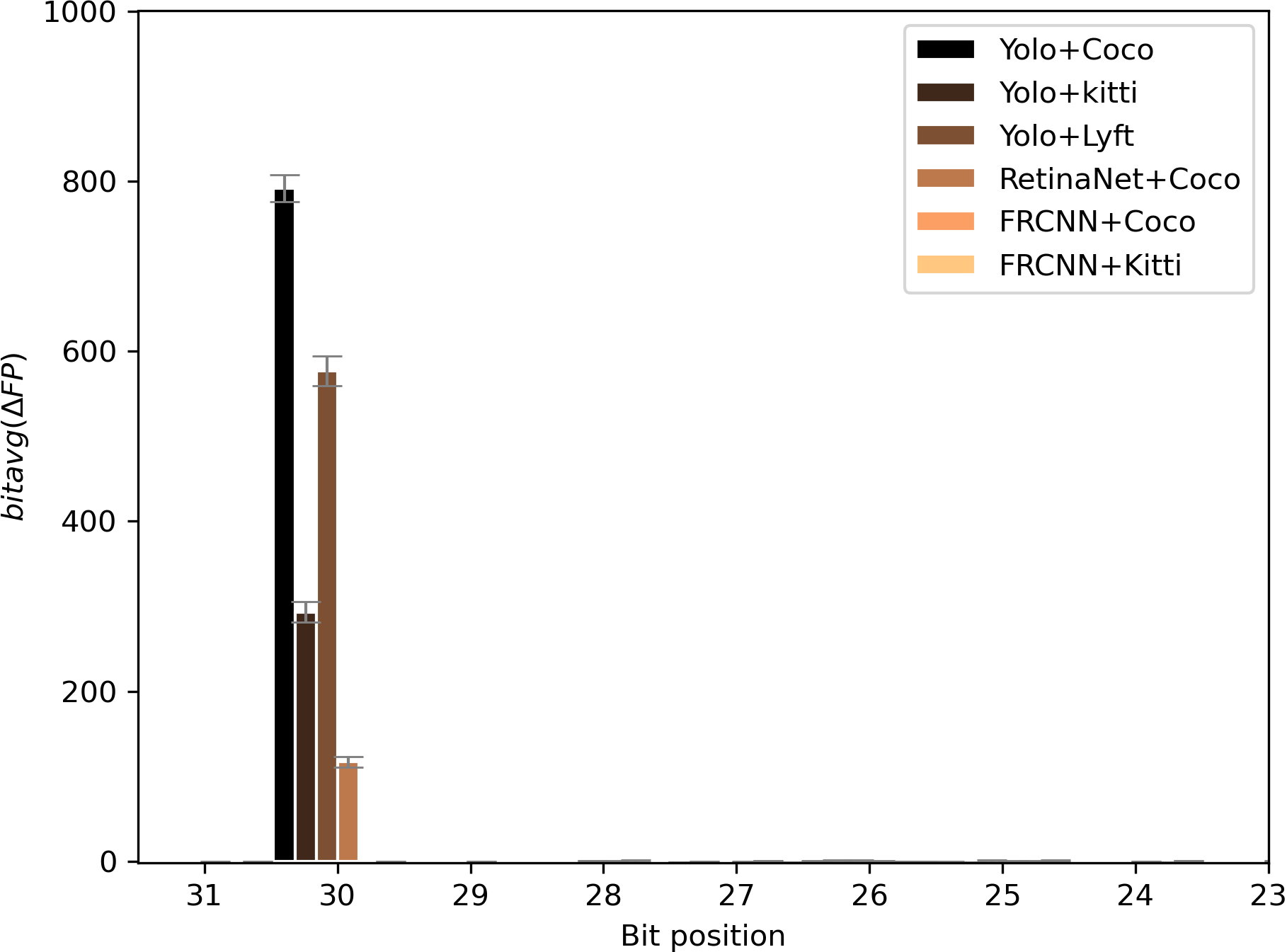}
		\caption{FP Weights}
    \end{subfigure}
		%\hspace{-.8cm}
		%
    \begin{subfigure}[b]{0.48\linewidth}
    \centering
    \includegraphics[width=0.9\textwidth, height=4cm]{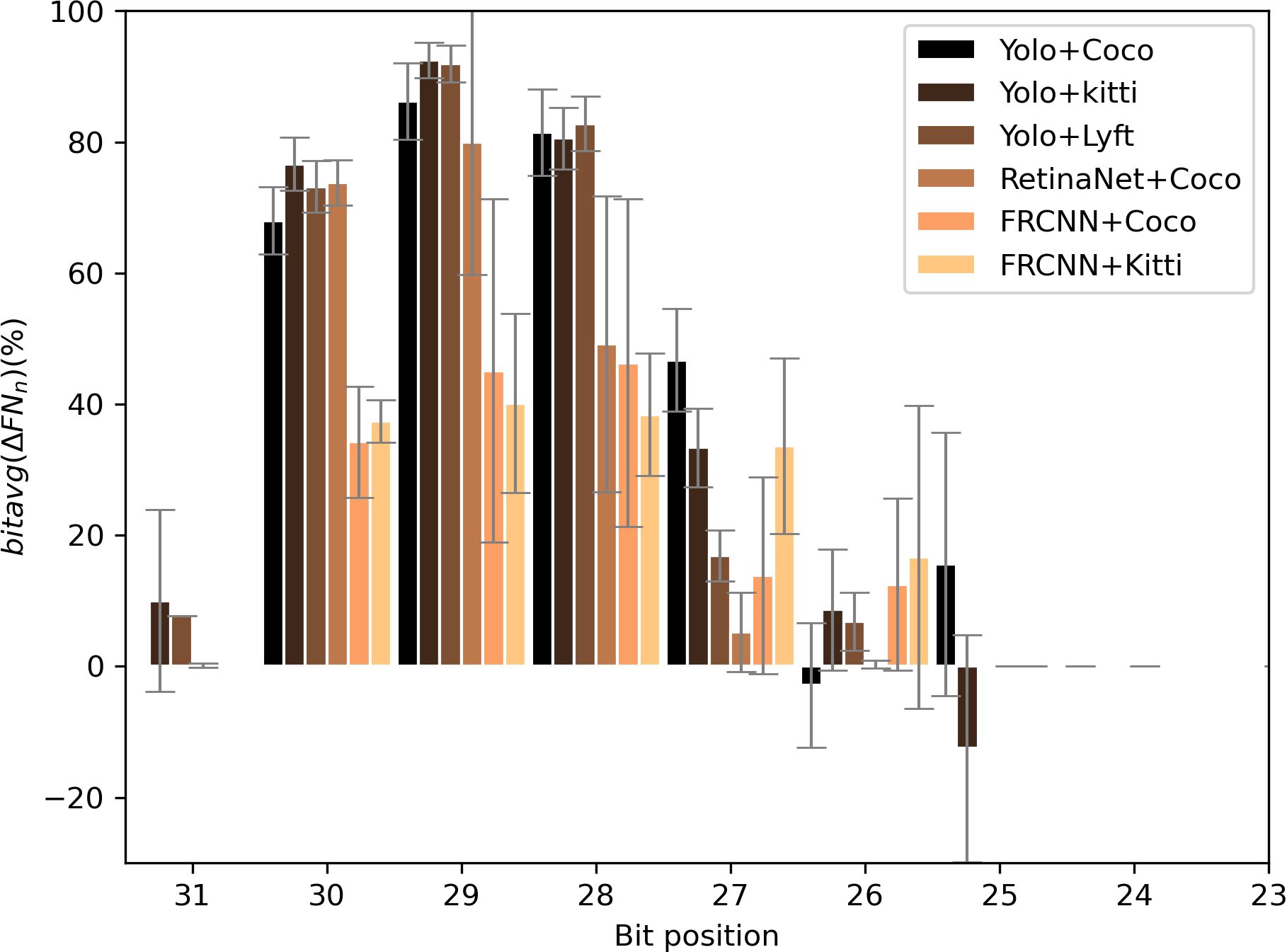}
    \caption{FN Neurons}
    %\label{fig:mouse}
    \end{subfigure}
		%\hspace{-.8cm}
    \begin{subfigure}[b]{0.48\linewidth}
		%\hspace{-.8cm}
    \centering
    \includegraphics[width=0.9\textwidth, height=4cm]{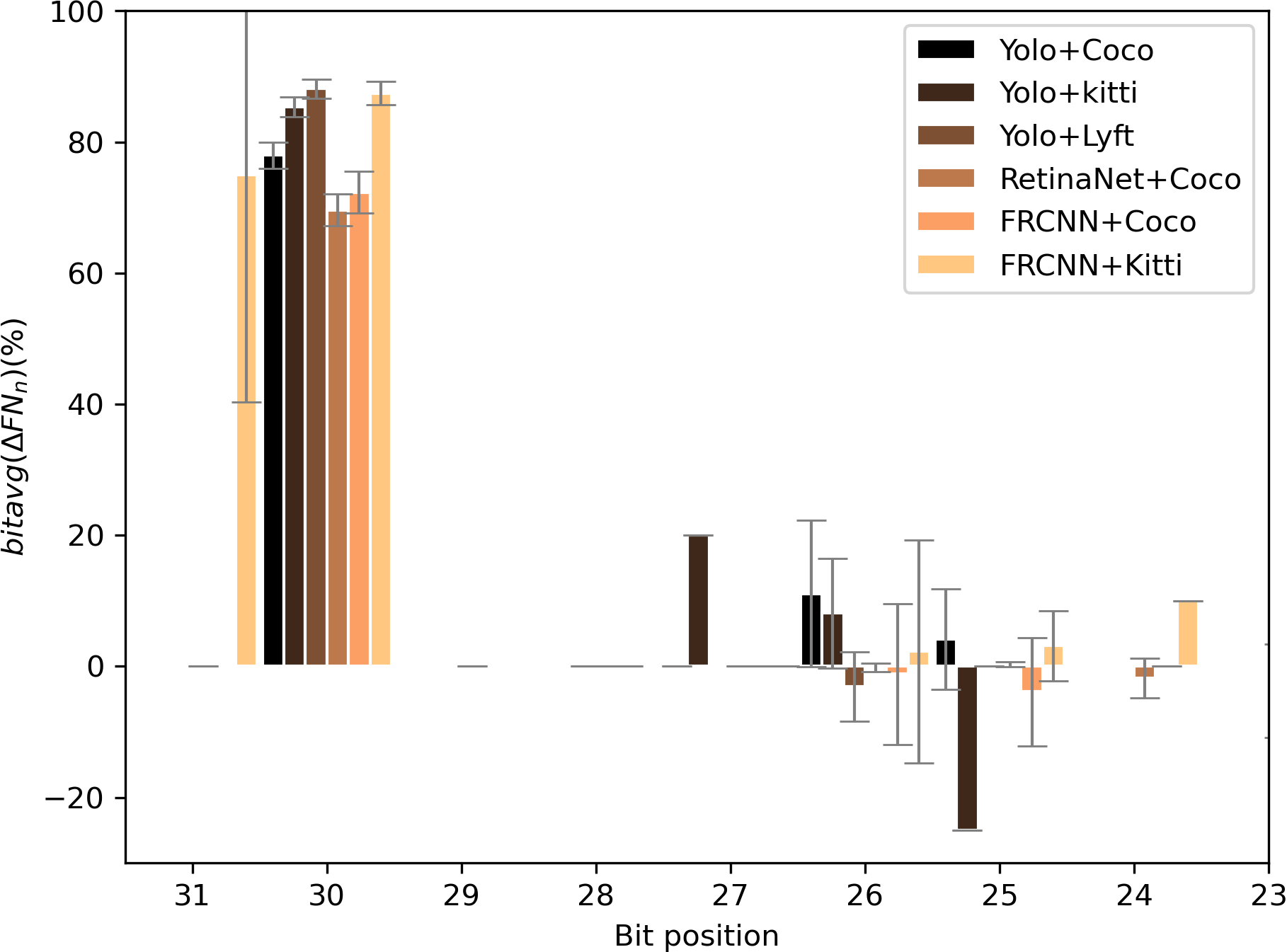}
    \caption{FN Weights}
    %\label{fig:mouse}
    \end{subfigure}
	\vspace{-8pt}
\caption{Bit-wise analysis of the severity of $\SDC$ events. Diagrams show the FP difference (a), (b) and FN rates (c), (d) for neurons and weights, respectively. Bit 31\textsuperscript{st} is the sign bit, 30\textsuperscript{th} bit being the most significant bit and 23\textsuperscript{rd} bit is the lowest bit of exponent part.}
\label{fig:FPFNvsbit}
\vspace{-8pt}
\end{figure}
\vspace{-10pt}

\subsection{Bit-wise analysis of false object count}
\label{sec:bit-wise transient}
% Interpretation Fig.~\ref{fig:FPFNvsbit}: \\
%We next aim to understand how faults that lead to SDC events actually corrupt images. 
The severity of an $\SDC$ event typically depends on the magnitude of the altered values, where values with a considerable absolute value are more likely to propagate and disrupt the network predictions \cite{Li2017, Hong2019, GeisslerQRAPDGP21}. 
Therefore, the severity features are expected to form a non-uniform distribution depending on the flipped bit position. 
To gain a better intuition, we here choose to present a bit-wise analysis of the $\FPd$ and $\FNnd$ samples during the $\SDC$ events. 
To quantify the impact of bits, we define the bit-averaged false-positive difference, $\bitavg(\FPd)$, which intuitively tells us how many FPs an SDC event with a particular bit position induces, on average. 
Similarly, for FNs, the normalized bit-averaged difference, $\bitavg(\FNnd)$, represents what portion of the originally detected objects disappears due to an SDC event with a specific bit position.
In Fig. ~\ref{fig:FPFNvsbit}, we observe that, for neuron faults, those additional FPs are typically caused by bitflips in either of the three highest exponential bits, as long as those do not lead to DUE instead. 
For weight faults, we find a situation similar to classifier networks where the specific value range of weights centered around zero is encoded in bit constellations where the MSB is in state '0' while the next higher exponential non-MSB bits are in state '1', see Ref. \cite{GeisslerQRAPDGP21}.
This explains why almost only MSB flips induce large values and $\SDC$ (with a high number of FPs).
Given the respective relevant neuron and weight bit flips, the $\FNnd$ ratio is increased up to $90\%$ (meaning that portion of all true positive detections is lost) in some of the models as shown in Fig. ~\ref{fig:FPFNvsbit}(c),(d)). In particular, due to MSB and other high exponential bit flips the average $\bitavg(\FNnd)$ is $\sim 47\%$. We observe that FN alterations can, to some extent, be induced also by lower exponential bits.

\vspace{-15pt}
% \vspace{-5pt}

%% file: results_2.tex
\begin{figure*}[!h]
  \settoheight{\tempdima}{\includegraphics[width=0.23\linewidth]{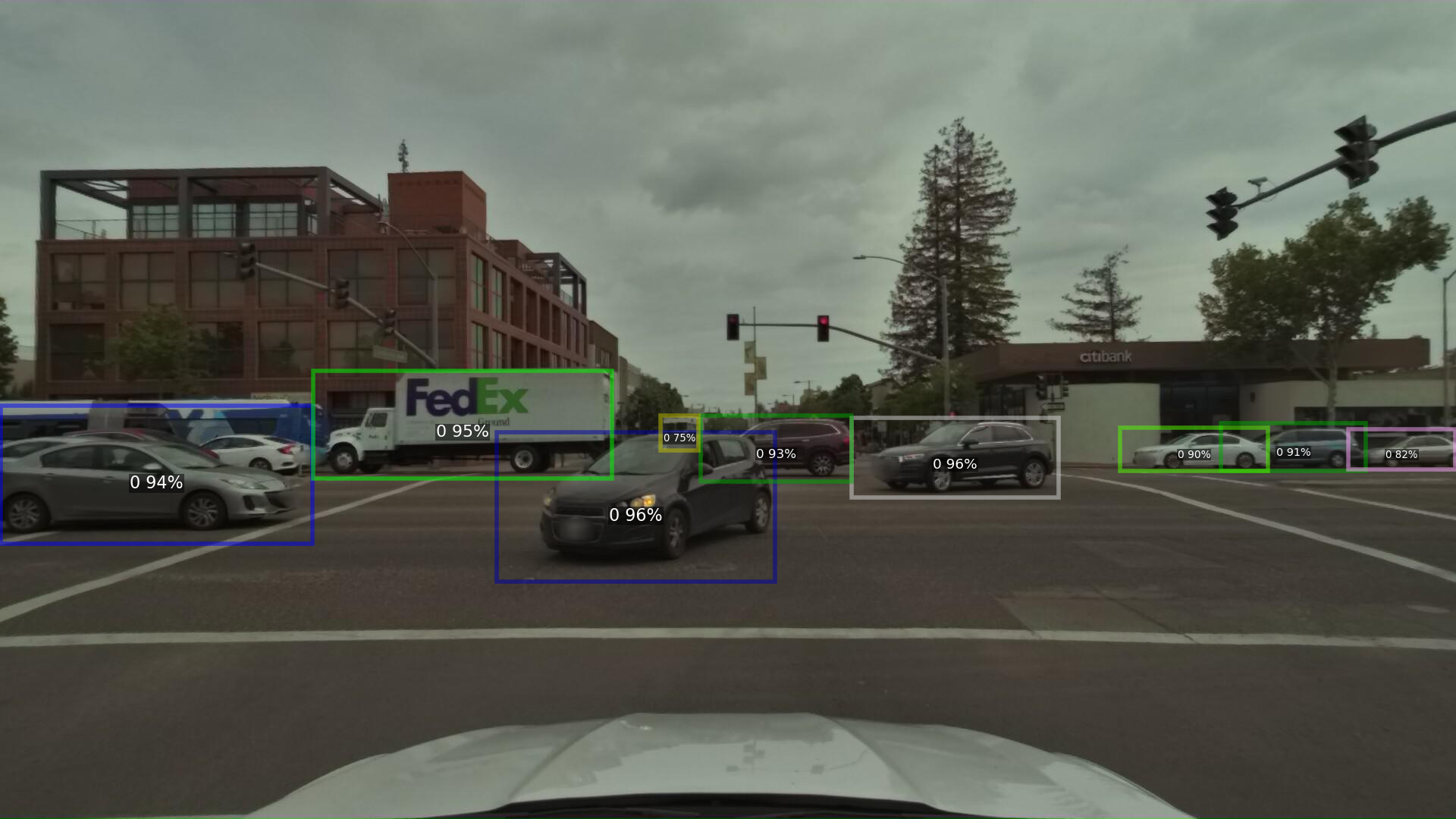}}%
  \centering\begin{tabular}{@{}c@{ }c@{ }c@{ }c@{}}
  &\textbf{frame t=94} & \textbf{frame t=96} & \textbf{frame t=98} \\
  \rowname{orig dt}&
  \frame{\includegraphics[width=.3\linewidth]{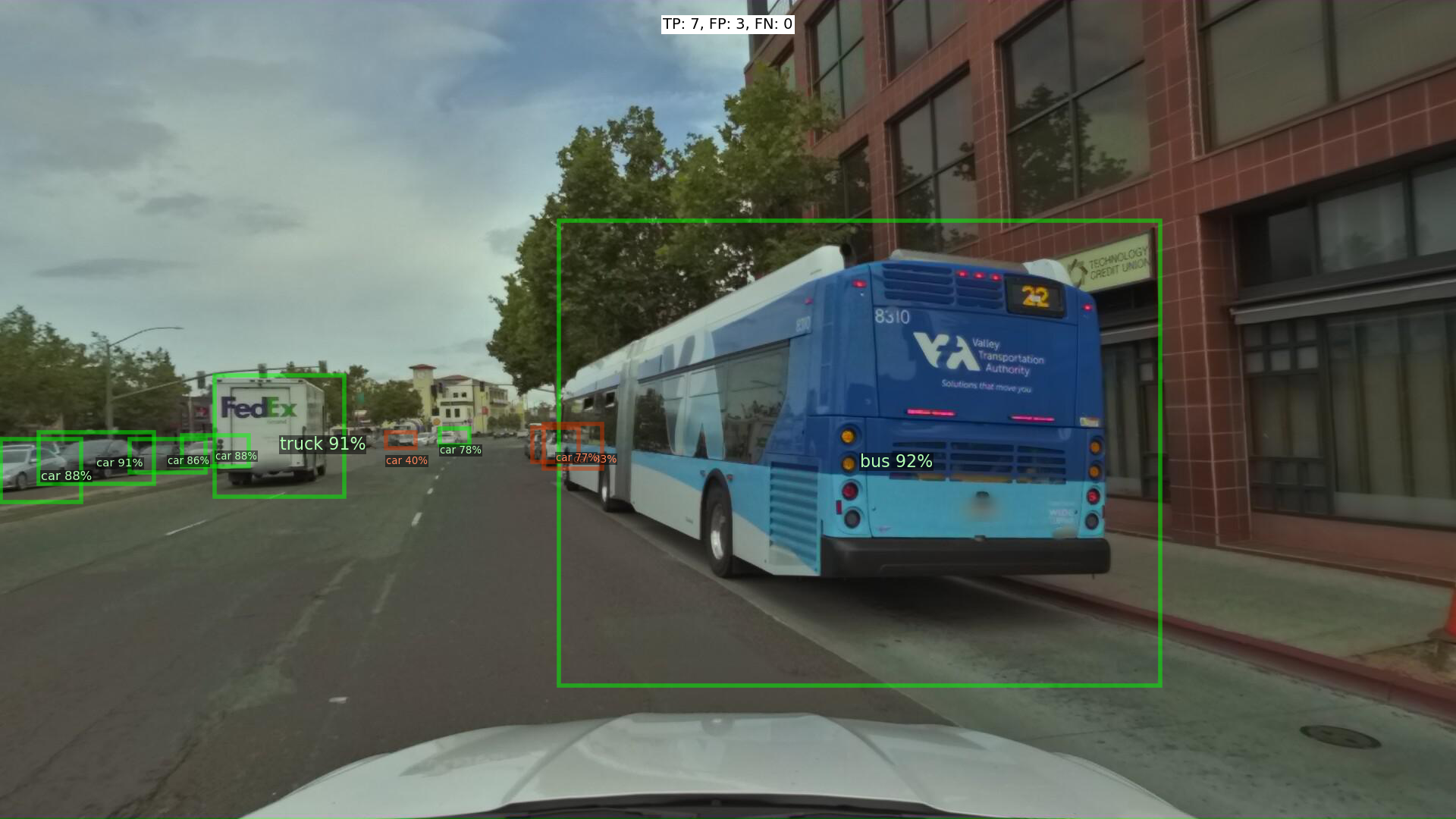}}&
  \frame{\includegraphics[width=.3\linewidth]{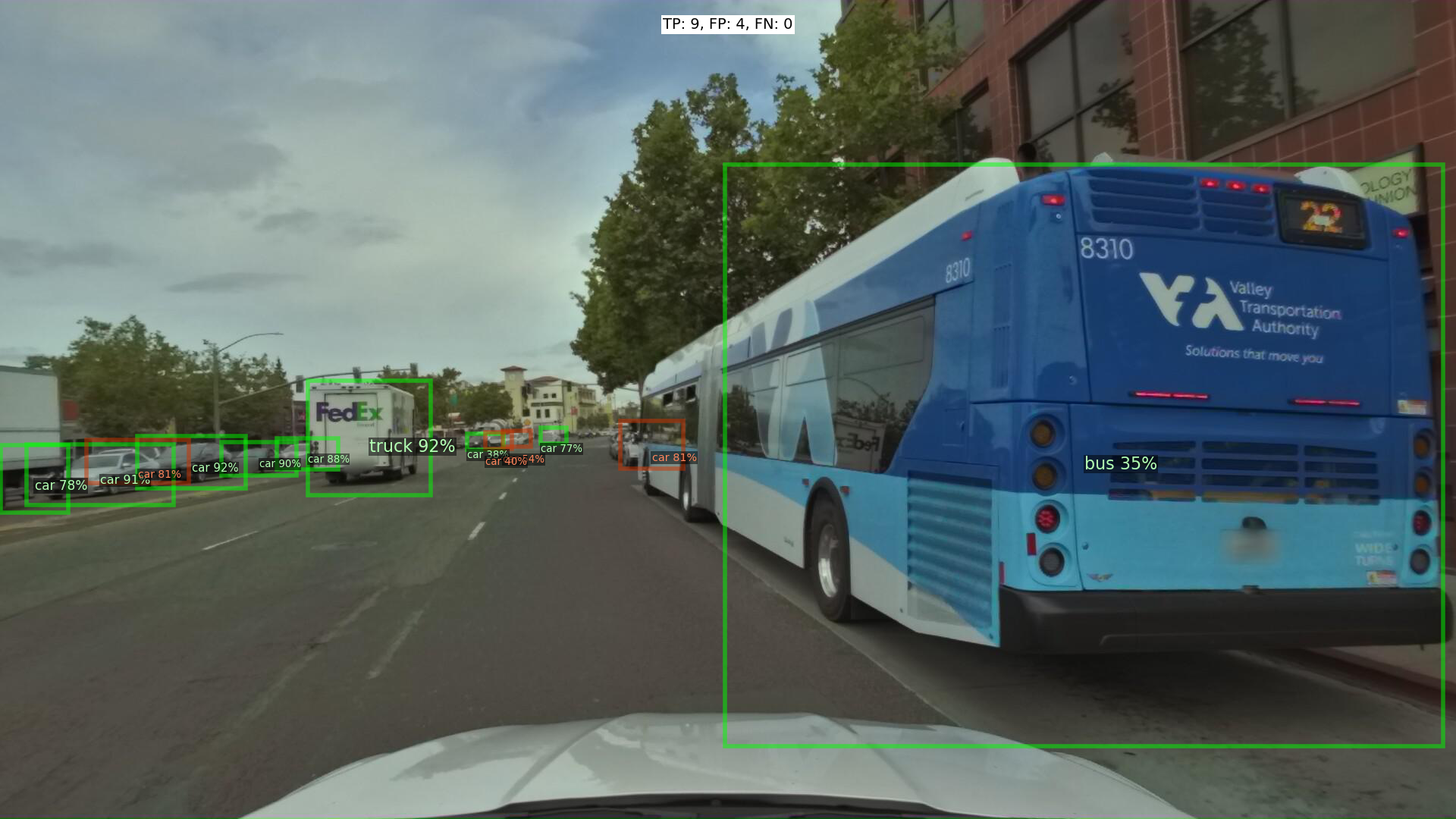}}&
  \frame{\includegraphics[width=.3\linewidth]{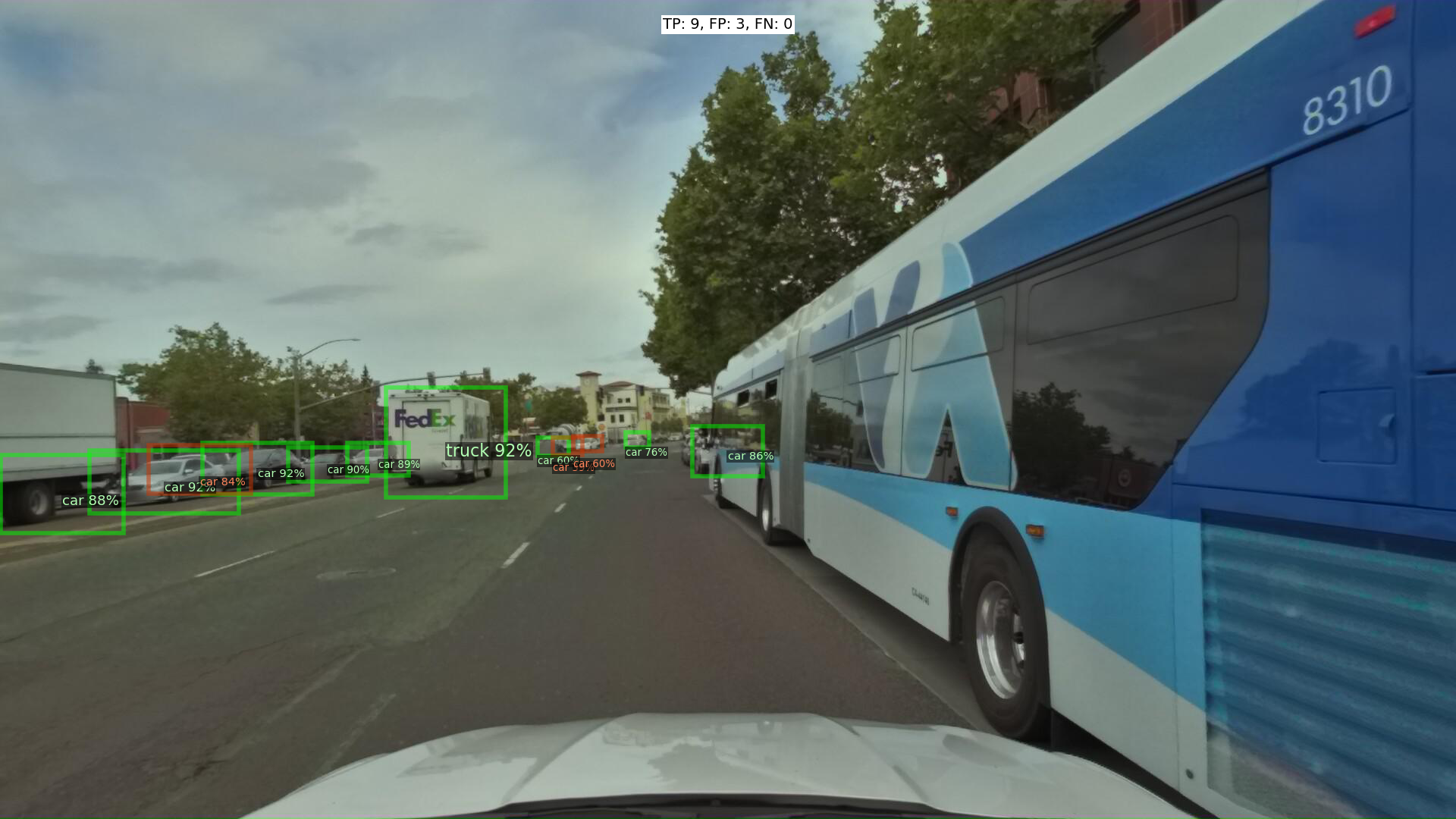}}\\[-1ex]
  % &\mycaption{0.2} & \mycaption{0.2} & \mycaption{0.3}\\
  \rowname{corr dt}&
  \frame{\includegraphics[width=.3\linewidth]{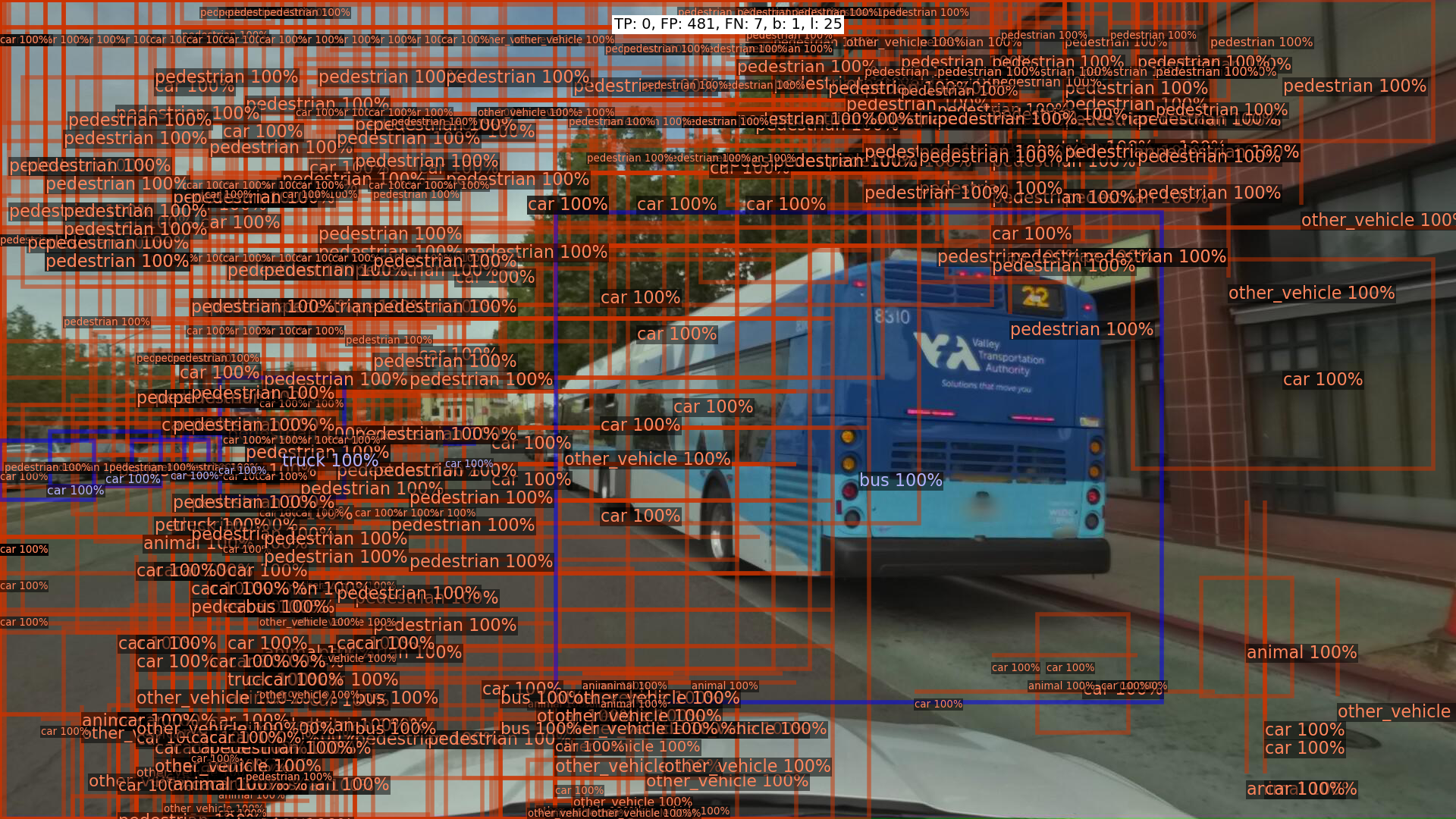}}&
  \frame{\includegraphics[width=.3\linewidth]{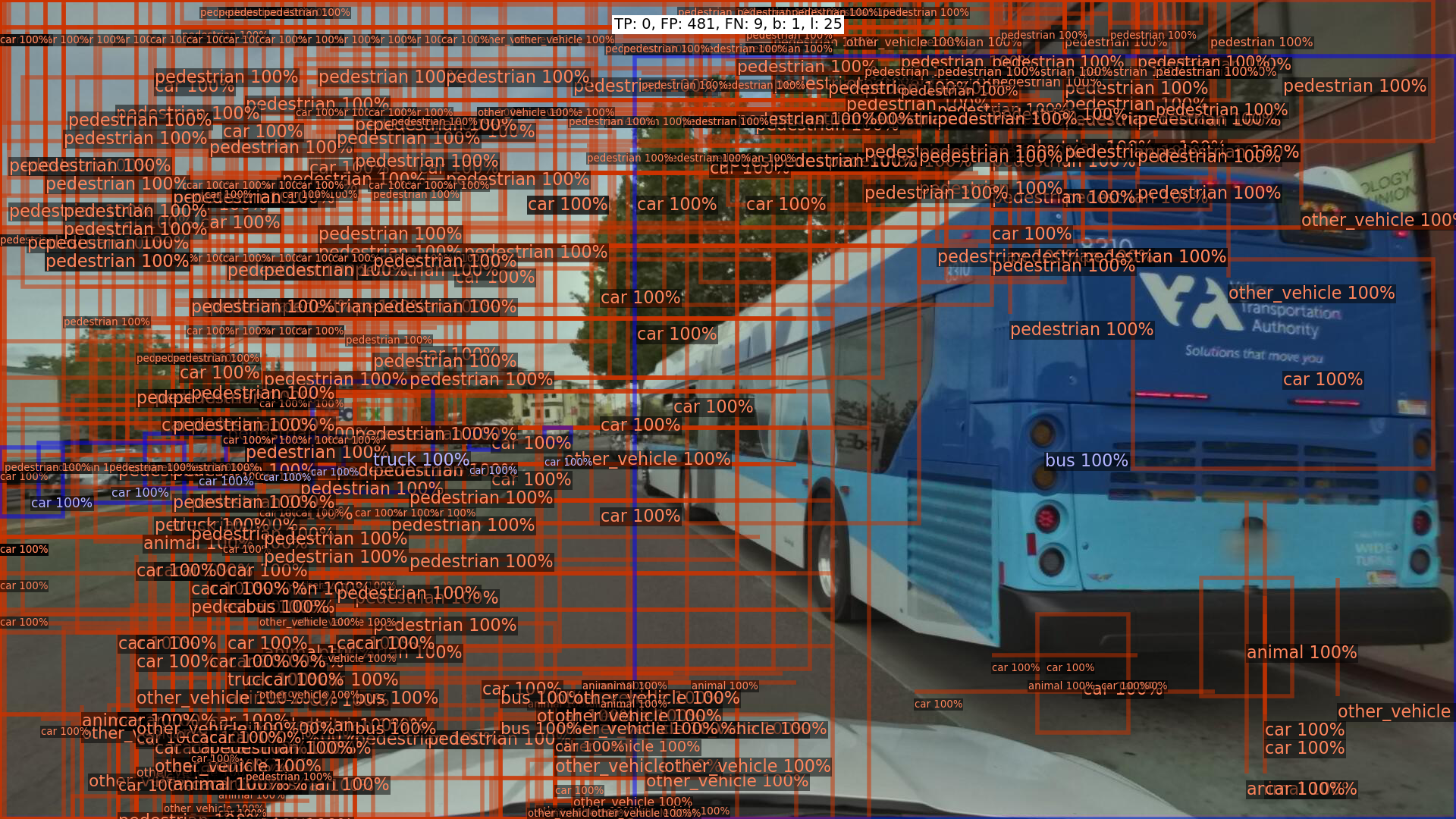}}&
  \frame{\includegraphics[width=.3\linewidth]{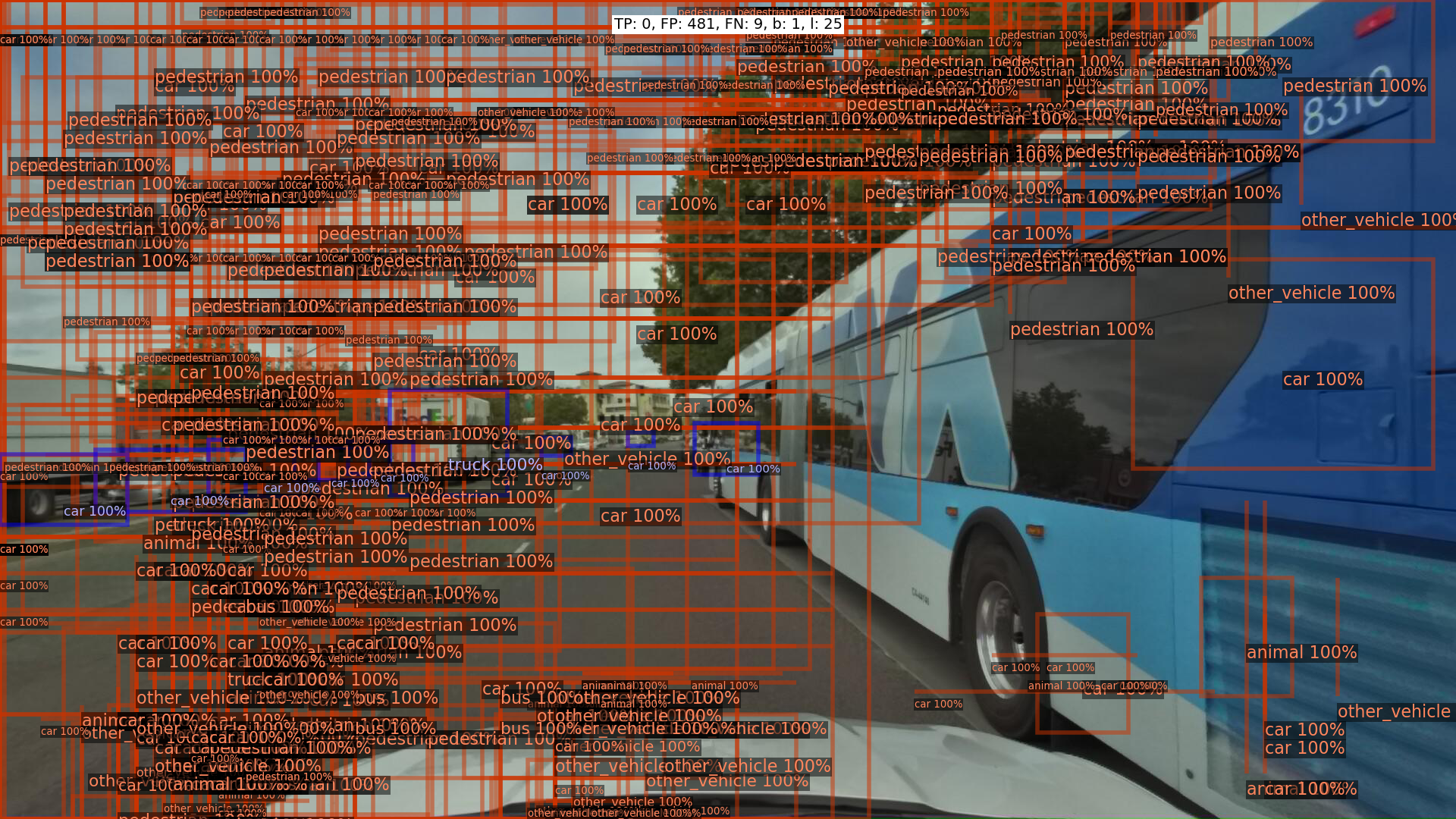}}\\[-1ex]
  % &\mycaption{0.5} & \mycaption{0.4} & \mycaption{0.6}\\
  \rowname{FP-blob}&
  \frame{\includegraphics[width=0.3\linewidth]{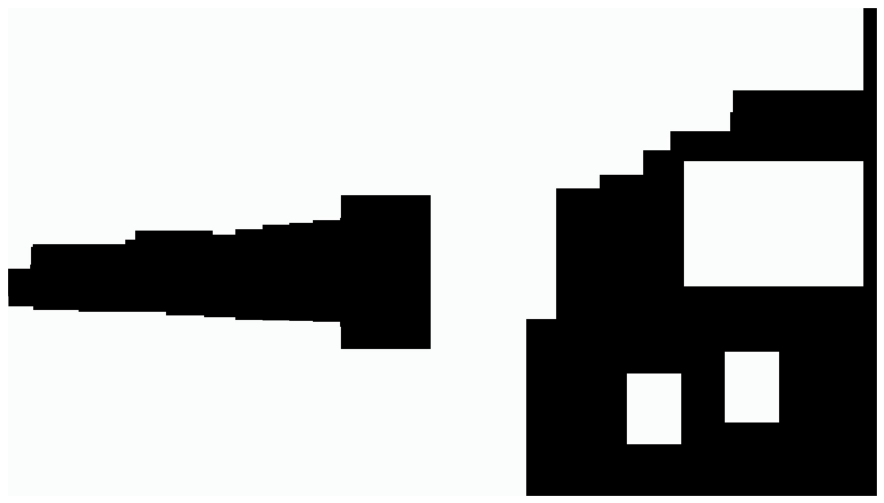}}&
  \frame{\includegraphics[width=0.3\linewidth]{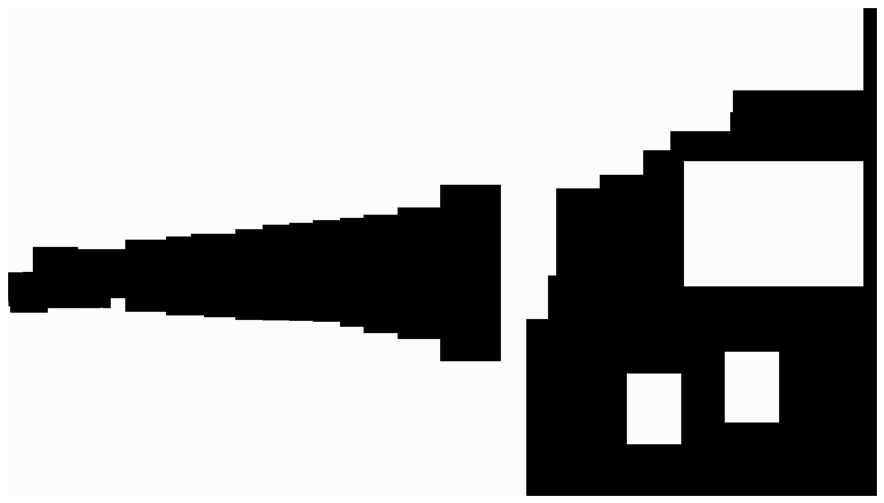}}&
  \frame{\includegraphics[width=0.3\linewidth]{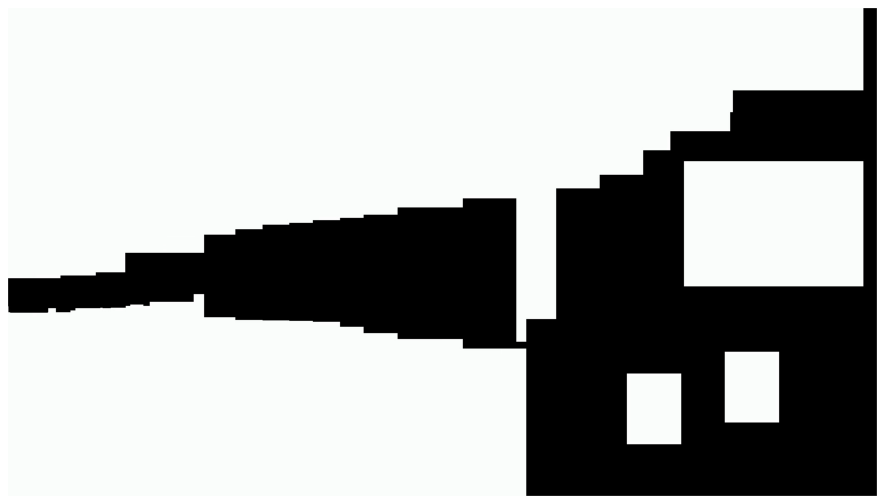}}\\[-1ex]
  % &\mycaption{0.5} & \mycaption{0.5} & \mycaption{0.7} \\

  \end{tabular}
  \vspace{-8pt}
  \caption{Pixel wise tracking of FP blobs. First row: orig dt are fault free detections. Second row: corr dt are faulty detections.
  Third row are tracked FP-blobs (white pixels are occupied by FP blobs).}%
  \label{fig:FP_blob_analysis: tracking}
  \vspace{-8pt}
\end{figure*}
\vspace{-8pt}

\begin{figure*}[!h]
  \settoheight{\tempdima}{\includegraphics[width=0.23\linewidth]{video_analysis/test_viz_0_dt.png}}%
  \centering\begin{tabular}{@{}c@{}c@{ }c@{ }c@{}}
  &\textbf{frame t=40} & \textbf{frame t=42} & \textbf{frame t=44} \\
  \rowname{orig dt}&
  \frame{\includegraphics[width=.3\linewidth]{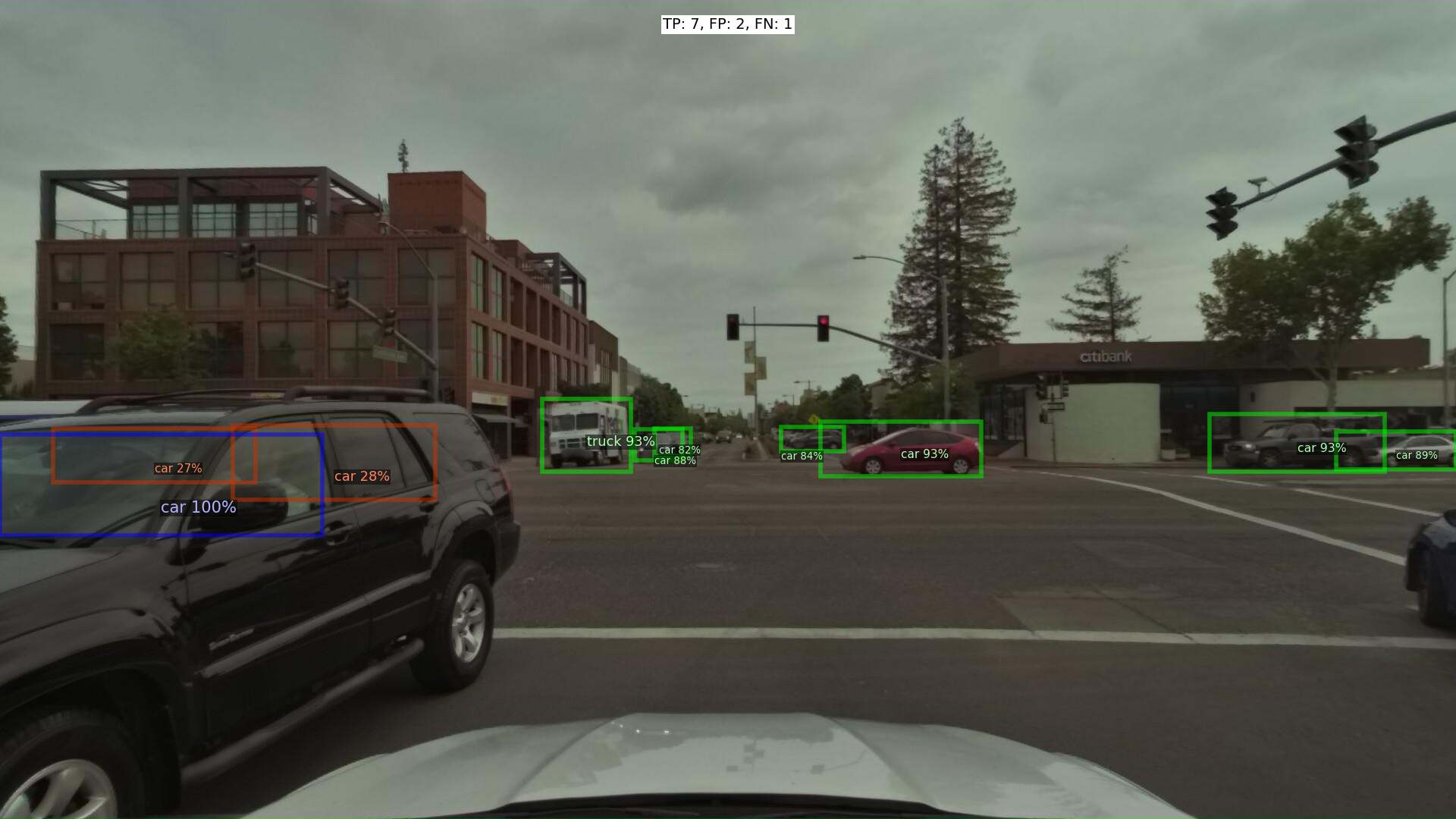}}&
  \frame{\includegraphics[width=.3\linewidth]{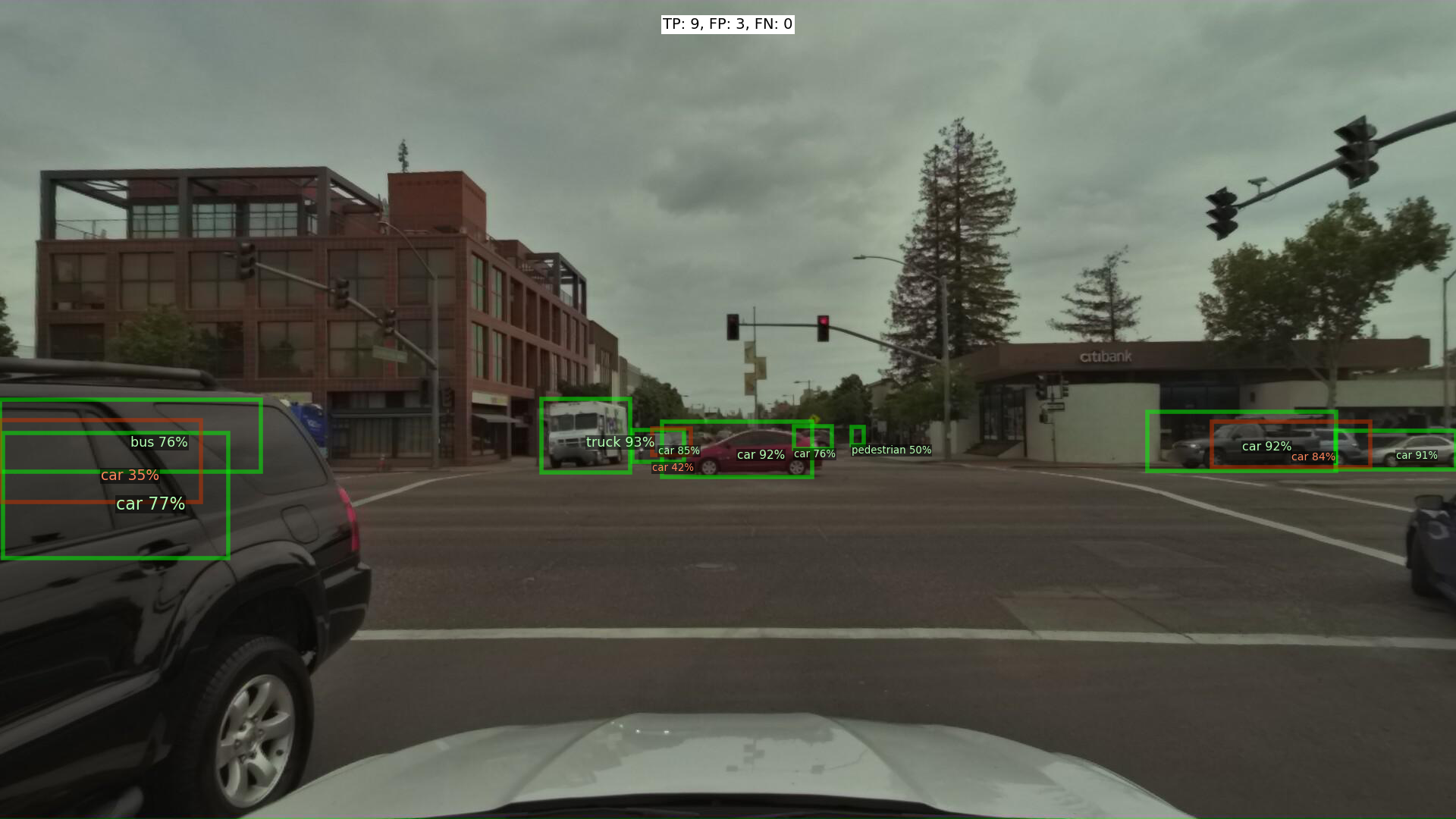}}&
  \frame{\includegraphics[width=.3\linewidth]{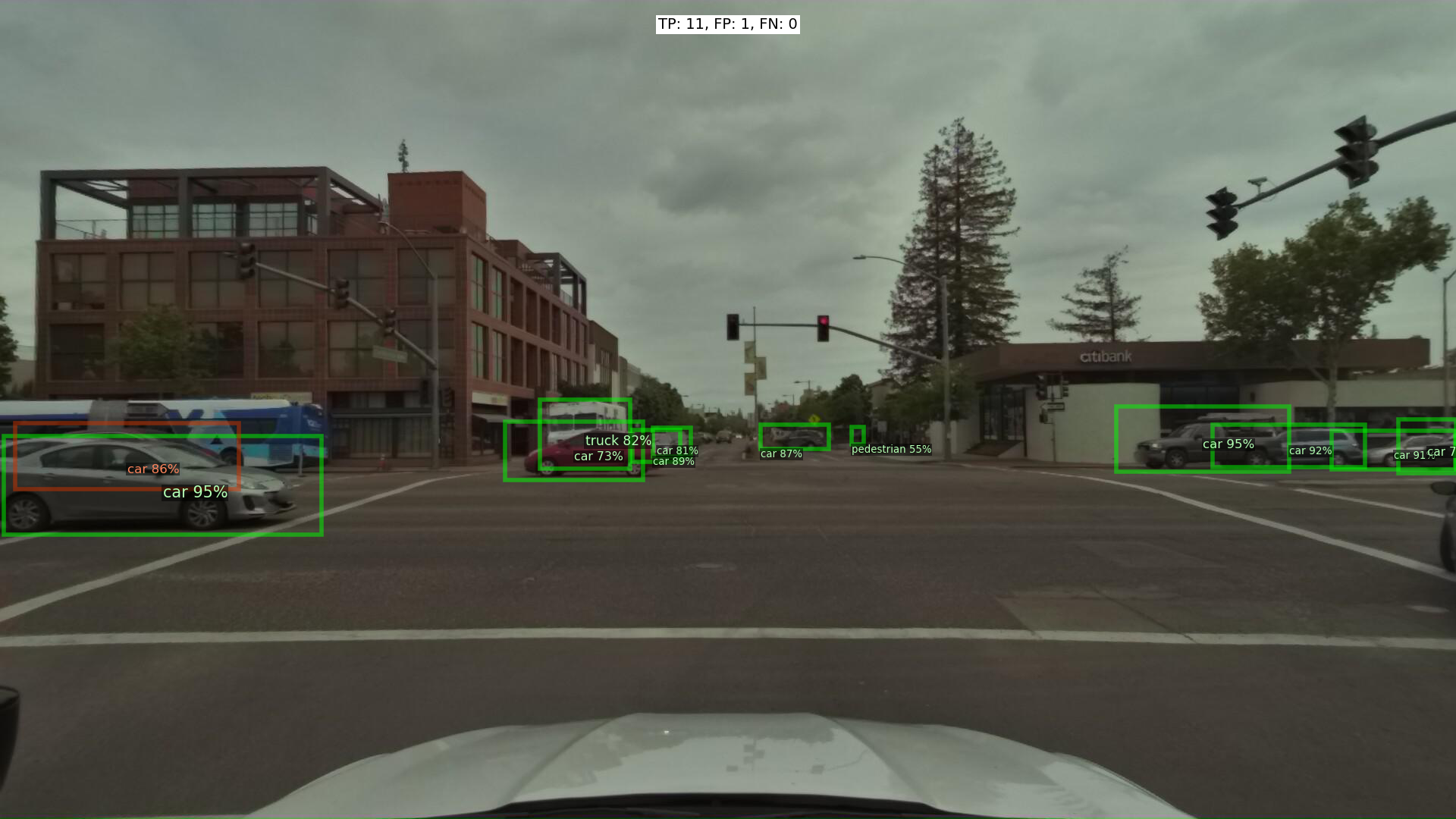}}\\[-1ex]
  % &\mycaption{0.2} & \mycaption{0.2} & \mycaption{0.3}\\
  \rowname{corr dt}&
  \frame{\includegraphics[width=.3\linewidth]{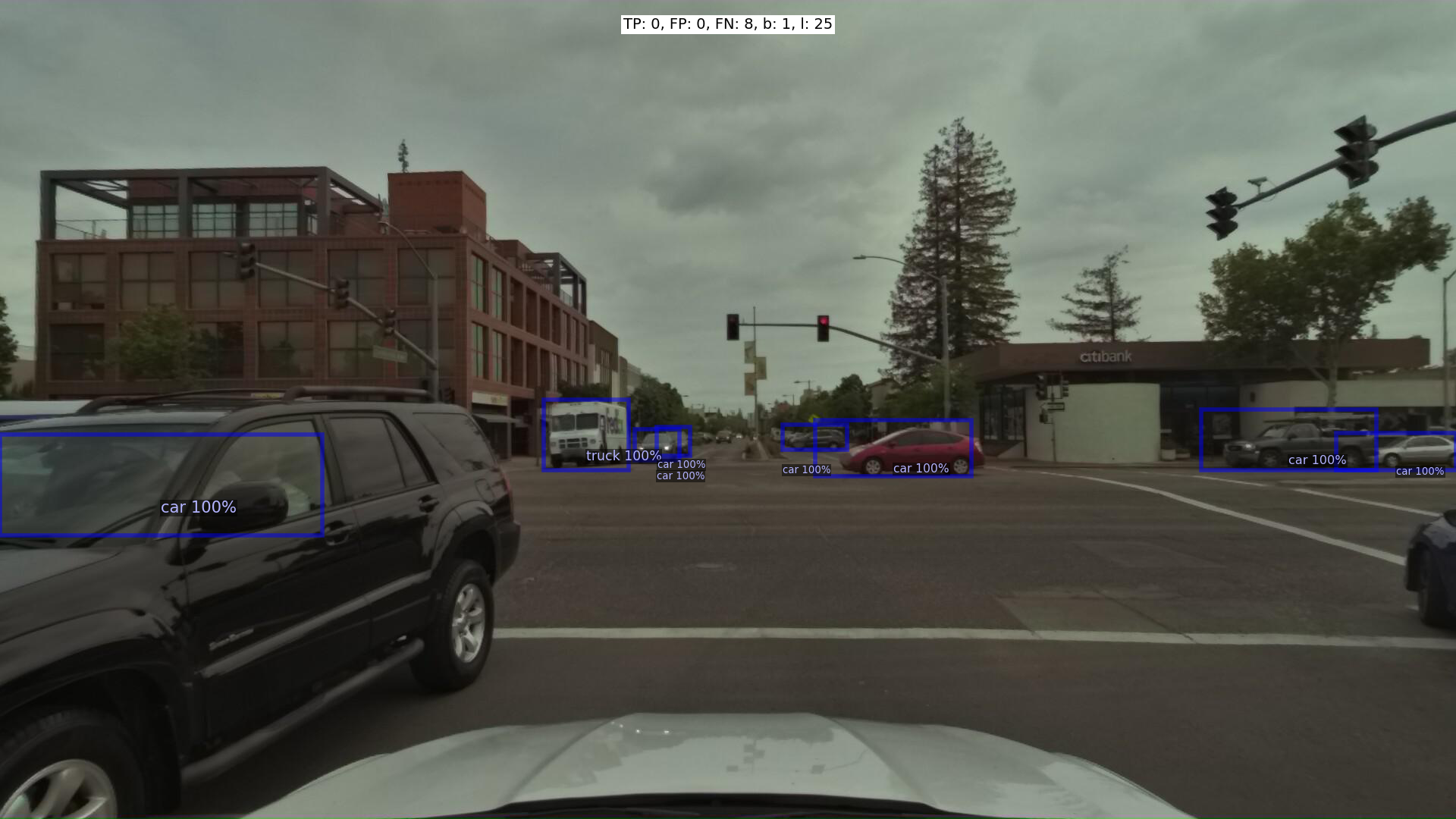}}&
  \frame{\includegraphics[width=.3\linewidth]{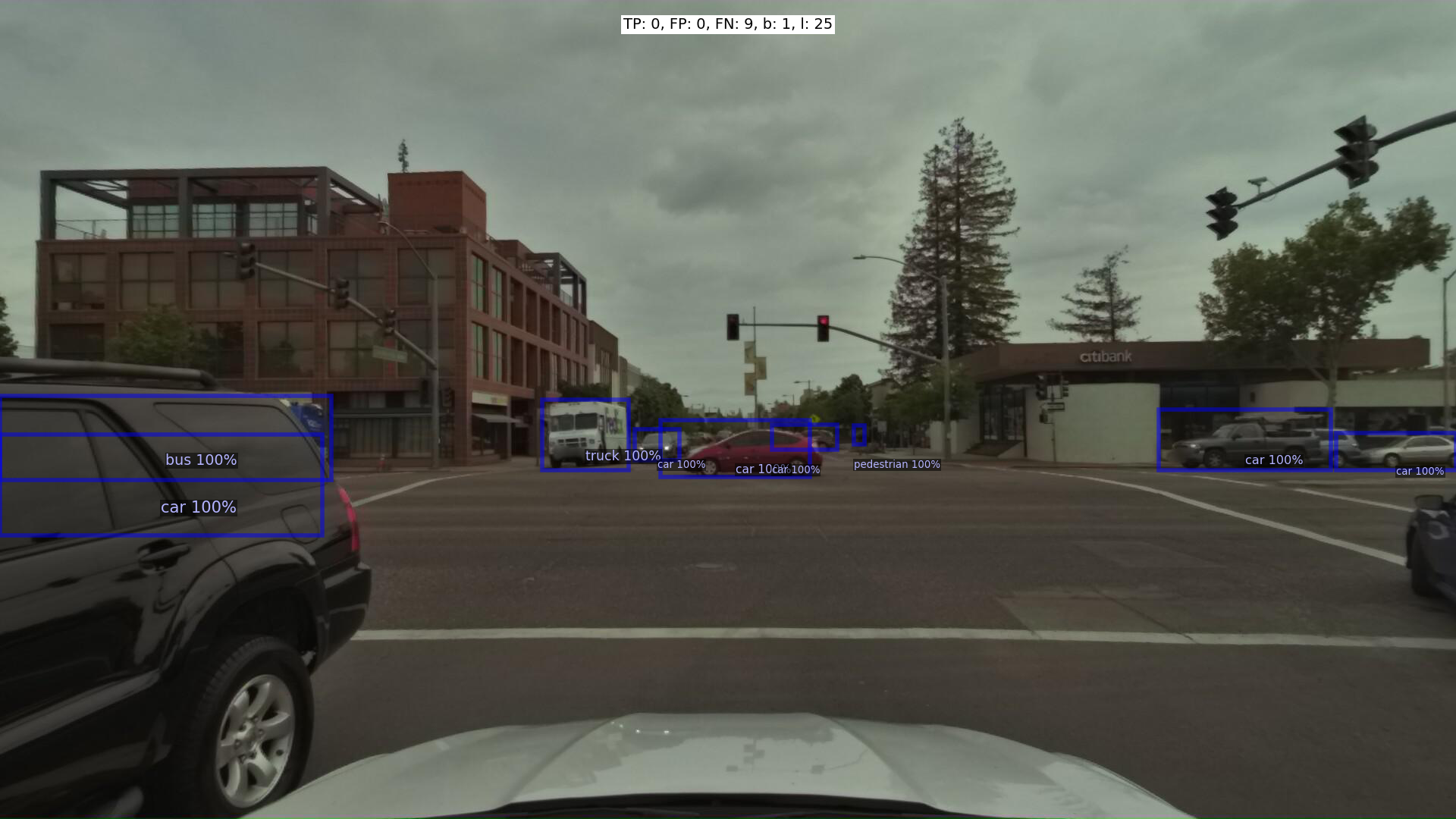}}&
  \frame{\includegraphics[width=.3\linewidth]{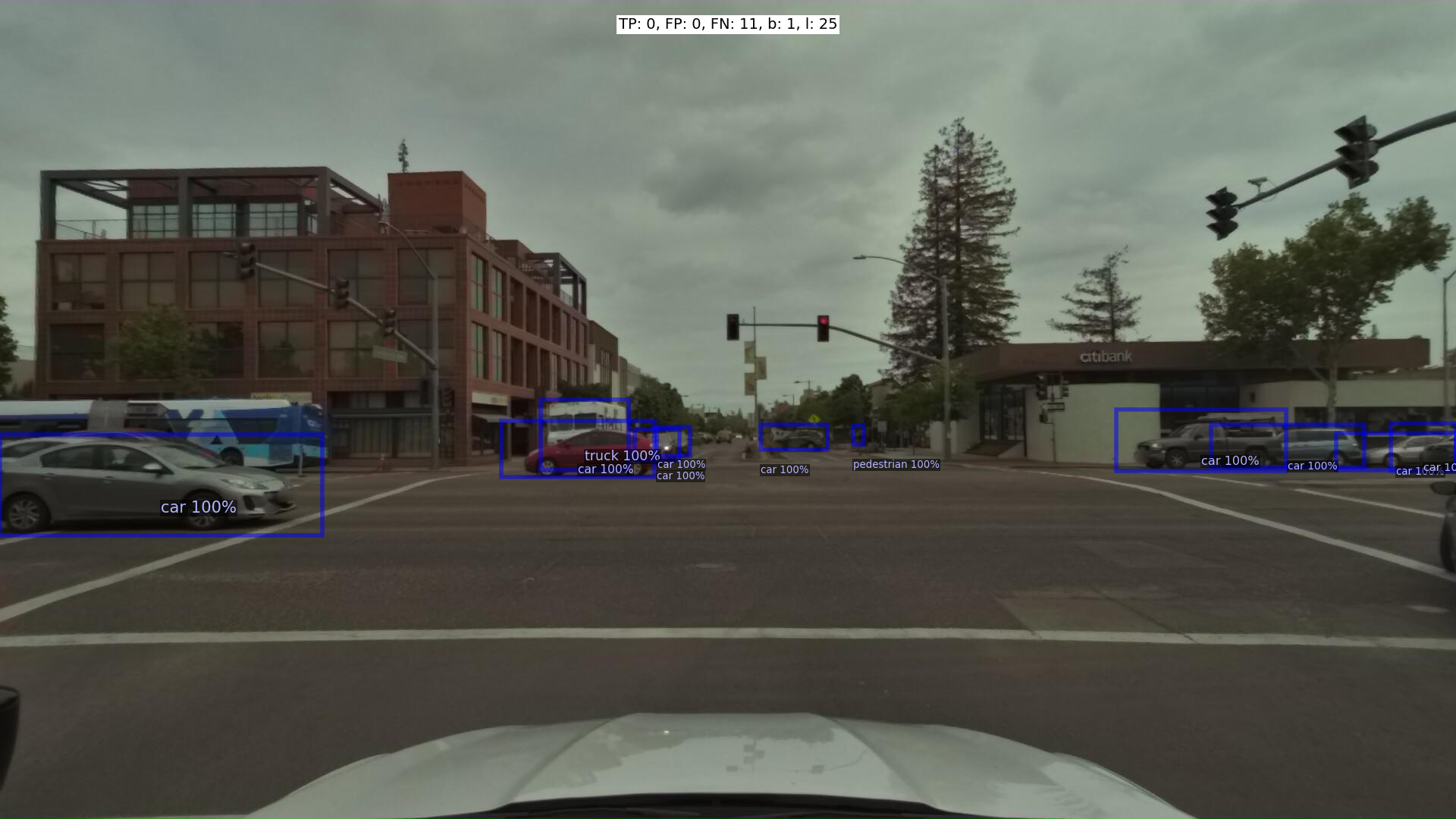}}\\[-1ex]
  % &\mycaption{0.5} & \mycaption{0.4} & \mycaption{0.6}\\
  \rowname{FN-blob}&
  \frame{\includegraphics[width=0.3\linewidth]{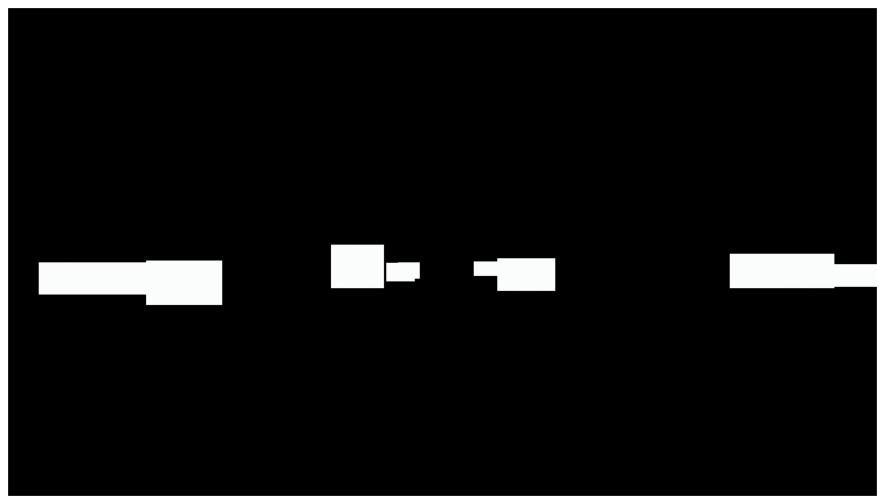}}&
  \frame{\includegraphics[width=0.3\linewidth]{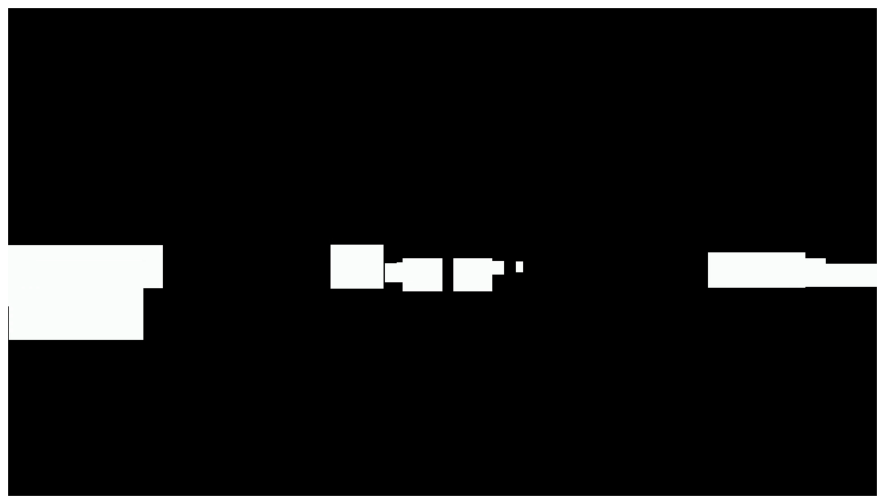}}&
  \frame{\includegraphics[width=0.3\linewidth]{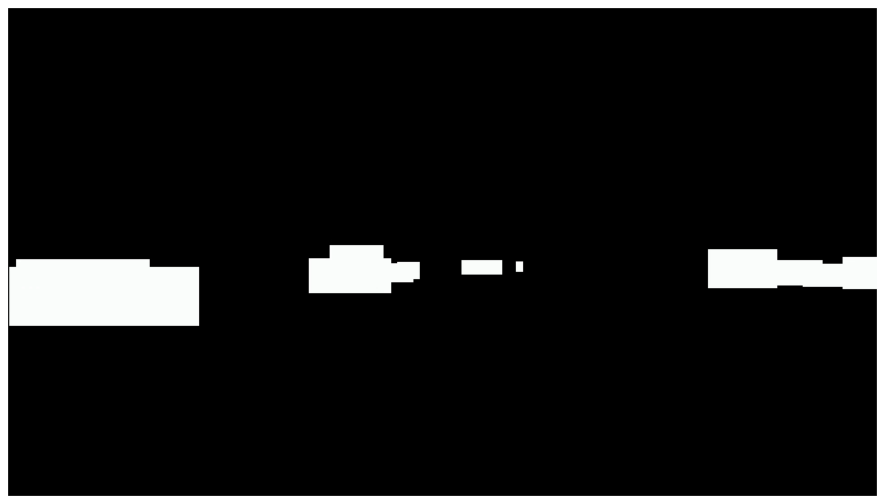}}\\[-1ex]
  % &\mycaption{0.5} & \mycaption{0.5} & \mycaption{0.7} \\

  \end{tabular}
  \vspace{-8pt}
  \caption{Pixel wise tracking of FN blobs. First row: orig dt are fault free detections. Second row: corr dt are faulty detections.
  Third row are tracked FN-blobs (white pixels is the free space created by FNs).}%
  \label{fig:FN_blob_analysis: tracking}
  \vspace{-8pt}
\end{figure*}
\vspace{-8pt}

\begin{figure*}[t]
  \begin{subfigure}{0.45\textwidth}
  \centering
  \includegraphics[width=\linewidth]{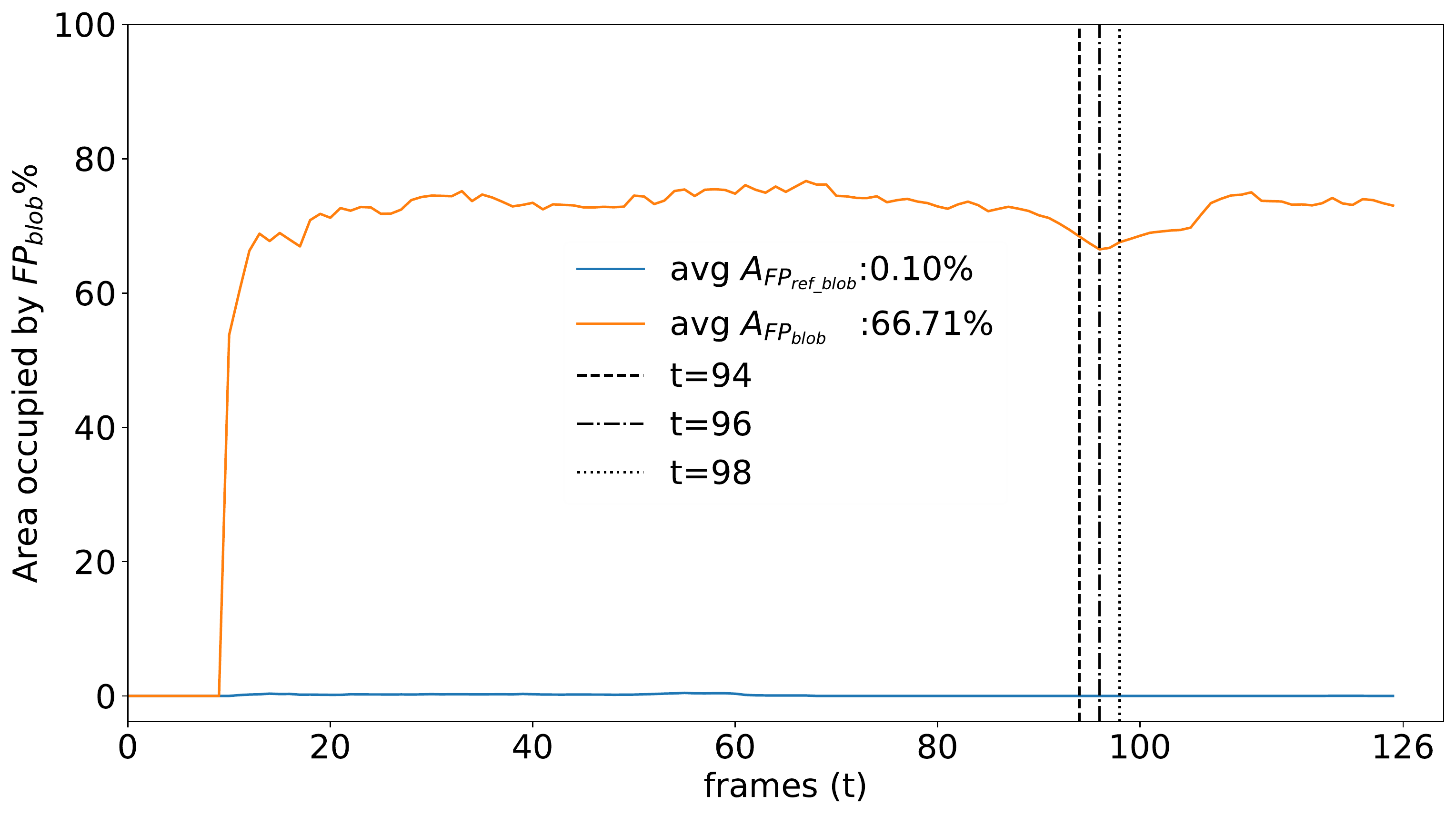}
  %\caption{A mouse}\label{fig:mouse}
  \caption{Average area occupancy by FP-blob}
  \label{fig:FP_blob_analysis: explanation} 
  \end{subfigure} %
  \qquad
  \begin{subfigure}{0.45\textwidth}
  \centering
  \includegraphics[width=\linewidth]{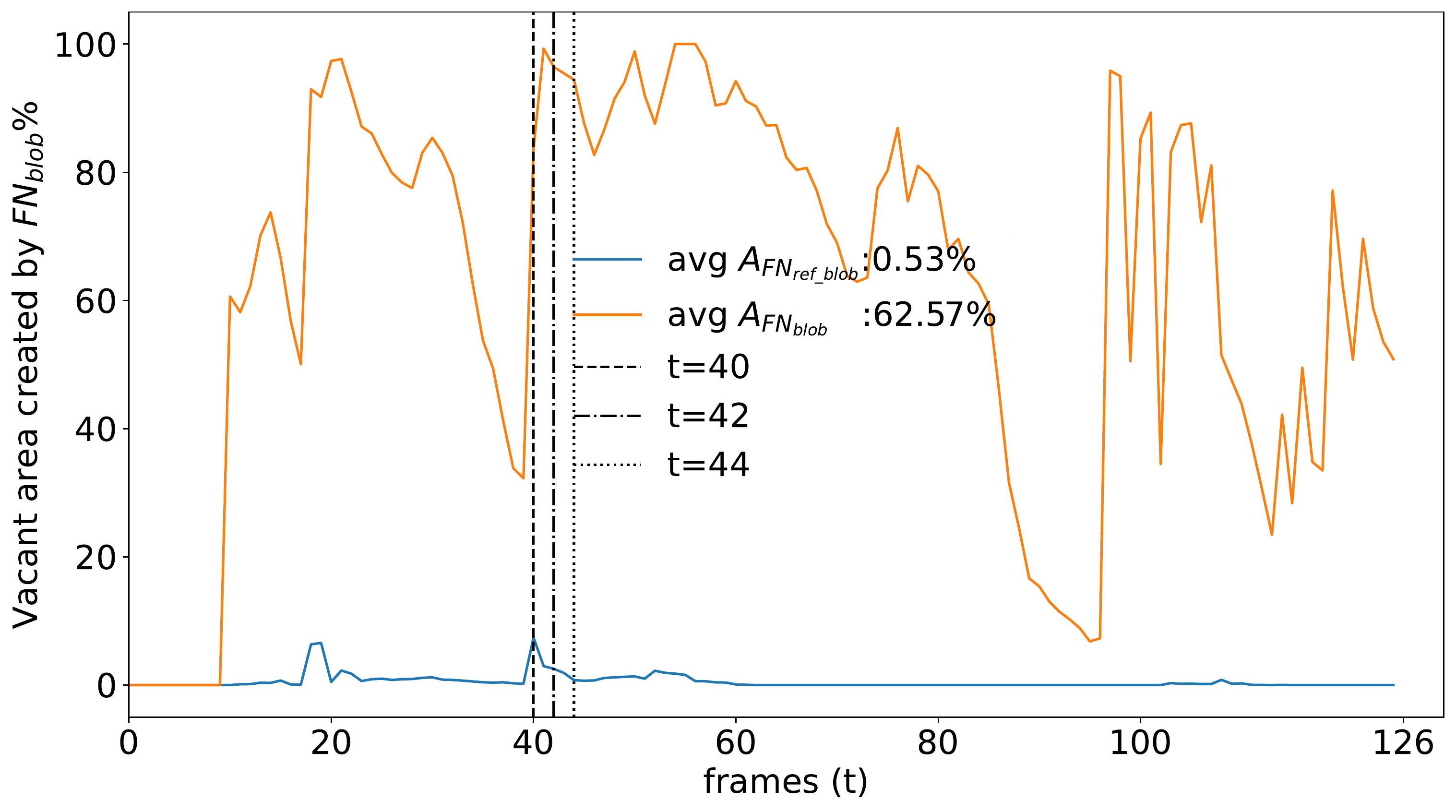}
    %\caption{A mouse}\label{fig:mouse}
  \caption{Average area occupancy by FN-blob}
  \label{fig:FN_blob_analysis: explanation}
  \end{subfigure}
  \vspace{-8pt}
\caption{Tracking of FP- and FN-blob area}
\label{fig.Tracked area explaination}
%   \label{fig:FP_blob_analysis: explanation}
% \vspace{-5pt}
\end{figure*}
\vspace{-8pt}

\begin{figure}[!h]
\begin{subfigure}{0.32\textwidth}
  \centering % <-- added
  \includegraphics[width=\linewidth, height=3.5cm]{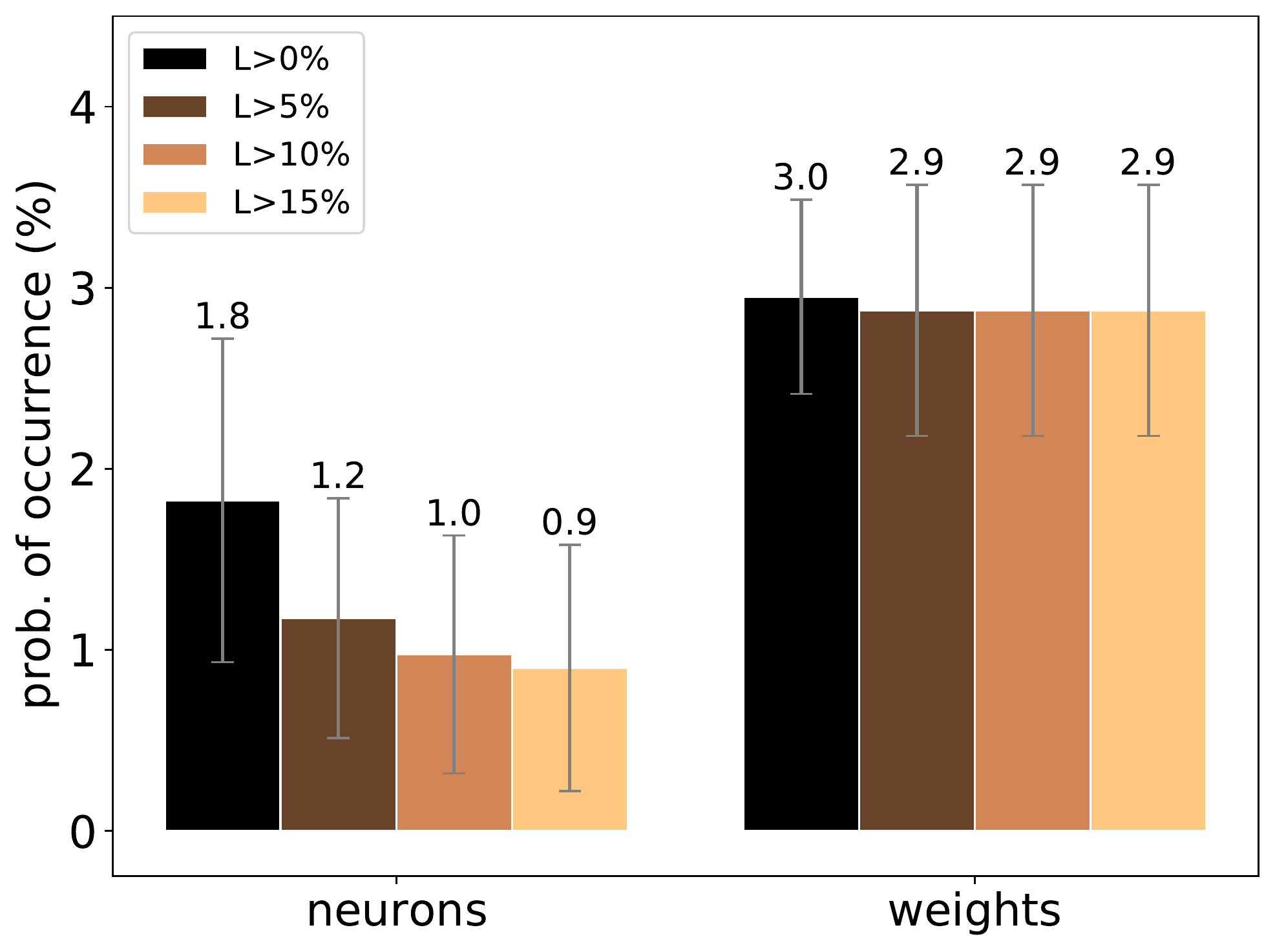}
  \caption{FP}
  \label{fig:1}
\end{subfigure}%\hfil % <-- added
\begin{subfigure}{0.32\textwidth}
  \centering % <-- added
  \includegraphics[width=\linewidth, height=3.5cm]{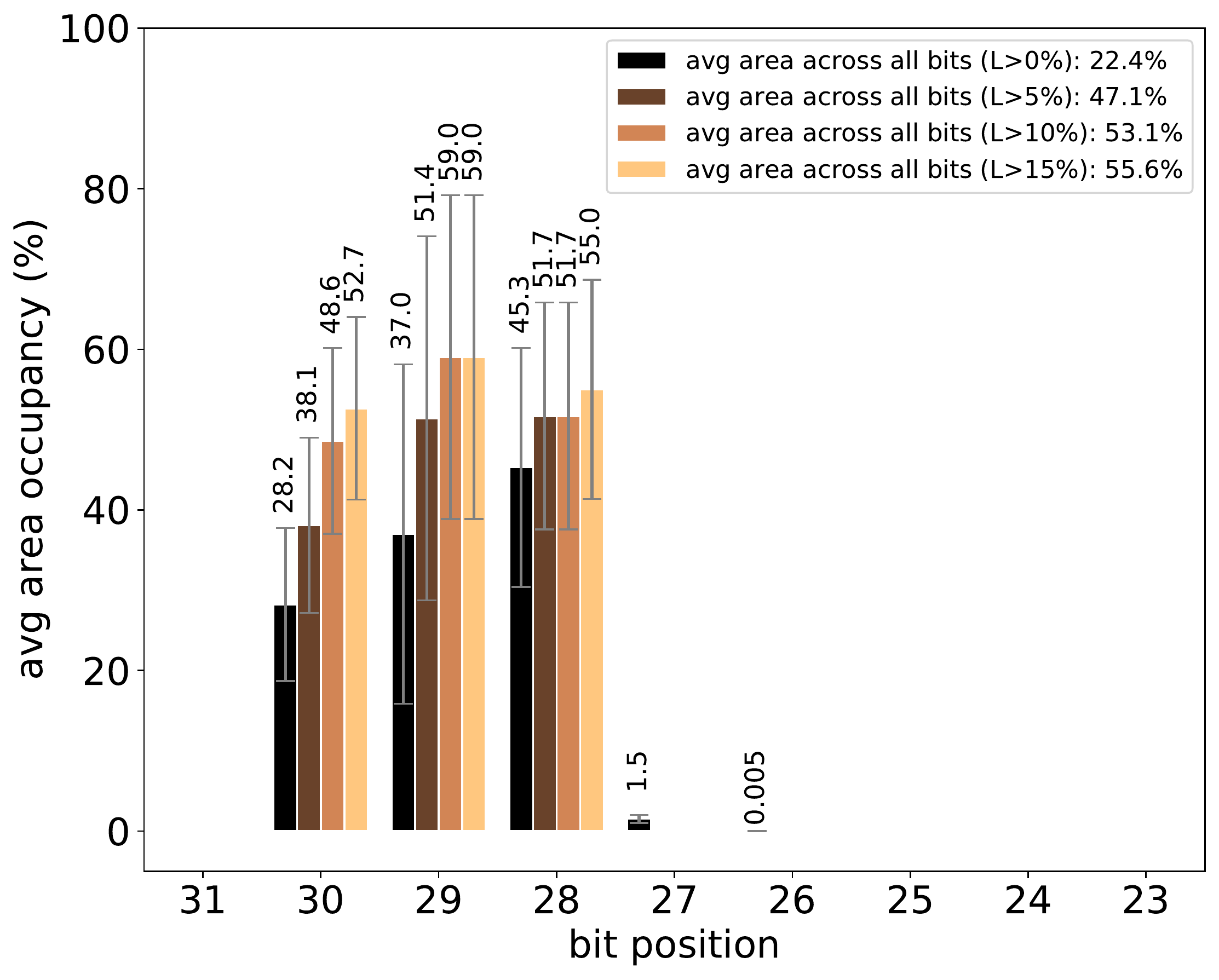}
  \caption{FP Neurons}
  \label{fig:2}
\end{subfigure}%\hfil % <-- added(\)  
\begin{subfigure}{0.32\textwidth}
  \centering % <-- added
  \includegraphics[width=\linewidth, height=3.5cm]{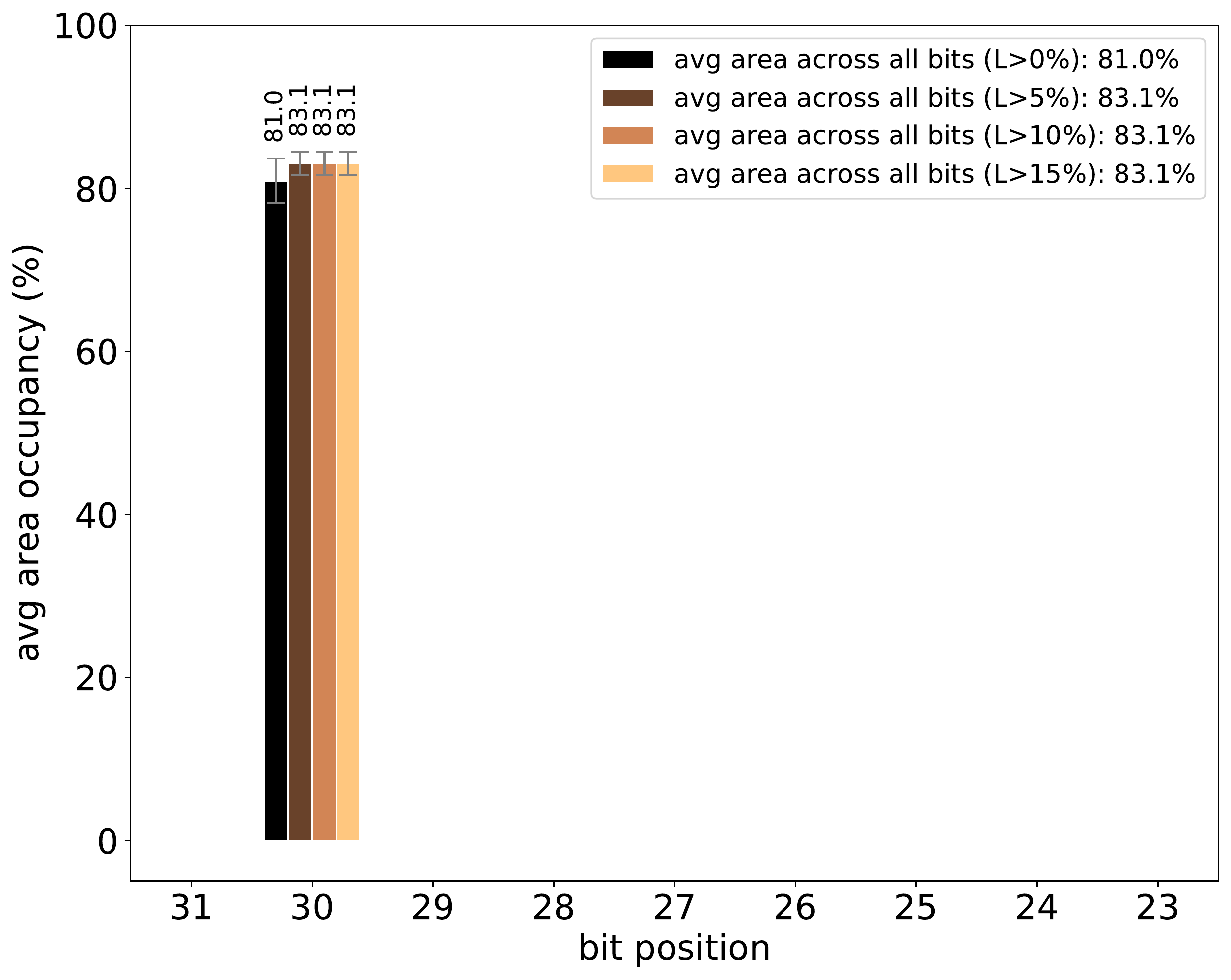}
  \caption{FP Weights}
  \label{fig:3}
\end{subfigure}
\medskip

\begin{subfigure}{0.32\textwidth}
  \centering % <-- added
  \includegraphics[width=\linewidth, height=3.5cm]{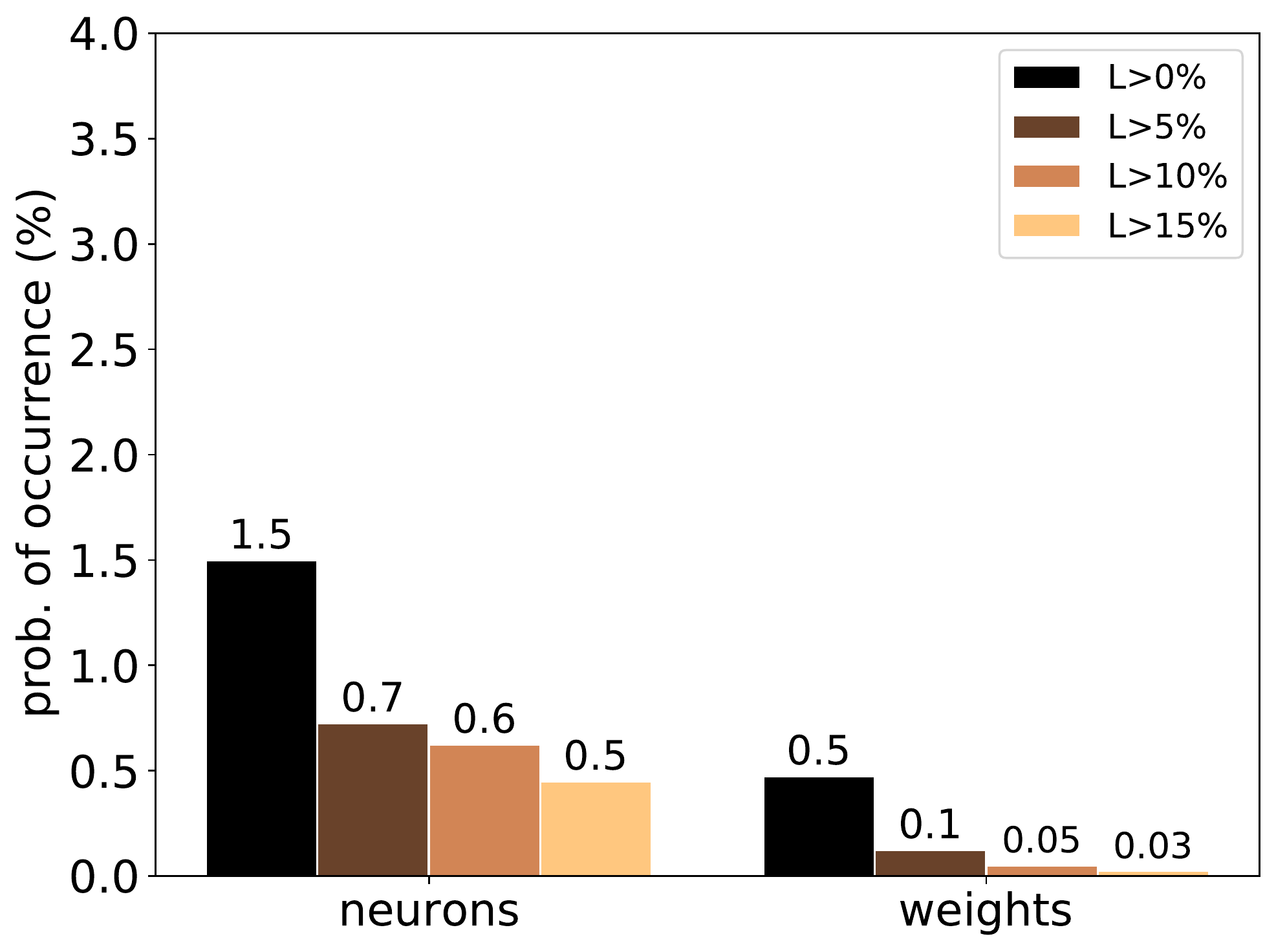}
  \caption{FN}
  \label{fig:4}
\end{subfigure}
\begin{subfigure}{0.32\textwidth}
  \centering % <-- added
  \includegraphics[width=\linewidth, height=3.5cm]{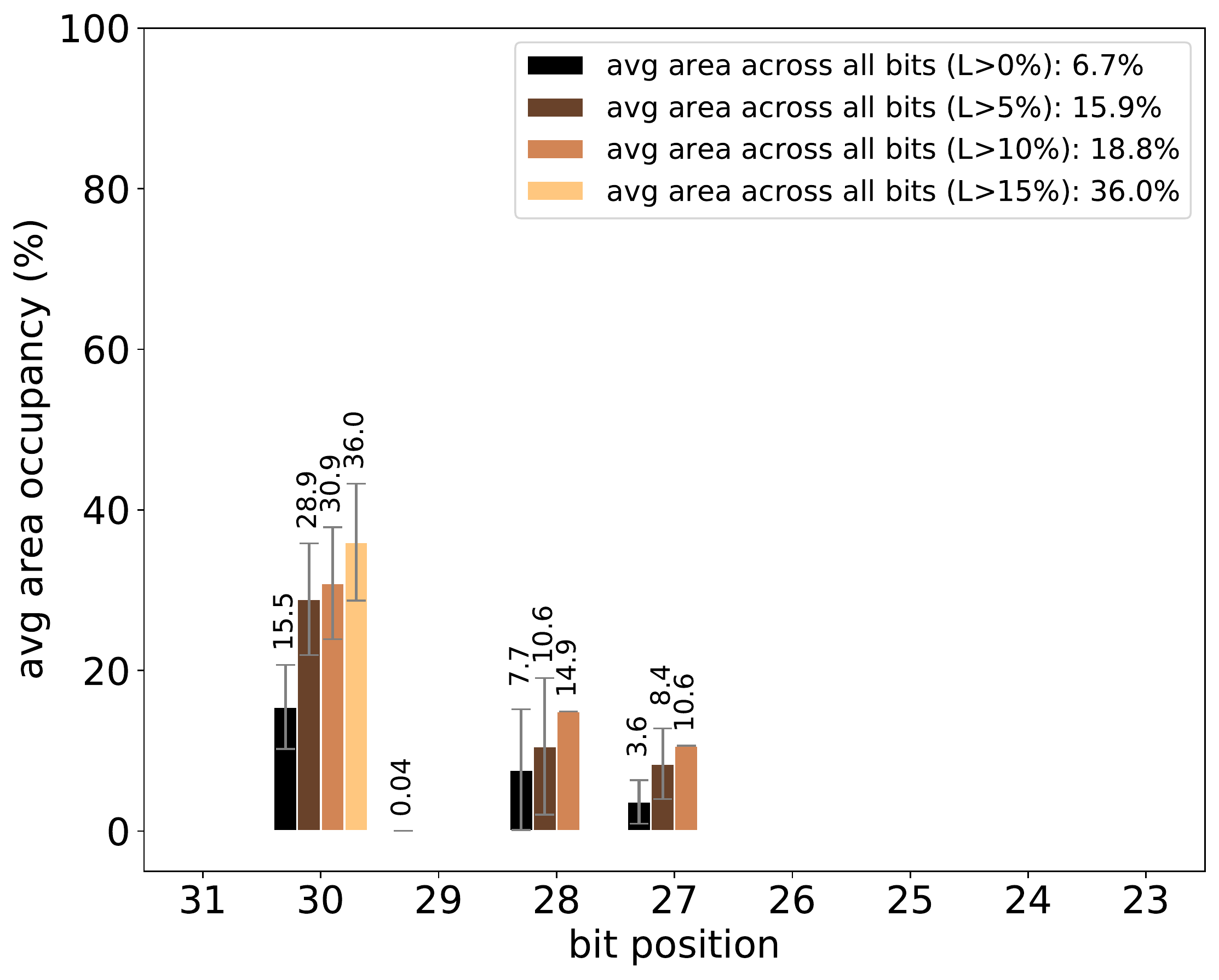}
  \caption{FN Neurons}
  \label{fig:5}
\end{subfigure}%\hfil % <-- added
\begin{subfigure}{0.32\textwidth}
  \centering % <-- added
  \includegraphics[width=\linewidth, height=3.5cm]{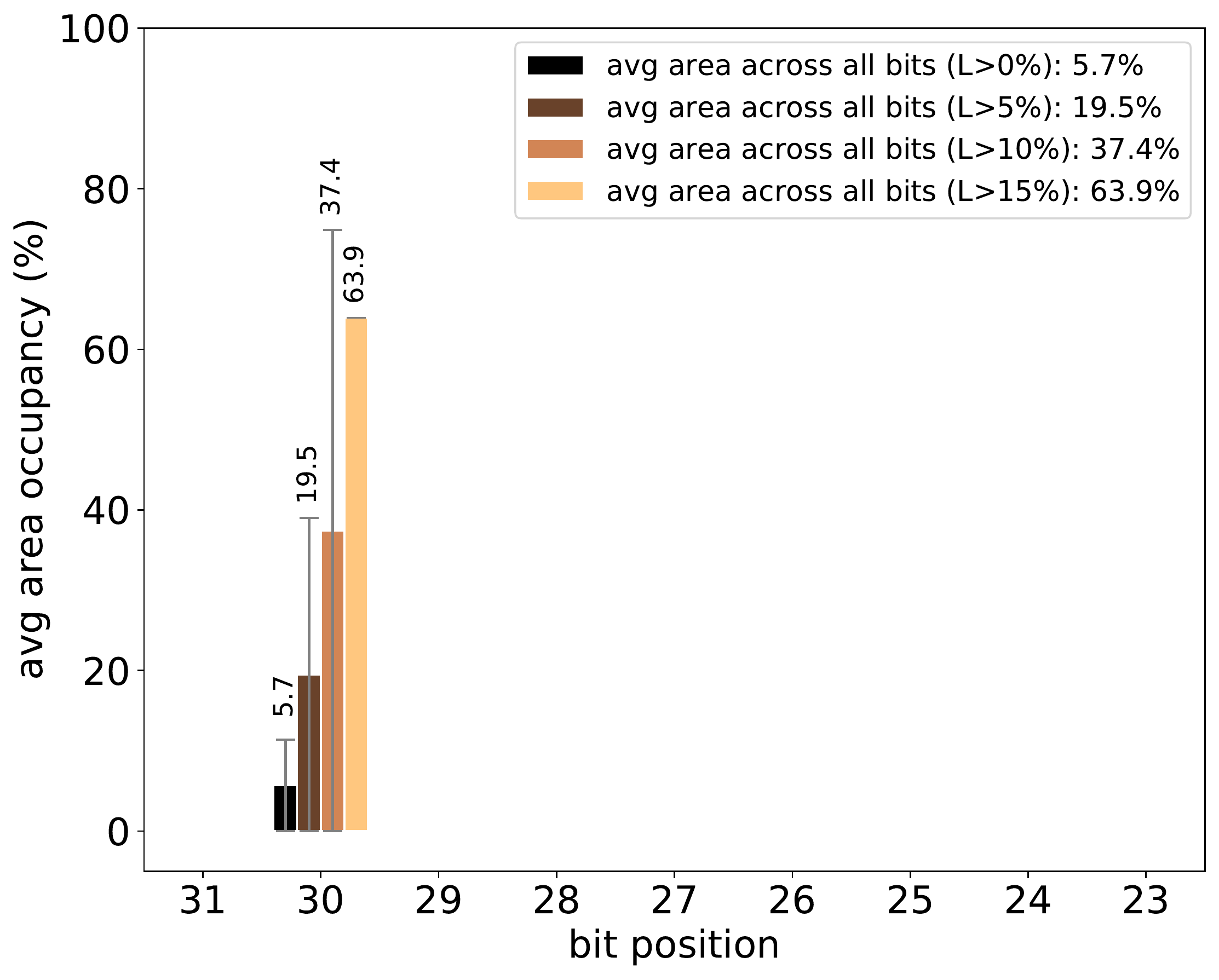}
  \caption{FN Weights}
  \label{fig:6}
\end{subfigure}
\vspace{-8pt}
\caption{Vulnerability of Yolov3 and Lyft for permanent faults.}
\label{fig:permanent_faults_stats}
\vspace{-8pt}
\end{figure}
\vspace{-15pt}
% \begin{figure*}[!h]
% \centering % <-- added
% \begin{subfigure}{0.32\textwidth}
%   \includegraphics[width=\linewidth]{video_analysis/probability_FP.pdf}
%   \caption{FP}
%   \label{fig:1}
% \end{subfigure}%\hfil % <-- added
% \begin{subfigure}{0.32\textwidth}
%   \includegraphics[width=\linewidth]{video_analysis/probability_FN.pdf}
%   \caption{FN}
%   \label{fig:4}
% \end{subfigure}\\

% \begin{subfigure}{0.32\textwidth}
%   \includegraphics[width=\linewidth]{video_analysis/FP_area_average_neurons.pdf}
%   \caption{FP Neurons}
%   \label{fig:2}
% \end{subfigure}%\hfil % <-- added
% \begin{subfigure}{0.32\textwidth}
%   \includegraphics[width=\linewidth]{video_analysis/FP_area_average_weights.pdf}
%   \caption{FP Weights}
%   \label{fig:3}
% \end{subfigure}
% \begin{subfigure}{0.32\textwidth}
%   \includegraphics[width=\linewidth]{video_analysis/FN_area_average_neurons.pdf}
%   \caption{FN Neurons}
%   \label{fig:5}
% \end{subfigure}%\hfil % <-- added
% \begin{subfigure}{0.32\textwidth}
%   \includegraphics[width=\linewidth]{video_analysis/FN_area_average_weights.pdf}
%   \caption{FN Weights}
%   \label{fig:6}
% \end{subfigure}
% \caption{Vulnerability of Yolov3 and Kitti for permanent faults.}
% \label{fig:permanent_faults_stats}
% \end{figure*}

%\section{Results \romannum{2}: Permanent faults effect on DNN}
\vspace{5pt}
\section{Permanent faults}
\label{sec:permanent_faults}
Our analysis in this section aims to understand whether permanent stuck-at faults (see Sec.~\ref{sec:hardware faults vocabulary}) leads to temporally consistent errors on an object level leading to continuous failure.
The object detection model typically receives sequential images from a continuous video stream in real time applications. 
We assume a permanent hardware fault hitting the inference module which in turn causes persistent miss detections on consecutive images.
% Suppose a permanent hardware fault causes persistent miss detections on consecutive images.
In this case, they will appear either as ghost objects in the output (as FPs) or lead to a consecutive miss of an object (as FNs) - both situations can be highly safety-critical.
A perception pipeline typically also includes a tracking module for detected objects, which can then be used to predict an object's trajectory and make an informed decision concerning the next maneuver of the vehicle. Therefore, we simulate a simple tracking of instantaneous fault-induced FPs and FNs clusters to determine whether they would be persistent in a realistic scenario.
For the analysis in this section, we use Yolov3 and the Lyft data set. This is the only dataset used in our analysis that provides consecutive images from video sequences (Lyft sequence of the \textit{CAM\_FRONT} channel featuring $126$ frames is considered).
% We inject 1000 single permanent faults at neurons and weights independently for this sequence.
From our experiments with transient faults injections in Sec.~\ref{sec:transient_faults}, we understand that no effect is observed by altering mantissa bits or by flips in the direction $'1' \to\ '0'$ since this does not generate large values.
Therefore, the experiments of this section are accelerated by using only \text{\statone} faults in the exponential bits of FP32. However, results have been rescaled to account for the probability of injections in all $32$ bits.
In this section, we designed an experiment where we inject each of 1000 single random permanent faults (exponential bits) at neurons and weights independently for the above considered sequence to understand its safety impact.

\vspace{-12pt}
\subsection{Evaluating fault persistence}
\vspace{-8pt}

We track the movement of blobs (Eq. \ref{eq:blob corr}) using a simple pixel-wise M/N tracking scheme \cite{Blackman1999}.
The proposed tracker incorporates the following criteria to establish that a given pixel of FP or FN blob is persistent, at a given frame $t$: i) The pixel occupied in at least M/N consecutive frames. (if it is also occupied in the current frame, this corresponds to t track update; otherwise it is a coasting track), ii) If the occupancy of that pixel in the last N frames is below M, we check the vicinity around that pixel for past occupancy. Deploying a simplified unidirectional motion model, we register a persistent dynamic pixel for the current frame if occupancies above M are found in the past $N$ frames in a close enough (here $50$ pixel, abbr. px) vicinity.\\
For FN blobs, we omit coasting due to the nature of detection misses. After registering the persistent pixels computed by the pixel-wise tracker, the occupied ($\Aoccfp$) or free-space ($\Aoccfn$) area is calculated using Eq. \ref{eq:area_blob corr}.
The tracking parameters are chosen as $(10/15)$: The upper frame number is hereby estimated from a critical time of reaction to a persistent false target ($\approx0.5\text{s}$) and the frame rate of the Lyft sequence ($30\text{Hz}$), leading to $N=0.5\text{s}\cdot 30\text{s}^{-1}=15$ key sequential frames. This estimated upper number can be application specific relevant to its safety specifications.

\vspace{-8pt}
\subsection{Corruption probability and severity}
\vspace{-5pt}
In Fig.~\ref{fig:FP_blob_analysis: tracking} and Fig.~\ref{fig:FN_blob_analysis: tracking}, we show examples of persistent FP and FN blobs in selected frames. The  occupied ($\Aoccfp$) and free ($\Aoccfn$) space of an entire video sequence is presented in Fig.~\ref{fig:FP_blob_analysis: explanation} and Fig.~\ref{fig:FN_blob_analysis: explanation}.
For orientation, we also give the area difference between original and ground truth predictions (Fig.~\ref{fig.Tracked area explaination}), $ \Aoccfporig = |\mathcal{I}(\text{det}_\text{orig} - \text{gt})|/ A_{\text{image}}$ and $ \Aoccfnorig = |\mathcal{I}(\text{gt} - \text{det}_\text{orig})|/ |\mathcal{I}(\text{det}_\text{orig})|$ (where \text{gt} is ground truth).
We neglect these contributions originating from the model imperfection as it is a function of training and is found to be small  (in the above examples $< 1\%$) compared to the fault-induced occupancy ($\sim66\%$ and $\sim62\%$, respectively). The example demonstrates that tracked FP blobs may persist across the entire image sequence and occupy a significant amount of free space. Similarly, a significant portion of the image can be lost persistently across the sequence (it reaches as high as  $\sim96\%$).
Our statistical evaluation from $1000$ permanent fault injections on the selected image sequence is given in Fig.~\ref{fig:permanent_faults_stats} for FP and FN.
% We discuss both the permanent SDC probability (Fig.~\ref{fig:permanent_faults_stats}(a) and (d)) and the severity ((b)-(c), (e)-(f)).
The Fig.~\ref{fig:permanent_faults_stats}(a) and (d) shows both the SDC probability (in the form of persistant occurance) and the severity ((b)-(c), (e)-(f)) in detail.
% The used metrics for permanent faults are the following:
We register an SDC for a given fault if any persistent FP or FN is found during the sequence with a severity of at least level $L$. The severity $L$ is quantified as the average area occupied by the blob (for FP normalized by the image size, for FN by the TP blob size, see above).
The severity levels are varied from $0\%$ to $15\%$ in Fig.~\ref{fig:permanent_faults_stats} to illustrate the effect of softening or hardening of the safety requirements.
As the severity of a fault is again expected to depend on the bit position of the injected fault, we present both bit-selected and bit-averaged numbers in Fig.~\ref{fig:permanent_faults_stats}(b,c,e,f).

% We see in Fig.~\ref{fig:permanent_faults_stats} that 
In this figure, the permanent faults in neurons and weights have a probability of $1.8\%$ and $3\%$ to create persistent ghost FP objects with a minimal area of $L>0$, respectively. With $L>15\%$ of an image area, this reduces to $0.9\%$ and $2.9\%$, respectively. On average, faults hitting MSB bit in weights on this model have $96\%$ probability to manifest into a persistent FP blob of area $>81\%$. 
%Accordingly, the chances are about $2\%$ and $2.5\%$ create these blobs of area $>15\%$. 
On the other hand, persistent FN blobs incorrectly indicating vacant spaces occur with a much lower chance. 
Bitflips cause persistent objects only in the highest exponential bits in case of neurons or in the MSB bits in the case of weights. This observation is consistent with the findings from transient faults in Sec.~\ref{sec:transient_faults}.
Using the given area occupancy metrics, permanent weight faults have a higher severity than neuron faults; in particular, weight faults on average induce massive ghost FP blobs of $>83\%$ of the image area.

%Fig. \ref{fig:1}, \ref{fig:2} and \ref{fig:3} quantify the severities of a yolov3 model and lyft dataset. 
%It can be seen that permanent faults on neurons have higher probablity of creating ghost blobs. This is similar to the effect of transient faults on neurons.
%This is not the same when the permanent faults hit the weights except the MSB bit. 
%Faults hitting the weights can create the ghots objects with an area as high as $79\%$ and $sim55\%$ in the case neuron faults.
%On an average ghost objects occurance probablity by persistent FPs is $3.3\%$. 
%Similarly, there is $2.35\%$ chance for yolov3 model to see ghost objects whose average area is $>$ $15\%$ which is a safety critical in a real time situation.
%
%
%\subsection{Impact of missed detections (FNs) due to permanent faults}
%
%The persistent miss-detection (FNs) due to these \textit{stuck-at} faults are most critical as they are degrading the base known performance of the model.
%There's an average probablity of $1.7\%$ chance of persistent miss-detections. Overall probability of continuosly missing detections of area $>$ $15\%$ is $0.2\%$.
%Similar to FP-blob manifestations, neurons faults have higher probablity for higher FN-blobs.